\documentclass{article}

\usepackage[final]{neurips_2022}

\usepackage[utf8]{inputenc} 
\usepackage[T1]{fontenc}    
\usepackage{microtype}
\usepackage{graphicx}
\usepackage{wrapfig}
\usepackage{subfigure}
\usepackage{nicefrac}       
\usepackage{booktabs} 
\usepackage{paralist}
\usepackage{multirow}
\usepackage{amsmath}
\usepackage{amssymb}
\usepackage{amsfonts}       
\usepackage{csquotes}
\usepackage[dvipsnames]{xcolor}
\usepackage[section]{placeins}
\usepackage{colortbl}
\usepackage{url}            
\usepackage{hyperref}
\usepackage[ruled,vlined]{algorithm2e}

\hypersetup{
  citebordercolor=Periwinkle,
}

\title{Towards Safe Reinforcement Learning \\ with a Safety Editor Policy}

\author{%
  Haonan Yu,\ \ Wei Xu,\ \ and\ \ Haichao Zhang \\
  Horizon Robotics\\
  Cupertino, CA 95014 \\
  \texttt{\{haonan.yu,wei.xu,haichao.zhang\}@horizon.ai} \\
}

\usepackage{scalerel,stackengine}
\stackMath
\newcommand\reallywidehat[1]{%
\savestack{\tmpbox}{\stretchto{%
  \scaleto{%
    \scalerel*[\widthof{\ensuremath{#1}}]{\kern-.6pt\bigwedge\kern-.6pt}%
    {\rule[-\textheight/2]{1ex}{\textheight}}
  }{\textheight}%
}{0.5ex}}%
\stackon[1pt]{#1}{\tmpbox}%
}

\newcommand{\expect}{\mathop{\mathbb{E}}}

\newcommand{\ie}{\textit{i.e.},}
\newcommand{\eg}{\textit{e.g.},}

\newcommand{\vs}{\textit{vs.}}

\newcommand{\name}{SEditor}
\newcommand{\piu}{\pi_{\phi}}
\newcommand{\pis}{\pi_{\psi}}
\newcommand{\pia}{\pi_{\psi\circ\phi}}

\definecolor{bleudefrance}{rgb}{0.19, 0.55, 0.91}
\definecolor{cadmiumorange}{rgb}{0.93, 0.53, 0.18}

\begin{document}

\maketitle

\begin{figure}[!h]
    \centering
    \includegraphics[width=0.6\textwidth]{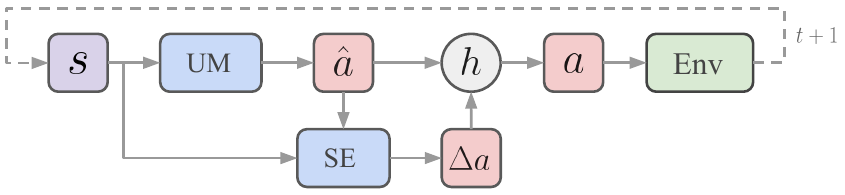}
    \caption{\name{}'s framework with two policies utility maximizer (UM) and safety editor (SE).
    Given a state $s$, UM proposes a preliminary action $\hat{a}$ that aims at maximizing the utility reward.
    Then SE outputs an adjustment $\Delta a$ to it based on $s$ and $\hat{a}$ itself, to ensure a low constraint violation rate.
    The action proposal $\hat{a}$ and edit $\Delta a$ are fed to an action editing function $h$ that outputs a final action $a$ to interact with the environment.
    Both UM and SE are stochastic policies modeled as neural networks and are trained via SGD.
    }
    \label{fig:framework}
\vspace{2ex}
\end{figure}

\begin{abstract}
We consider the safe reinforcement learning (RL) problem of maximizing utility with extremely low constraint violation rates.
Assuming no prior knowledge or pre-training of the environment safety model given a task, an agent has to learn, via exploration, which states and actions are safe.
%
%
A popular approach in this line of research is to combine a model-free RL algorithm with the Lagrangian method to adjust the weight of the constraint reward relative to the utility reward dynamically.
It relies on a single policy to handle the conflict between utility and constraint rewards, which is often challenging.
We present \name{}, a two-policy approach that learns a safety editor policy transforming potentially unsafe actions proposed by a utility maximizer policy into safe ones.
The safety editor is trained to maximize the constraint reward while minimizing a hinge loss of the utility state-action values before and after an action is edited.
\name{} extends existing safety layer designs that assume simplified safety models, to general safe RL scenarios where the safety model can in theory be arbitrarily complex.
As a first-order method, it is easy to implement and efficient for both inference and training.
%
%
On 12 Safety Gym tasks and 2 safe racing tasks, \name{} obtains much a higher overall safety-weighted-utility (SWU) score than the baselines, and demonstrates outstanding utility performance with constraint violation rates as low as once per 2k time steps, even in obstacle-dense environments.
On some tasks, this low violation rate is up to 200 times lower than that of an unconstrained RL method with similar utility performance.
Code 
\ifdefined\isaccepted
is available at \url{https://github.com/hnyu/seditor}.
\else 
will be made public.
\fi
%

%
%
%
\end{abstract}

\section{Introduction}
\label{sec:introduction}
\vspace{-2ex}
Safety has been one of the major roadblocks in the way of deploying reinforcement learning (RL) 
to the real world.
Although RL for strategy and video games has achieved great successes~\citep{Silver2016,Vinyals2019GrandmasterLI,OpenAI2019Five}, the cost of executing actions that lead to catastrophic failures in these cases is low: at most losing a game.
On the other hand, RL for control has also mostly been studied in virtual simulators~\citep{Brockman2016,tassa2020dmcontrol}. 
Aside from the data collection consideration, to circumvent the safety issue (\eg\ damages to real robots and environments) is another important cause of using these simulators.
When RL is sometimes applied to real robots~\citep{Kim2003,Levine2016,OpenAI2019}, data collection and training are configured in restrictive settings to ensure safety.
Thus to promote the deployment of RL to more real-world scenarios, safety is a critical topic.
In this paper, we study a safe RL problem of maximizing utility with extremely low constraint violation rates.
%

There are two general settings of safe RL. 
Some existing works make the assumption that the environment safety model is a known prior. The agent is able to query an oracle function to see if any state is safe or not, \emph{without} actually visiting that state.
This includes having access to a well calibrated dynamics model~\citep{berkenkamp2017safe,chow2018lyapunovbased,thomas2021safe} or the set of safe/unsafe states ~\citep{turchetta2020safe,Luo2021,Li2021SafeRL}.
Or they assume that a pre-training stage of safe/unsafe states or policies is performed on offline data, and then the learned safety knowledge is transferred to main tasks~\citep{Dalal2018,miret2020safety,Thananjeyan2021}.
With these assumptions, they are able to achieve few or even zero constraint violations during online exploration.
However, such assumptions also put restrictions on their applicable scenarios.

This paper focuses on another setting where no prior knowledge or pre-training of safety models is assumed.
The agent only gets feedback on which states are unsafe from its exploration experience.
In other words, it has to learn the safety model (implicitly or explicitly) from scratch and the constraint budget can only be satisfied \emph{asymptotically}.
%
%
Although we cannot completely avoid constraint violations during online exploration, this setting can be especially helpful for sim-to-real transfer~\citep{Zhao2020}, or when the training process can be protected from damages.
Prior works combine model-free RL~\citep{Bhatnagar2012,Ray2019,Tessler2019,bohez2019value,stooke2020responsive,zhang2020order,qin2021density} or model-based RL~\citep{as2022constrained} with the Lagrangian method~\citep{Bertsekas1999}.
This primal-dual optimization dynamically adjusts the weight of the constraint reward relative to the utility reward, depending on how well the violation rate target is being met.
A single policy is then trained from a weighted combination of the utility and constraint rewards.
However, reconciling utility maximization with constraint violation minimization usually poses great challenges to this single policy.

In this paper, we present \textbf{\name{}}, a general safe RL approach that decomposes policy learning across two polices (Figure~\ref{fig:framework}). 
The utility maximizer (UM) policy is only responsible for maximizing the utility reward without considering the constraints.
Its output actions are potentially unsafe. 
The safety editor (SE) policy then transforms these actions into safe ones. 
It is trained to maximize the constraint reward while minimizing a hinge loss of the utility state-action values before and after an action is edited.
Both UM and SE are trained in an off-policy manner for good sample efficiency.
Our two-policy paradigm is largely inspired by existing safety layer designs~\citep{Dalal2018,pham2018optlayer,Cheng2019,Li2021SafeRL} which simplify (\eg\ to linear or quadratic) environment safety models.
The high-level idea is the same though: modifying a utility-maximizing action only when necessary.

We evaluate \name{} on 12 Safety Gym~\citep{Ray2019} tasks and 2 safe car racing tasks adapted from \citet{Brockman2016}, targeting at \emph{very} low violation rates.
%
%
%
\name{} obtains a much higher overall safety-weighted-utility (SWU) score  (defined in Section~\ref{sec:experiments}) than four baselines.
It demonstrates outstanding utility performance with constraint violation rates as low as once per 2k time steps, even in obstacle-dense environments.
Our results reveal that the two-policy cooperation is critical, while simply doubling the size of a single policy network will not lead to comparable results.
The choices of the action distance function and editing function are also important in certain circumstances.
%
%
%
%
In summary, our contributions are: 
\begin{compactenum}[a)]
    \item We extend the existing safety layer works to more general safe RL scenarios where the environment safety model can in theory be arbitrarily complex.
    \item We present \name{}, a first-order, easy-to-implement approach that is trained by SGD like most model-free RL methods. 
    It is efficient during both inference and training, as it does not solve a multi-step inner-level optimization problem when transforming an unsafe action into a safe one.
    \item When measuring the distance between an action and its edited version, we show that in some cases the hinge loss of their state-action values is better than the usual L2 distance~\citep{Dalal2018,pham2018optlayer,Li2021SafeRL} in the action space.
    We further show that an additive action editing function introduces an effective inductive bias for the safety editor.
    \item We propose the safety-weighted-utility (SWU) score for quantitatively evaluating a safe RL method. The score is a soft indicator of the \emph{dominance} defined by \citet{Ray2019}.
    \item Finally, we achieve outstanding utility performance even with an extremely low constraint violation rate ($5\times10^{-4}$) in dense-obstacle environments. 
    On some tasks, this low violation rate is up to 200 times lower than that of a unconstrained RL method with similar utility performance.
\end{compactenum}

\section{Preliminaries}
The safe RL problem can be defined as policy search in a constrained Markov decision process (CMDP) $(\mathcal{S},\mathcal{A},p,r,c)$.
The state space $\mathcal{S}$ and action space $\mathcal{A}$ are both assumed to be continuous.
The environment transition function $p(s_{t+1}|s_t,a_t)$ determines the probability density of reaching $s_{t+1}\in\mathcal{S}$ after taking action $a_t\in\mathcal{A}$ at state $s_t\in\mathcal{S}$.
The initial state distribution $\mu(s_0)$ determines the probability density of an episode starting at state $s_0$.
Both $p(s_{t+1}|s_t,a_t)$ and $\mu(s_0)$ are usually unknown to the agent.
For every transition $(s_t,a_t,s_{t+1})$, the environment outputs a scalar $r(s_t,a_t,s_{t+1})$ which we call the \emph{utility} reward.
Sometimes one uses the expected reward of taking $a_t$ at $s_t$ as $r(s_t,a_t)\triangleq\expect_{s_{t+1}\sim p(\cdot|s_t,a_t)}r(s_t,a_t,s_{t+1})$ for a simpler notation.
Similarly, the environment also outputs a scalar $c(s_t,a_t)$ as the cost.
To unify the reward and cost notations, we define the \emph{constraint} reward $r_c(s_t,a_t)\triangleq -c(s_t,a_t)\le 0$, and the CMDP becomes $(\mathcal{S},\mathcal{A},p,r,r_c)$.
For both utility and constraint rewards, the value is \emph{the higher the better}.
Finally, we denote the agent's policy as $\pi(a_t|s_t)$ which dictates the probability density of taking $a_t$ at $s_t$.

For each $s_t$, the utility state value of following $\pi$ is denoted by $V^{\pi}(s_t)=\expect_{\pi,p}\sum_{t'=t}^{\infty}\gamma^{t'-t}r(s_{t'},a_{t'})$, and the utility state-action value is denoted by $Q^{\pi}(s_t,a_t)=r(s_t,a_t)+\gamma \expect_{s_{t+1}\sim p}V^{\pi}(s_{t+1})$,
where $\gamma\in[0, 1)$ is the discount for future rewards.
Similarly, we can define $V_c^{\pi}$ and $Q_c^{\pi}$ for the constraint reward.
Then we consider the safe RL objective:
\begin{equation}
\label{eq:objective}
\max_{\pi}\expect_{s_0\sim\mu}V^{\pi}(s_0),\ \ s.t.\ \ \expect_{s_0\sim\mu}V_c^{\pi}(s_0) + C\ge 0,
\end{equation}
where $C\ge 0$ is the constraint violation budget.
It might be unintuitive to specify $C$ for a discounted return, so one can rewrite
\vspace{-2ex}
\begin{equation}
\label{eq:constraint_c}
\expect_{\mu,\pi,p}\sum_{t=0}^{\infty}\gamma^t (r_c(s_t,a_t)+c)\ge 0,
\end{equation}
and specify the per-step budget $c$ instead, relating to $C$ by $\sum_{t=0}^{\infty}\gamma^t c = \frac{c}{1-\gamma} = C$.
Note that $c$ is not strictly imposed on every step. 
Instead, it is only in the average sense (by a discount factor), treated as the \emph{violation rate target}.

The Lagrangian method~\citep{Bertsekas1999} converts the constrained optimization problem Eq.~\ref{eq:objective} into an unconstrained one by introducing a multiplier $\lambda$:
\begin{equation}
\label{eq:unconstrained_objective}
\min_{\lambda\ge 0}\max_{\pi}\left[\expect_{s_0\sim\mu}V^{\pi}(s_0)+\lambda\left(\expect_{s_0\sim\mu}V_c^{\pi}(s_0) + C \right) \right].
\end{equation}
Intuitively, it dynamically adjusts the weight $\lambda$ according to how well the constraint state value satisfies the budget, by evaluating (approximately) the difference
\begin{equation}
\label{eq:lambda_train}
\Lambda_{\pi} \triangleq \expect_{s_0\sim\mu}V_c^{\pi}(s_0) + C = \expect_{\mu,\pi,p}\sum_{t=0}^{\infty}\gamma^t (r_c(s_t,a_t)+c),
\end{equation}
which is the gradient of $\lambda$ given $\pi$ in Eq.~\ref{eq:unconstrained_objective}.
%
%
When optimizing $\pi$ given $\lambda$, Eq.~\ref{eq:unconstrained_objective} becomes
\begin{equation*}
\begin{array}{l}
    \displaystyle\max_{\pi}\expect_{s_0\sim\mu}\left[V^{\pi}(s_0)+\lambda V_c^{\pi}(s_0)\right]
    =\displaystyle\max_{\pi}\expect_{\mu,\pi,p}\sum_{t=0}^{\infty}\gamma^t\left(r(s_t,a_t)+\lambda r_c(s_t,a_t)\right).\\
\end{array}
\end{equation*}
Thus $\lambda$ can be seen as the weight of the constraint reward when it is combined with the utility reward to convert multi-objective RL into single-objective RL.
This objective can be solved by typical model-free RL algorithms.
For practical implementations, previous works~\citep{Ray2019,Tessler2019,bohez2019value} usually perform gradient ascent on the parameters of $\pi$ and gradient descent on $\lambda$ simultaneously, potentially with different learning rates.

\vspace{-2ex}
\section{Approach}
\label{sec:approach}
We consider a pair of cooperative policies.
The first policy utility maximizer (UM) denoted by $\piu$, optimizes the utility reward by proposing a preliminary action $\hat{a}\sim \piu(\cdot|s)$ which is potentially unsafe. 
The second policy safety editor (SE) denoted by $\pis$, edits the preliminary action by $\Delta a\sim \pis(\cdot|s,\hat{a})$ to ensure safety, and the result action 
$a= h(\hat{a},\Delta a)$ is output to the environment, where $h$ represents an editing function.
Note that we condition SE on UM's output $\hat{a}$.
Together the two policies cooperate to maximize the agent's utility while maintaining a safe condition (Figure~\ref{fig:framework}). 
For simplicity, we will denote the overall composed policy by $\pia(a|s)$.

\textbf{Motivation.} \name{} decomposes a difficult policy learning task that maximizes both utility and safety, into two easier subtasks that focus on either utility or safety, based on the following considerations:
\begin{compactenum}[a)]
    \item Different effective horizons. In most scenarios, safety requires either responsive actions~\citep{Dalal2018}, or planning a number of steps ahead to prevent the agent entering non-recoverable states. 
    Thus SE's actual decision horizon could be short depending on the nature of the safety constraints. 
    This is in contrast to UM’s decision horizon which is usually long for goal achieving. 
    In other words, we could expect the optimization problem of SE to be easier than that of UM, and SE has a chance of being learned faster if separated from UM.
    \item Guarded exploration. From the perspective of UM, its MDP (precisely, action space) is altered by SE. 
    UM’s actions are guarded by the barriers set up by the SE. 
    Instead of UM being discouraged for an unsafe action (\ie\, punished with negative signal), SE gives suggestions to UM by redirecting the unsafe action to a safe but also utility-high action to continue its exploration. 
    This guarded exploration leads to a better overall exploration strategy because safety constraints are less likely to hinder UM’s exploration (Figure~\ref{fig:pointgoal2_trajs} illustration).
\end{compactenum}

%
\textbf{Objectives.}\ \ We employ an off-policy actor-critic setting for training the two policies.
Given an overall policy $\pia$, we can use typical TD backup to learn $Q^{\pia}(s,a)$ and $Q_c^{\pia}(s,a)$ parameterized as $Q(s,a;\theta)$ and $Q_c(s,a;\theta)$ respectively, where we use $\theta$ to collectively represent the network parameters of the two state-action values.
Given $s_{t+1}\sim p(\cdot|s_t,a_t)$ and $a_{t+1}\sim\pia(\cdot|s_{t+1})$, the Bellman backup operator (point estimate) for the utility state-action value is 
\begin{equation}
\label{eq:bellman}
\mathcal{T}^{\pia}Q(s_t,a_t;\theta) = r(s_t,a_t) + \gamma Q(s_{t+1},a_{t+1};\theta),
\end{equation}
with the backup operator for the constraint state-action value $Q_c$ defined similarly with $r_c$.
Both $Q(s,a;\theta)$ and $Q_c(s,a,;\theta)$ can be learned on transitions $(s_t,a_t,s_{t+1})$ sampled from a replay buffer.
%

For off-policy training of $\phi$ and $\psi$, we first transform Eq.~\ref{eq:unconstrained_objective} into a bi-level optimization surrogate as:
\begin{equation}
\label{eq:surrogate_obj}
\begin{array}{llll}
     \text{(a)} & \displaystyle\max_{\phi,\psi}\left[\expect_{s\sim\mathcal{D},a\sim\pia(\cdot|s)}\left(Q(s,a;\theta) + \lambda Q_c(s,a;\theta) \right)\right],  &
     \text{(b)} & \displaystyle\min_{\lambda\ge 0}\lambda\Lambda_{\pia}\\
\end{array}.
\end{equation}
%
%
$\mathcal{D}$ denotes a replay buffer and $\Lambda_{\pia}$ is defined by Eq.~\ref{eq:lambda_train}.
We basically have a historical marginal state distribution for training the policies, but still \emph{use the initial state distribution} $\mu$ for training $\lambda$.
The motivation for this difference is, when tuning $\lambda$, we should always care about how well the policy satisfies our constraint budget starting with $\mu$ but not with some historical state distribution.

We continue transforming the off-policy objective (Eq.~\ref{eq:surrogate_obj}, a) into two, 
\begin{equation}
\label{eq:objectives2}
\begin{array}{@{}llll@{}}
\text{(a)} & \displaystyle\max_{\textcolor{bleudefrance}{\phi}}\expect_{\substack{s\sim\mathcal{D}, \textcolor{bleudefrance}{\hat{a}}\sim\pi_{\textcolor{bleudefrance}{\phi}}(\cdot|s),\\ \textcolor{bleudefrance}{\Delta a}\sim\pis(\cdot|s,\textcolor{bleudefrance}{\hat{a}}), \\ \textcolor{bleudefrance}{a}=h(\textcolor{bleudefrance}{\hat{a}},\textcolor{bleudefrance}{\Delta a})}} \Big[ Q(s,\textcolor{bleudefrance}{a};\theta)\Big], & 
\text{(b)} & \displaystyle\max_{\textcolor{cadmiumorange}{\psi}}\expect_{\substack{s\sim\mathcal{D}, \hat{a}\sim\piu(\cdot|s),\\ \textcolor{cadmiumorange}{\Delta a}\sim\pi_{\textcolor{cadmiumorange}{\psi}}(\cdot|s,\hat{a}), \\ \textcolor{cadmiumorange}{a}=h(\hat{a},\textcolor{cadmiumorange}{\Delta a})}}\Big[-d(\textcolor{cadmiumorange}{a},\hat{a})+\lambda Q_c(s,\textcolor{cadmiumorange}{a};\theta)\Big],\\
\end{array}
\end{equation}
where $d(a,\hat{a})$ is a distance function measuring the change from $\hat{a}$ to $a$.
It is not necessarily proportional to $\Delta a$ because the editing function $h$ could be nonlinear.
%
The role of SE $\pis$ is to maximize the constraint reward while minimizing some distance between the actions before and after the modification.
The role of UM $\piu$ is to only maximize the utility reward. 
However, it only can only do so \emph{through the lens of} SE $\pis$. 
In other words, SE has actually changed the action space (also MDP) of UM.
This property corresponds to the motivation of guarded exploration mentioned earlier. 

We would like to emphasize that, unlike the safety layer~\citep{Dalal2018} which projects unsafe actions based on instantaneous safety costs, the training objective (Eq.~\ref{eq:objectives2}, b) for SE relies on a safety critic $Q_c$ which is learned as the expected future constraint return. 
Therefore, maximizing this critic will take into account long-term safety behaviors. 
In other words, if there are non-recoverable states where SE can’t do anything to ensure safety, it will edit the agent’s actions long before that to avoid entering those states.
%

%
\textbf{Action editing function.}\ \ We choose the editing function $h$ to be mainly additive and non-parametric. 
Without loss of generality, we assume a bounded action space $[-A,A]^M$, and that both $\hat{a}$ and $\Delta a$ are already in this space.
Then we define $a=h(\hat{a},\Delta a)=\min(\max(\hat{a}+2\Delta a, -A), A)$, where $\min$ and $\max$ are element-wise. 
The multiplication by 2 and the clipping make sure that $a\in[-A,A]^M$, namely, SE has a full control over the final action in that it can overwrite UM's action if necessary.
Because SE can be arbitrarily complex, its output $\Delta a$ could depend on the current state $s$ and the action proposal $\hat{a}$ in an arbitrarily complex way. 
Thus even though the additive operation is simple, the overall editing process is already general enough to represent any modification.

This additive editing function is motivated by \emph{constraint sparsity}. 
Usually, constraint violations are only triggered for some states. 
Most often, the action proposal by UM is already safe if the agent is far away from obstacles. 
To explicitly introduce this inductive bias, we use the additive editing function which ensures that the majority of SE’s modifications are close to 0. 
This makes the optimization landscape of SE easier (see Figure~\ref{fig:editing_function_frames} Appendix~\ref{app:editing_function} for some empirical observations).

\textbf{Distance function.}\ \ Prior works on safety layer design, such as \citet{Dalal2018,pham2018optlayer,Li2021SafeRL}, set the distance function $d(\cdot,\cdot)$ as the L2 distance.
Later we will show that L2 is not always the best option.
%
%
Instead, we use the hinge loss of the utility state-action values of $\hat{a}$ and $a$:
\begin{equation}
\label{eq:hinge_loss}
d(a,\hat{a}) \triangleq \max(0, Q(s,\hat{a};\theta) - Q(s,a;\theta))
\end{equation}
This loss is zero if the edited action $a$ already obtains a higher utility state-action value than the preliminary action $\hat{a}$.
In this case, only the constraint Q is optimized by $\pis$.
Otherwise, the inner part of Eq.~\ref{eq:objectives2} (b) is recovered as $Q(s,a;\theta)+\lambda Q_c(s,a;\theta)$,
as we can drop the term $-Q(s,\hat{a};\theta)$ due to its gradient \textit{w.r.t.} $\psi$ being zero.
Our distance function in the utility Q space is more appropriate than the L2 distance in the action space, because eventually we care about how the utility changes after the action is edited. 
The L2 distance between the two actions is only an approximation to the change, from the perspective of the Taylor series of $Q(s,a;\theta)$.

\begin{figure}[!t]
    \centering
    \includegraphics[width=0.5\columnwidth]{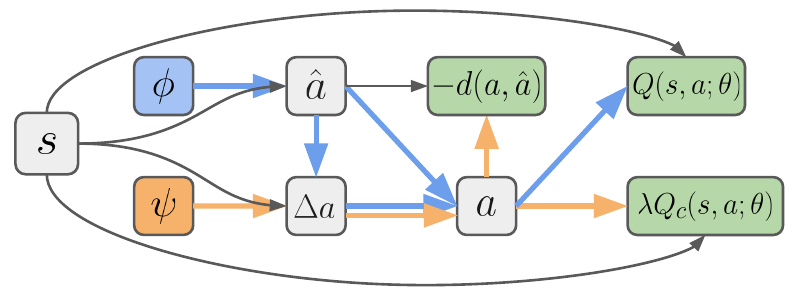}
    \caption{The computational graph of Eq.~\ref{eq:objectives2}.
    Nodes denote variables and edges denote operations.
    The {\color{LimeGreen}green} blocks are negative losses,
    the {\color{RoyalBlue}blue} paths are the (reversed) gradient paths of $\phi$, and the {\color{BurntOrange}orange} paths are the (reversed) gradient paths of $\psi$.
    Paths in black or {\color{RoyalBlue}blue} are detached for $\psi$, and paths in black or {\color{BurntOrange}orange} are detached for $\phi$.
    }
    \label{fig:computation_graph}
\vspace{-2ex}
\end{figure}

\textbf{Evaluating $\Lambda_{\pia}$.}\ \ Given a batch of rollout experiences $\{(s_n,a_n)\}_{n=1}^N$ following $\pia$, we approximate the gradient of $\lambda$ (Eq.~\ref{eq:lambda_train}) as 
\vspace{-3ex}
\begin{equation}
\label{eq:lambda_gradient}
\Lambda_{\pia} \approx \frac{1}{N}\sum_{n=1}^N r_c(s_n,a_n) + c,
\end{equation}
where $c$ is the violation rate target defined in Eq.~\ref{eq:constraint_c}.
Namely, after every rollout, we collect a batch of constraint rewards, compare each of them to $-c$, and use the mean of differences to adjust $\lambda$.
%
%
This approximation allows us to update $\lambda$ using mini-batches of data instead of having to wait for whole episodes to finish, or having to rely on estimated $V^{\pia}_c$ which is usually not accurate.
To reduce temporal correlation (which might affect the constraint evaluation) in the rollout batch data, we use multiple parallel environments (Appendix~\ref{app:hyperparameters}). 
This rollout batch is also the one that will be put into the replay buffer.

%

\textbf{Training.}\ \ To practically train the objectives, we apply SGD to Eq.~\ref{eq:surrogate_obj} (b) and Eq.~\ref{eq:objectives2} simultaneously, resulting in a first-order method for approximated bi-level optimization~\citep{likhosherstov2021debiasing}.
SGD is made possible by applying the re-parameterization trick (Appendix~\ref{app:parameterization}) to both $\hat{a}\sim\piu$ and $\Delta a\sim\pis$.
A computational graph of Eq.~\ref{eq:objectives2} is illustrated in Figure~\ref{fig:computation_graph}.
We would like to highlight that a big difference between our SE and the previous safety layers~\citep{Dalal2018,pham2018optlayer} is that SE directly uses a feedforward prediction $\Delta a\sim\pis$ (Eq.~\ref{eq:objectives2}, a) to replace the optimization result of the objective of transforming unsafe actions into safe ones.
%
%
In contrast, previous safety layers construct closed-form solutions or inner-level differentiable optimization steps (\eg\, quadratic programming~\citep{Amos2017}) for simplified safety models at every inference step.
Our overall approach generalizes to various constraint rewards (safety models) and action distance functions.
Finally, to encourage exploration, we add the entropy terms of $\piu$ and $\pis$ into Eq.~\ref{eq:objectives2}, with their weights dynamically adjusted according to two entropy targets following \citet{Haarnoja2018}.

Both UM and SE are trained from scratch in our experiments. 
Here we discuss an alternative situation where an existing policy is pre-trained to maximize utility, and we use it as the initialization for UM.
In this case, UM \emph{cannot} be frozen because its MDP (and hence its optimum) will be changed by the evolving SE.
Instead, it should be fine-tuned to adapt to the changing behaviors of SE. 
Potentially, a pre-trained UM will speed up the convergence of \name{}.
We will leave this to our future work.

\vspace{-2ex}
\section{Experiments}
\label{sec:experiments}
\vspace{-1ex}
\textbf{Baselines.}\ We compare \name{} with four baselines.
\begin{compactenum}[-]
    \item \textit{PPO-Lag} combines PPO~\citep{schulman2017proximal} with the Lagrangian method, as done in ~\citet{Ray2019}. 
    Since \textit{PPO} is an on-policy algorithm, we expect its sample efficiency to be much lower than off-policy algorithms.
    \item \textit{FOCOPS}~\citep{zhang2020order} is analogous to PPO-Lag with two differences: 1) there is no clipping of the importance ratio, and 2) a KL divergence regularization term with a fixed weight is added to the policy improvement loss, with an early stopping when this term averaged over a rollout batch data violates the trust region constraint.
    \item \textit{SAC-actor2x-Lag} combines SAC~\citep{Haarnoja2018} with the Lagrangian method. 
    Similar to \name{}, SAC-actor2x-Lag trains its policy on states sampled from the replay buffer but trains $\lambda$ on states generated by the \emph{rollout policy}.
    %
    %
    Its gradient of $\lambda$ is also estimated by Eq.~\ref{eq:lambda_gradient}.
    The policy network size is doubled. This is to match the model capacity of having two actors in \name{}.
    %
    %
    %
    %
    %
    %
    %
    \item \textit{SAC} serves as an unconstrained optimization baseline to calibrate the utility return.
\end{compactenum}
We choose not to compare with second-order CMDP approaches such as CPO~\citep{achiam17a} or PCPO~\citep{Yang2020Projection-Based} because they require (approximately) computing the inverse of the Fisher information matrix, which is prohibitive when the parameter space is large. 
Especially for our CNN based policies, second-order methods are impractical.

To analyze the key components of \name{}, we also evaluate two variants of it for ablation studies. 
\begin{compactenum}[-]
    \item \textit{\name{}-L2} defines $d(a,\hat{a})\triangleq\Vert a-\hat{a}\Vert^2$ as done in most prior works of safety layer. 
    %
    %
    \item \textit{\name{}-overwrite} makes SE directly overwrite UM's action proposal by $a=h(\hat{a},\Delta a)=\Delta a$. 
    %
    %
\end{compactenum}

All compared approaches including the variants of \name{}, share a common training configuration (\eg\, replay buffer size, mini-batch size, learning rate, etc) as much as possible. 
Specific changes are made to accommodate to particular algorithm properties (Appendix~\ref{app:hyperparameters}).

\textbf{Evaluation metric.}\ Following \citet{Ray2019}, one method \emph{dominates} another if ``it strictly improves on either return or cost rate and does at least as well on the other''. 
Accordingly, we emphasize that the utility reward or constraint violation rate should never be compared in isolation, as one could easily find a method that optimizes either metric very well.
Particularly in the experiments, we set the learning rate of the Lagrangian multiplier $\lambda$ much larger than that of the remaining parameters.
This ensures that any continued constraint violation will lead to a quick increase of $\lambda$ and drive the violation rate back to the target level. 
This conservative strategy makes some compared methods have similar constraint violation curves, but vastly different utility performance.
%
%
For a quantitative comparison of their final performance, we compute the \emph{safety-weighted-utility} (SWU) scores of the compared methods towards the end of training. The score is a soft indicator of dominance, and it is a product of two ratios:
\begin{equation*}
\text{SWU}\triangleq \min \left\{1,\frac{\text{ConstraintViolationRateTarget}}{\text{ConstraintViolationRate}}\right\} \times \frac{\text{UtilityScore}}{\text{UtilityScore}_{\text{UnconstrainedRL}}}.
\end{equation*}
%
%
The reason for having $\min\{1,\cdot\}$ is because we only care when the violation rate is higher than the target.
In practice, UtilityScore can be the episodic utility return (assuming positive) or the success rate.
We choose $\text{UtilityScore}_{\textsc{SAC}}$ as $\text{UtilityScore}_{\text{UnconstrainedRL}}$.
%
%
In our experiments, both UtilityScore and ConstraintViolationRate are averaged over the last $\frac{1}{10}$ of the training steps to reduce variances.

\begin{figure}[!t]
\centering
\resizebox{0.85\textwidth}{!}{
\begin{tabular}{@{}c@{}}
\includegraphics[width=\textwidth]{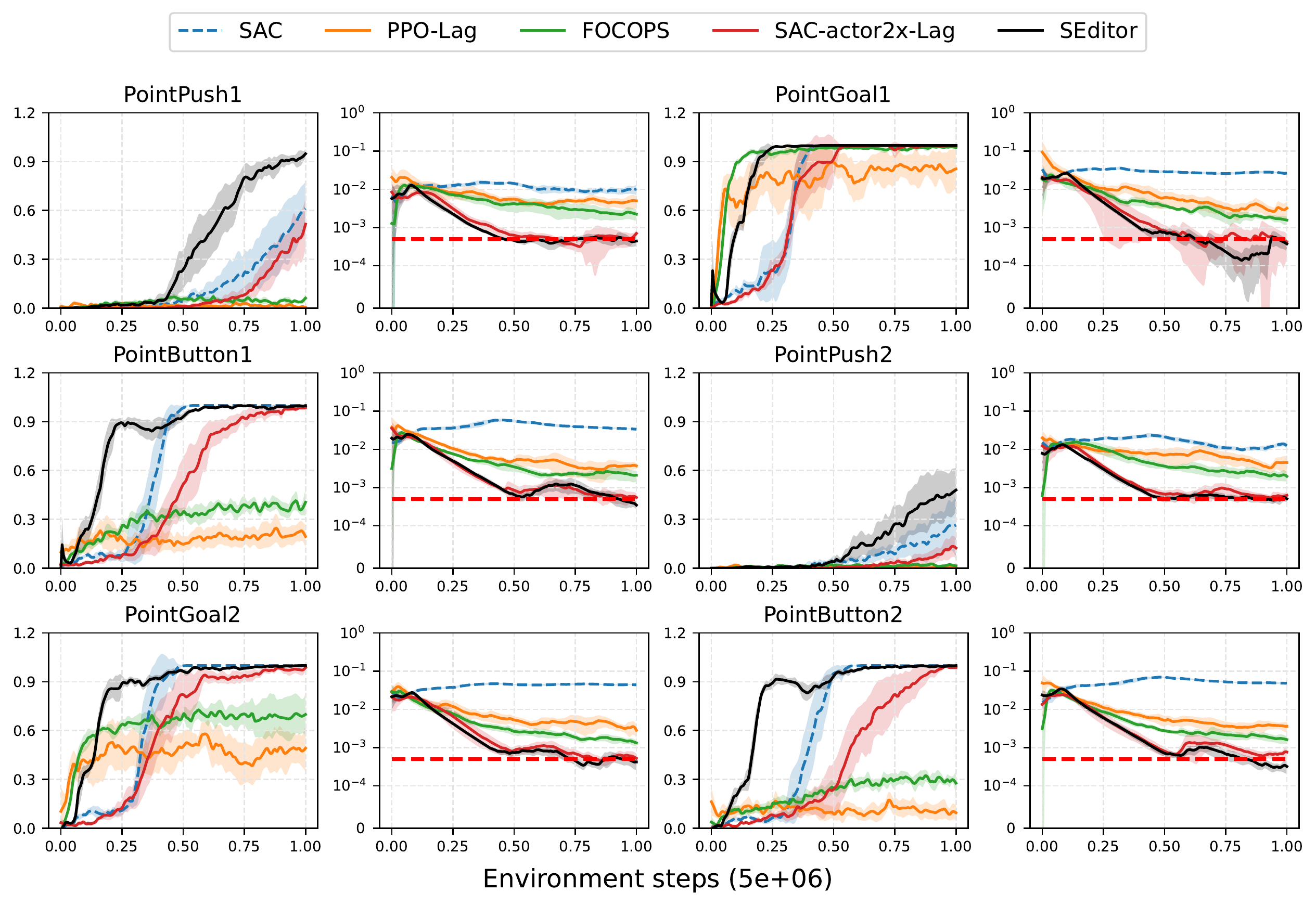}\\
\includegraphics[width=\textwidth]{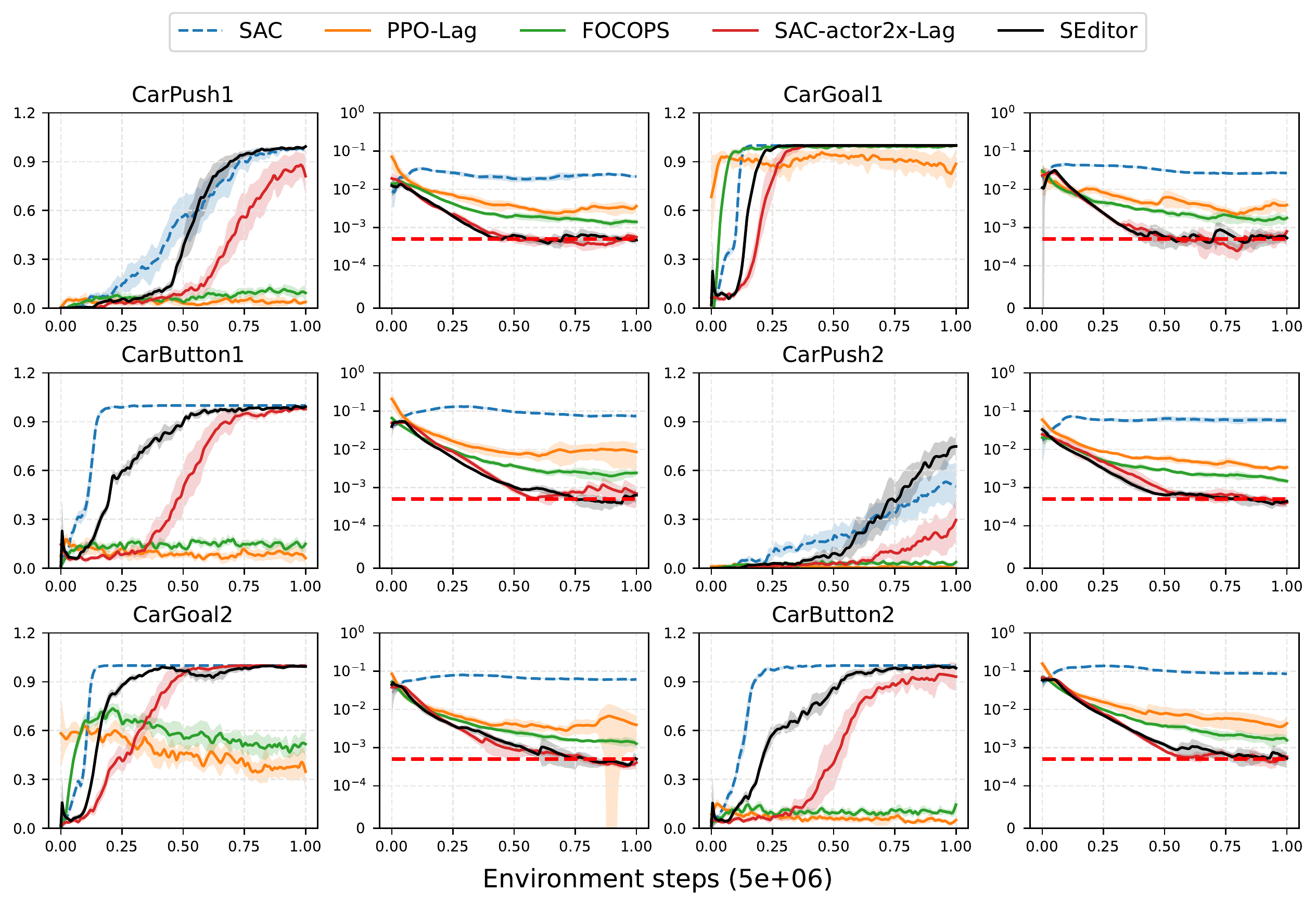}
\end{tabular}
}
\caption{
The training curves of the 12 Safety Gym tasks. 
Odd columns: $\uparrow$ success rate. Even columns: $\downarrow$ constraint violation rate (log scale). Red dashed horizontal lines: violation rate target $c=5\times10^{-4}$. Shaded areas: 95\% confidence interval (CI).
}
\label{fig:safety_gym_curves}
\vspace{-2ex}
\end{figure}

\textbf{Safety Gym tasks.}\ In Safety Gym~\citep{Ray2019}, a robot with lidar sensors navigates through cluttered environments to achieve goals.
We use the \textsc{Point} and \textsc{Car} robots in our experiments. 
Either of them has three tasks: \textsc{Goal}, \textsc{Button}, and \textsc{Push}.
Each task has two levels, where level 2 has more obstacles and a larger map size than level 1.
In total we have $3\times 2 \times 2=12$ tasks.
Whenever an obstacle is in contact with the robot, a constraint reward of $-1$ is given.
Thus the constraint violation rate can be calculated as the negative average constraint reward.
The utility reward is calculated as the decrement of the distance between the robot (\textsc{Goal} and \textsc{Button}) or box (\textsc{Push}) and the goal at every step.
We define a success as finishing a task within a time limit of $1000$.
We emphasize that the agent has \emph{no} prior knowledge of which states are unsafe.
The map layout is \emph{randomized} at the beginning of each episode.
This randomization poses a great challenge to the agent because it has to generalize the safety knowledge to new scenarios.

We customize the environments to equip the agent with a more advanced lidar sensor in two ways:
\begin{compactenum}[i)]
    \item It is a natural lidar instead of a pseudo one used by \citet{Ray2019}. 
    The natural lidar simulates a ray intersecting with an object from an origin. 
    The pseudo lidar can only detect object centers, while the natural lidar reveals shape information of objects, which is required to achieve a very low constraint threshold.    
    \item Our lidar has 64 bins while the original lidar has only 16 bins. 
    The 16-bin pseudo lidar's low precision is a bottleneck of achieving much fewer safety violations. 
    Without perception precision, the problem is ill-posed for a robot suffering from blind spots to avoid touching objects at all.
\end{compactenum}
We refer the reader to Appendix~\ref{app:safety_gym} for more details of the environment and tasks.
\ifdefined\isaccepted
Our customized Safety Gym is available at \url{https://github.com/hnyu/safety-gym}.
\else\fi

\begin{table}[!t]
    \centering
    \resizebox{\columnwidth}{!}{
    \begin{tabular}{@{}r|cccccccccccc|cc|cc@{}}
         & \multicolumn{12}{c}{Safety Gym} & \multicolumn{2}{|c|}{Safe Racing} & \multirow{2}{*}{Overall}& \multirow{2}{*}{Improvement}\\
         &\textsc{CP1}&\textsc{CG1}&\textsc{CB1}&\textsc{CP2}&\textsc{CG2}&\textsc{CB2}&\textsc{PP1}&\textsc{PG1}&\textsc{PB1}&\textsc{PP2}&\textsc{PG2}&\textsc{PB2}
         &\textsc{SR}&\textsc{SRO}&\\
         \hline
        
        SAC&0.02&0.02&0.01&0.01&0.01&0.01&0.05&0.02&0.01&0.04&0.01&0.01&0.14&0.03&\cellcolor{blue!15}0.03&\cellcolor{blue!15}3567\%\\
        PPO-Lag&0.01&0.11&0.00&0.00&0.04&0.01&0.00&0.13&0.03&0.00&0.09&0.01&0.04&0.04&\cellcolor{blue!15}0.04&\cellcolor{blue!15}2650\%\\
        FOCOPS&0.03&0.28&0.03&0.03&0.20&0.05&0.02&0.32&0.10&0.01&0.26&0.09&0.04&0.04&\cellcolor{blue!15}0.11&\cellcolor{blue!15}900\%\\
        SAC-actor2x-Lag&0.74&0.63&0.70&0.59&1.00&0.81&0.60&1.00&0.89&0.37&0.94&0.64&0.37&0.08&\cellcolor{blue!15}0.67&\cellcolor{blue!15}64\%\\
        \name{}&\textbf{1.01}&\textbf{0.94}&0.78&\textbf{1.49}&0.99&\textbf{0.95}&\textbf{1.55}&1.00&1.00&\textbf{1.78}&1.00&\textbf{1.00}&\textbf{1.28}&\textbf{0.57}&\cellcolor{blue!15}\textbf{1.10}&\cellcolor{blue!15}-\\

    \end{tabular}
    }
    \caption{
    The SWU scores of different methods.
    Task name abbreviations: \textsc{CP} - \textsc{CarPush}, \textsc{CG} - \textsc{CarGoal}, \textsc{CB} - \textsc{CarButton}, 
    \textsc{PP} - \textsc{PointPush}, \textsc{PG} - \textsc{PointGoal}, \textsc{PB} - \textsc{PointButton}, \textsc{SR} - \textsc{SafeRacing}, 
    and \textsc{SRO} - \textsc{SafeRacingObstacle}.
    }
    \label{tab:SWU}
\vspace{-4ex}
\end{table}

We set the constraint violation target $c=5\times 10^{-4}$, meaning that the agent is allowed to
violate any constraint only once every 2k steps on average.
This threshold is only $\frac{1}{50}$ of the threshold $c=0.025$ used by the original Safety Gym experiments~\citep{Ray2019}, highlighting the difficulty of our task.
%
%
Each compared approach is trained with 9 random seeds. 
The training curves are in Figure~\ref{fig:safety_gym_curves} and the SWU scores are in Table~\ref{tab:SWU}.
We see that \name{} obtains much higher SWU scores than the baselines on 7 of the 12 tasks, while being comparable on the rest.
%
%
%
While SAC-actor2x-Lag performs well among the baselines, its double-size policy network does not lead to results comparable to \name{}.
This suggests that \name{} does not simply rely on the large combined capacity of two policy networks to improve the performance, instead, its framework in Figure~\ref{fig:framework} matters.
Both PPO-Lag and FOCOPS missed all the constraint violation rate targets.
The unconstrained baseline SAC obtains better success rate sample efficiency than \name{} with the \textsc{Car} robot, but violates constraints much more.
Surprisingly, SAC is worse than \name{} with the \textsc{Point} robot regarding success rates, even without constraints.
%
%
Finally, we highlight that towards the end of training on \textsc{CarButton1/2}, \textsc{CarPush2}, and \textsc{CarGoal2}, \name{} violates constraints up to 200 times less than SAC does, while achieving success rates on par with SAC (black curves \vs dashed blue curves)!

\begin{figure}[!t]
\centering
\resizebox{0.97\columnwidth}{!}{
\begin{tabular}{@{}ll@{}}
\includegraphics[width=\columnwidth]{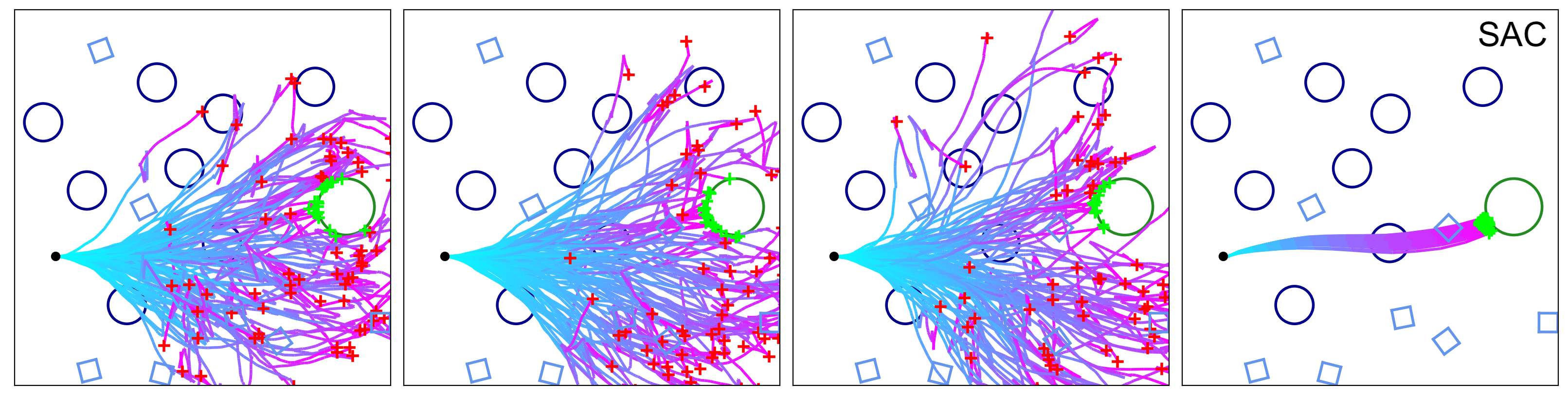}&
\includegraphics[width=\columnwidth]{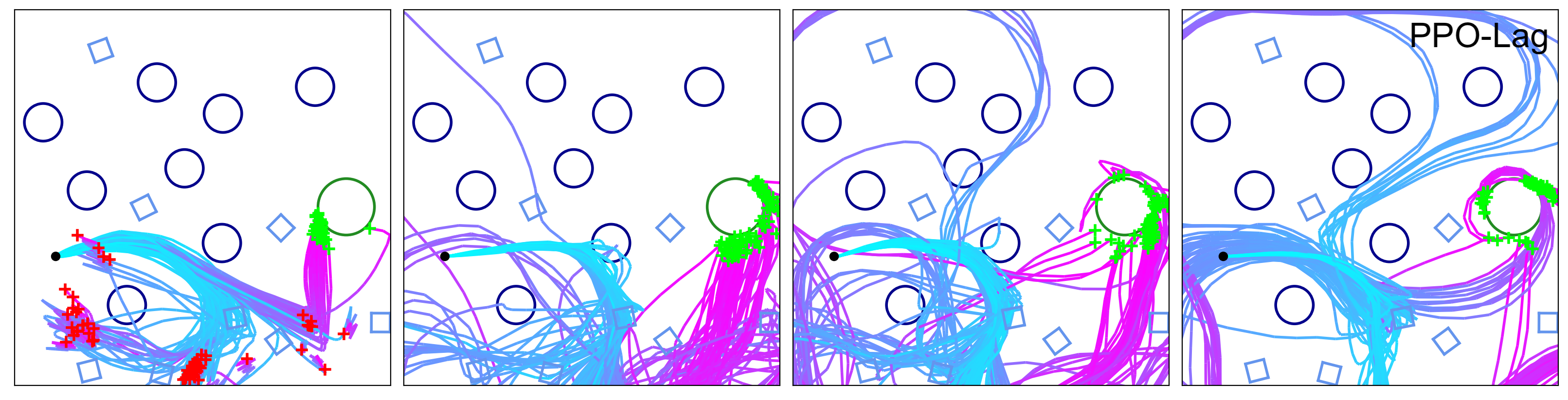}\\
\includegraphics[width=\columnwidth]{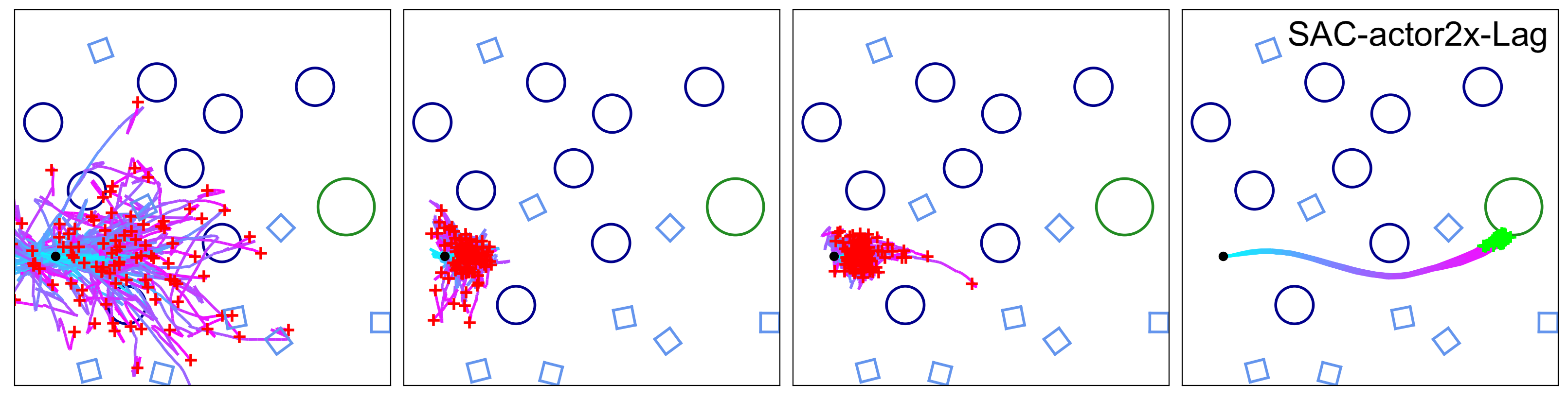}&
\includegraphics[width=\columnwidth]{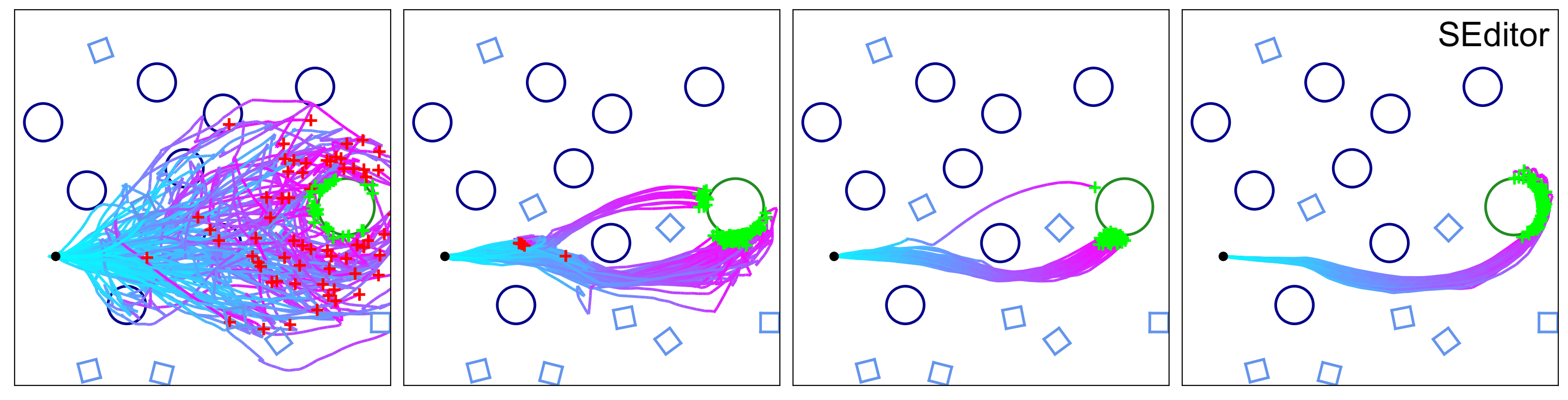}\\
\end{tabular}
}
\caption{Visualization of rollout trajectories at different training stages.
%
%
%
For each method from left to right, the trajectories are generated by checkpoints at 10\%, 20\%, 30\%, and 50\% of the training process.
On each trajectory, time flows from \textcolor{Cyan}{cyan} to \textcolor{Mulberry}{purple}.
A red cross \textcolor{red}{+} denotes a failed trajectory while a green cross \textcolor{green}{+} denotes a successful trajectory.
Each bird's-eye map is a sketch of the environment layout.
The small black dot is the robot's initial position.
The big green circle is the goal location and the other shapes are obstacles.
In summary, \name{} has a better exploration strategy because the barriers set up by SE can redirect UM's unsafe actions to continue its exploration, while SAC-actor2x-Lag's exploration will be hindered in when encountering similar obstacles.
}
\label{fig:pointgoal2_trajs}
\vspace{-3ex}
\end{figure}

To analyze the exploration behaviors of SAC, SAC-actor2x-Lag, PPO-Lag, and \name{}, we visualize their rollout trajectories on an example map of \textsc{PointGoal2}, at different training stages (the training is done on randomized maps but here we only evaluate on one map).
For each approach, the checkpoints at 10\%, 20\%, 30\%, and 50\% of the training process are evaluated and 100 rollout trajectories are generated for each of them. 
The trajectories are then drawn on a bird's-eye sketch map (Figure~\ref{fig:pointgoal2_trajs}).
We can see that although SAC is able to find the goal location in the very beginning, it tends to explore more widely regardless of the obstacles, and its final paths ignore obstacles.
SAC-actor2x-Lag learns to respect the constraints after 10\% of training, but the obstacles greatly hinder its exploration: the trajectories are confined in a small region. 
As a result, it takes some time for it to find the goal location.
PPO-Lag is slower at learning the constraints and it even hits obstacles at 50\%.
\name{} is able to quickly explore regions between obstacles and refine the navigation paths passing them instead of being blocked.

Our ablation studies (Appendix~\ref{app:ablation} Figure~\ref{fig:safety_gym_ablation}) show that, in this particular safe RL scenario,
\name{} is not sensitive to the choice of distance function: \name{}-L2 achieves similar results with \name{}.
However, the editing function does make a difference: \name{}-overwrite is clearly worse than \name{} in terms of utility performance.
This shows that the inductive bias of the edited action $a$ being close to the preliminary action $\hat{a}$ is very effective (see Appendix~\ref{app:editing_function}). 
Our two-stage ``propose-and-edit'' strategy is more efficient than outputting an action in one shot.

\begin{figure}[!t]
\centering
\resizebox{0.8\textwidth}{!}{
\begin{tabular}{@{}c@{}c@{}}
\multicolumn{2}{c}{\includegraphics[width=0.95\textwidth]{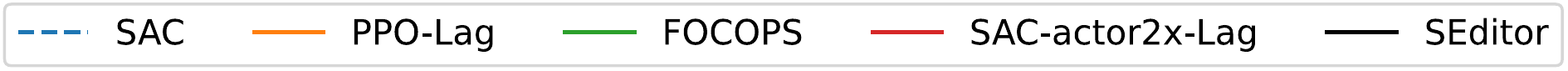}}\\
\includegraphics[width=0.5\textwidth]{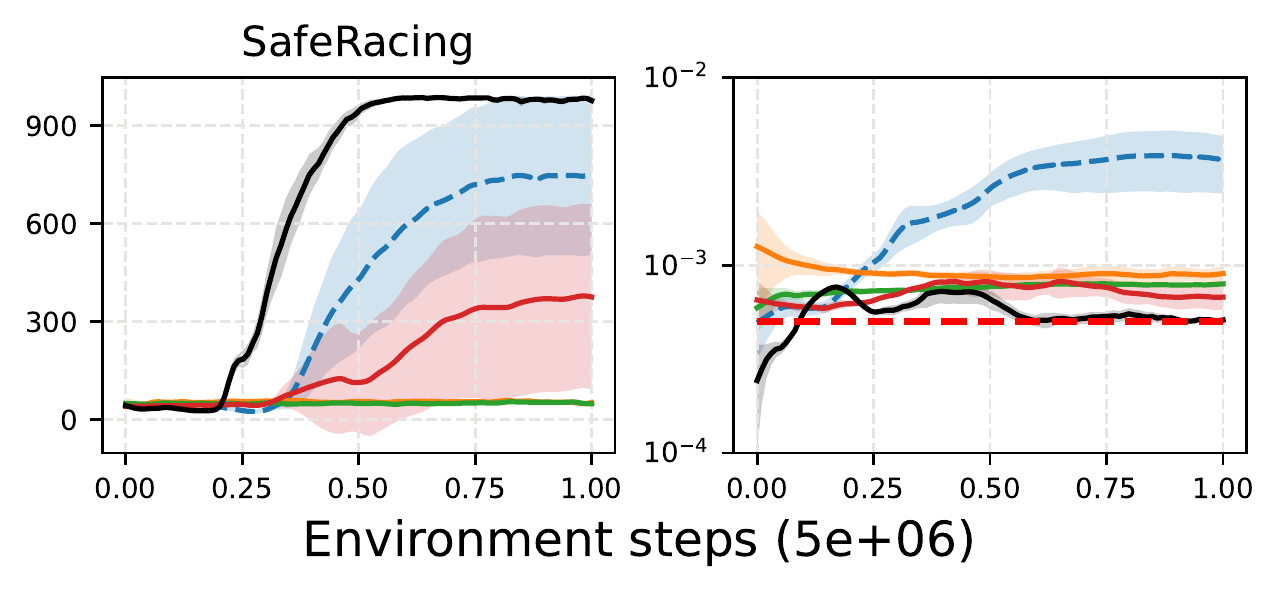}&
\includegraphics[width=0.5\textwidth]{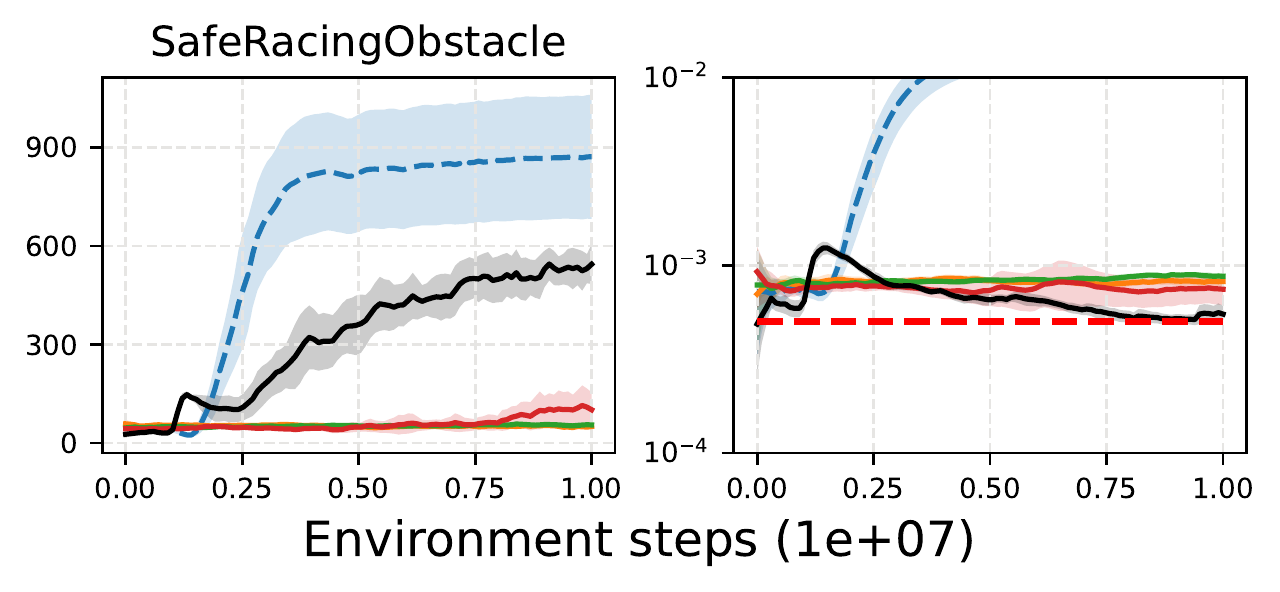}\\
\end{tabular}
}
\caption{The training curves of the safe racing tasks. 
Odd columns: $\uparrow$ undiscounted episode return. Even columns: $\downarrow$ constraint violate rate (log scale). Red dashed horizontal lines: violation rate target $c=5\times10^{-4}$. Shaded areas: 95\% confidence interval (CI).
}
\label{fig:safe_car_racing_curves}
\vspace{-2ex}
\end{figure}

%
%
\textbf{Safe racing tasks.}\ Our next two tasks, \textsc{SafeRacing} and \textsc{SafeRacingObstacle}, are adapted from the unconstrained car racing task in \citet{Brockman2016}.
The goal of either task is to finish a racetrack as fast as possible. 
The total reward of finishing a track is $1000$, and it is evenly distributed on the track tiles.
%
%
In \textsc{SafeRacing}, the car has to stay on the track and receives a constraint reward of $-1$ whenever driving outside of it.
In \textsc{SafeRacingObstacle}, the car receives a constraint reward of $-1$ if it hits an obstacle on the track, but there is no penalty for being off-track.
An episode finishes after every track tile is visited by the car, or after $1000$ time steps.
The track (length and shape) and obstacles (positions and shapes) are \emph{randomly} generated for each episode.
As in Safety Gym, this randomization could expose a never-experienced safety scenario to the agent at any time.
The agent's observations include a bird's-eye view image and a car status vector.
Note that this high-dimensional input space usually poses great challenges to second-order safe RL methods. 
Again, the constraint violation rate can be computed as the negative average constraint reward.
We set the constraint violation rate target $c=5\times 10^{-4}$. 
For evaluation, we use undiscounted episode return as the UtilityScore.
Each compared approach is trained with 9 random seeds.
The training curves are shown in Figure~\ref{fig:safe_car_racing_curves} and SWU scores in Table~\ref{tab:SWU}.
\name{} has much higher SWU scores than all baselines.
Moreover, it is the only one that satisfies the harsh violation rate target towards the end of training.
Surprisingly, even with constraints \name{} gets a much better utility return than SAC on \textsc{SafeRacing}.
One reason is that without the out-of-track penalty as racing guidance, the car easily gets lost on the map and collects lots of meaningless timesteps.
Although SAC gets a much higher return on \textsc{SafeRacingObstacle}, it greatly violates the constraint budget. 
%
%
Interestingly, the ablation studies show results that are complementary to those on the Safety Gym tasks (Appendix~\ref{app:ablation} Figure~\ref{fig:safe_racing_ablation}).
Now the action distance function makes a big difference.
Changing it to the L2 distance greatly impacts the utility return, especially on \textsc{SafeRacingObstacle} where almost no improvement is made.
%
%
This demonstrates that the closeness of two actions does not necessarily reflect the closeness of their state-action values (Appendix~\ref{app:hinge_vs_l2}).

\vspace{-2ex}
\section{Related Work}
\vspace{-1ex}
Safe RL is closely related to multi-objective RL~\citep{Roijers2013}, where the agent optimizes a scalarization of multiple rewards given a preference~\citep{Moffaert2013}, or finds a set of policies covering the Pareto front if no preference is provided~\citep{moffaert2015}.
%
%
In our case, the preference for the constraint reward is always changing, because we try to maintain it to a certain level instead of maximizing it.
%

%
%
Trust-region methods~\citep{achiam17a,chow2019lyapunovbased,Yang2020Projection-Based,Liu2022} for solving CMDPs put a constraint on the KL divergence between the new and old policies, where the KL divergence is second-order approximated.
For solving the surrogate objective at each iteration, the inverse of the Fisher information matrix is approximated.
\citet{Liu2022} decomposes the policy update step of CPO~\citep{achiam17a} into two steps, where the first E-step computes a non-parametric form of the optimal policy and the second M-step distills it to a parametric policy.
These methods often involve complex formulations and high computational costs.
%
%
In contrast, \name{} is a conceptually simple and computationally efficient first-order method.

The line of works closest to ours is safety layer~\citep{Dalal2018, pham2018optlayer, Cheng2019, Li2021SafeRL}.
Although they also transform unsafe actions into safe ones, their safety layers assume simple safety models which can be solved in closed forms or by inner-level optimization (\eg\, quadratic programming~\citep{Amos2017}).
We extend this idea to more general safe RL scenarios without assuming a particular safety model.
Our SE can be also treated as a teacher under the teacher-student framework, where UM is the student whose actions are corrected by SE.
%
%
However, existing works~\citep{turchetta2020safe,langlois2021rl} under this framework more or less rely on various heuristics and task-specific rules, or have access to the simulator's internal state.
In complex realistic environments, such assumptions are invalidated.

Recovery RL~\citep{Thananjeyan2021} also decomposes policy learning across two policies. 
It makes a \emph{hard} switch between a task policy and a recovery policy by comparing a pre-trained constraint critic with a manual threshold. 
It requires an offline dataset of demonstrations to learn the constraint critic and fix it during online exploration.
This offline training requirement and the hard switch scheme reduce the flexibility of Recovery RL.
DESTA~\citep{Mguni2021} fully decouples utility from safety by introducing two separate agents. 
The safety agent can decide at which states it takes control of the system while the task agent can only maximize utility at the remaining states.
There is hardly any communication or cooperation between the two agents.
In contrast, \name{} lets UM and SE cooperate with a ``propose-and-edit'' strategy, instead of switching between them.
\citet{Flet-Berliac2022} also adopts a two-policy design. 
However, their two policies are adversarial. 
More specifically, one of their policies tries to intentionally maximize risks by behaving unsafely, to shrink the feasibility region of the agent’s value function. 
Their overall policy will bias towards being conservative. 
Our SE minimizes risks, and UM and SE always cooperate to resolve the conflict between utility and safety.

\vspace{-2ex}
\section{Limitations and Conclusions}
\label{sec:conclusion}
\vspace{-1ex}
While \name{} is able to maintain an extremely low constraint violation rate in expectation, oscillation does happen occasionally on the training curves around the target threshold (Figure~\ref{fig:safety_gym_curves}).
In general, if no oracle safety model can be accessed by the agent, avoiding this oscillation issue would be very difficult.
On one hand, our tasks always generate new random environment layouts for new episodes, and in theory the agent has to generalize its safety knowledge (\eg\ one obstacle being dangerous at this location usually implies it being also dangerous even when the background changes).
On the other hand, neural networks have the notorious issue of catastrophic forgetting, meaning that while the agent learns some new safety knowledge, it might forget that already learned.
Finally, from a practical point of view, as the violation target reaches 0, it takes an exponential number of samples to evaluate if the agent truly can meet the target.
These challenging issues are currently ignored by \name{} and would be interesting future directions.

In summary, we have introduced \name{}, for general safe RL scenarios, that makes a safety editor policy transform preliminary actions output by a utility maximizer policy into safe actions.
In problems where a utility maximizing action is often safe already, this architecture leads to faster policy learning.
We train this two-policy framework with simple primal-dual optimization, resulting in an efficient first-order approach.
On 14 safe RL tasks with very harsh constraint violation rates, \name{} achieves an outstanding overall SWU score.
%
%
We hope that \name{} can serve as a preliminary step towards safe RL with harsh constraint budgets.
%
%

\bibliography{main}

\begin{thebibliography}{}

\bibitem[Achiam et~al., 2017]{achiam17a}
Achiam, J., Held, D., Tamar, A., and Abbeel, P. (2017).
\newblock Constrained policy optimization.
\newblock In {\em ICML}.

\bibitem[Amos and Kolter, 2017]{Amos2017}
Amos, B. and Kolter, J.~Z. (2017).
\newblock Optnet: Differentiable optimization as a layer in neural networks.
\newblock In {\em ICML}.

\bibitem[As et~al., 2022]{as2022constrained}
As, Y., Usmanova, I., Curi, S., and Krause, A. (2022).
\newblock Constrained policy optimization via bayesian world models.
\newblock In {\em International Conference on Learning Representations}.

\bibitem[Berkenkamp et~al., 2017]{berkenkamp2017safe}
Berkenkamp, F., Turchetta, M., Schoellig, A.~P., and Krause, A. (2017).
\newblock Safe model-based reinforcement learning with stability guarantees.
\newblock In {\em NeurIPS}.

\bibitem[Berner et~al., 2019]{OpenAI2019Five}
Berner, C., Brockman, G., Chan, B., Cheung, V., Debiak, P., Dennison, C.,
  Farhi, D., Fischer, Q., Hashme, S., Hesse, C., J{\'{o}}zefowicz, R., Gray,
  S., Olsson, C., Pachocki, J., Petrov, M., de~Oliveira~Pinto, H.~P., Raiman,
  J., Salimans, T., Schlatter, J., Schneider, J., Sidor, S., Sutskever, I.,
  Tang, J., Wolski, F., and Zhang, S. (2019).
\newblock Dota 2 with large scale deep reinforcement learning.
\newblock {\em arXiv}.

\bibitem[Bertsekas, 1999]{Bertsekas1999}
Bertsekas, D. (1999).
\newblock {\em Nonlinear Programming}.
\newblock Athena Scientific.

\bibitem[Bhatnagar and Lakshmanan, 2012]{Bhatnagar2012}
Bhatnagar, S. and Lakshmanan, K. (2012).
\newblock An online actor–critic algorithm with function approximation for
  constrained markov decision processes.
\newblock {\em Journal of Optimization Theory and Applications}, 153.

\bibitem[Bohez et~al., 2019]{bohez2019value}
Bohez, S., Abdolmaleki, A., Neunert, M., Buchli, J., Heess, N., and Hadsell, R.
  (2019).
\newblock Value constrained model-free continuous control.
\newblock {\em arXiv}.

\bibitem[Brockman et~al., 2016]{Brockman2016}
Brockman, G., Cheung, V., Pettersson, L., Schneider, J., Schulman, J., Tang,
  J., and Zaremba, W. (2016).
\newblock Openai gym.

\bibitem[Cheng et~al., 2019]{Cheng2019}
Cheng, R., Orosz, G., Murray, R.~M., and Burdick, J.~W. (2019).
\newblock End-to-end safe reinforcement learning through barrier functions for
  safety-critical continuous control tasks.
\newblock In {\em AAAI}.

\bibitem[Chou et~al., 2017]{chou2017improving}
Chou, P.-W., Maturana, D., and Scherer, S. (2017).
\newblock Improving stochastic policy gradients in continuous control with deep
  reinforcement learning using the beta distribution.
\newblock In {\em ICML}, pages 834--843.

\bibitem[Chow et~al., 2018]{chow2018lyapunovbased}
Chow, Y., Nachum, O., Duenez-Guzman, E., and Ghavamzadeh, M. (2018).
\newblock A lyapunov-based approach to safe reinforcement learning.
\newblock In {\em NeurIPS}.

\bibitem[Chow et~al., 2019]{chow2019lyapunovbased}
Chow, Y., Nachum, O., Faust, A., Duenez-Guzman, E., and Ghavamzadeh, M. (2019).
\newblock Lyapunov-based safe policy optimization for continuous control.
\newblock {\em arXiv}.

\bibitem[Dalal et~al., 2018]{Dalal2018}
Dalal, G., Dvijotham, K., Vecer{\'{\i}}k, M., Hester, T., Paduraru, C., and
  Tassa, Y. (2018).
\newblock Safe exploration in continuous action spaces.
\newblock {\em CoRR}.

\bibitem[Flet{-}Berliac and Basu, 2022]{Flet-Berliac2022}
Flet{-}Berliac, Y. and Basu, D. (2022).
\newblock {SAAC:} safe reinforcement learning as an adversarial game of
  actor-critics.
\newblock In {\em Conference on Reinforcement Learning and Decision Making}.

\bibitem[Haarnoja et~al., 2018]{Haarnoja2018}
Haarnoja, T., Zhou, A., Hartikainen, K., Tucker, G., Ha, S., Tan, J., Kumar,
  V., Zhu, H., Gupta, A., Abbeel, P., and Levine, S. (2018).
\newblock Soft actor-critic algorithms and applications.
\newblock {\em arXiv}, abs/1812.05905.

\bibitem[Kim et~al., 2004]{Kim2003}
Kim, H., Jordan, M., Sastry, S., and Ng, A. (2004).
\newblock Autonomous helicopter flight via reinforcement learning.
\newblock In {\em Advances in Neural Information Processing Systems},
  volume~16.

\bibitem[Langlois and Everitt, 2021]{langlois2021rl}
Langlois, E.~D. and Everitt, T. (2021).
\newblock How rl agents behave when their actions are modified.
\newblock In {\em AAAI}.

\bibitem[Levine et~al., 2016]{Levine2016}
Levine, S., Pastor, P., Krizhevsky, A., and Quillen, D. (2016).
\newblock Learning hand-eye coordination for robotic grasping with deep
  learning and large-scale data collection.
\newblock {\em arXiv}.

\bibitem[Li et~al., 2021]{Li2021SafeRL}
Li, Y., Li, N., Tseng, H.~E., Girard, A.~R., Filev, D., and Kolmanovsky, I.~V.
  (2021).
\newblock Safe reinforcement learning using robust action governor.
\newblock In {\em L4DC}.

\bibitem[Likhosherstov et~al., 2021]{likhosherstov2021debiasing}
Likhosherstov, V., Song, X., Choromanski, K., Davis, J., and Weller, A. (2021).
\newblock Debiasing a first-order heuristic for approximate bi-level
  optimization.
\newblock In {\em ICML}.

\bibitem[Liu et~al., 2022]{Liu2022}
Liu, Z., Cen, Z., Isenbaev, V., Liu, W., Wu, Z., Li, B., and Zhao, D. (2022).
\newblock Constrained variational policy optimization for safe reinforcement
  learning.
\newblock In {\em ICML}.

\bibitem[Luo and Ma, 2021]{Luo2021}
Luo, Y. and Ma, T. (2021).
\newblock Learning barrier certificates: Towards safe reinforcement learning
  with zero training-time violations.
\newblock In {\em NeurIPS}.

\bibitem[Mguni et~al., 2021]{Mguni2021}
Mguni, D., Jennings, J., Jafferjee, T., Sootla, A., Yang, Y., Yu, C., Islam,
  U., Wang, Z., and Wang, J. (2021).
\newblock {DESTA:} {A} framework for safe reinforcement learning with markov
  games of intervention.
\newblock {\em arXiv}.

\bibitem[Miret et~al., 2020]{miret2020safety}
Miret, S., Majumdar, S., and Wainwright, C. (2020).
\newblock Safety aware reinforcement learning (sarl).
\newblock {\em arXiv}.

\bibitem[Moffaert and Now{{\'e}}, 2014]{moffaert2015}
Moffaert, K.~V. and Now{{\'e}}, A. (2014).
\newblock Multi-objective reinforcement learning using sets of pareto
  dominating policies.
\newblock {\em JAIR}.

\bibitem[OpenAI et~al., 2019]{OpenAI2019}
OpenAI, Akkaya, I., Andrychowicz, M., Chociej, M., Litwin, M., McGrew, B.,
  Petron, A., Paino, A., Plappert, M., Powell, G., Ribas, R., Schneider, J.,
  Tezak, N., Tworek, J., Welinder, P., Weng, L., Yuan, Q., Zaremba, W., and
  Zhang, L. (2019).
\newblock Solving rubik's cube with a robot hand.
\newblock {\em arXiv}.

\bibitem[Pham et~al., 2018]{pham2018optlayer}
Pham, T.-H., Magistris, G.~D., and Tachibana, R. (2018).
\newblock Optlayer - practical constrained optimization for deep reinforcement
  learning in the real world.
\newblock In {\em ICRA}.

\bibitem[Qin et~al., 2021]{qin2021density}
Qin, Z., Chen, Y., and Fan, C. (2021).
\newblock Density constrained reinforcement learning.
\newblock In {\em ICML}.

\bibitem[Ray et~al., 2019]{Ray2019}
Ray, A., Achiam, J., and Amodei, D. (2019).
\newblock {Benchmarking Safe Exploration in Deep Reinforcement Learning}.

\bibitem[Roijers et~al., 2013]{Roijers2013}
Roijers, D.~M., Vamplew, P., Whiteson, S., and Dazeley, R. (2013).
\newblock A survey of multi-objective sequential decision-making.
\newblock {\em JAIR}.

\bibitem[Schulman et~al., 2017]{schulman2017proximal}
Schulman, J., Wolski, F., Dhariwal, P., Radford, A., and Klimov, O. (2017).
\newblock Proximal policy optimization algorithms.
\newblock {\em arXiv}.

\bibitem[Silver et~al., 2016]{Silver2016}
Silver, D., Huang, A., Maddison, C.~J., Guez, A., Sifre, L., van~den Driessche,
  G., Schrittwieser, J., Antonoglou, I., Panneershelvam, V., Lanctot, M.,
  Dieleman, S., Grewe, D., Nham, J., Kalchbrenner, N., Sutskever, I.,
  Lillicrap, T., Leach, M., Kavukcuoglu, K., Graepel, T., and Hassabis, D.
  (2016).
\newblock Mastering the game of {Go} with deep neural networks and tree search.
\newblock {\em Nature}, 529(7587):484--489.

\bibitem[Stooke et~al., 2020]{stooke2020responsive}
Stooke, A., Achiam, J., and Abbeel, P. (2020).
\newblock Responsive safety in reinforcement learning by pid lagrangian
  methods.
\newblock In {\em ICML}.

\bibitem[Tassa et~al., 2020]{tassa2020dmcontrol}
Tassa, Y., Tunyasuvunakool, S., Muldal, A., Doron, Y., Liu, S., Bohez, S.,
  Merel, J., Erez, T., Lillicrap, T., and Heess, N. (2020).
\newblock dm\_control: Software and tasks for continuous control.

\bibitem[Tessler et~al., 2019]{Tessler2019}
Tessler, C., Mankowitz, D.~J., and Mannor, S. (2019).
\newblock Reward constrained policy optimization.
\newblock In {\em ICLR}.

\bibitem[Thananjeyan et~al., 2021]{Thananjeyan2021}
Thananjeyan, B., Balakrishna, A., Nair, S., Luo, M., Srinivasan, K., Hwang, M.,
  Gonzalez, J.~E., Ibarz, J., Finn, C., and Goldberg, K. (2021).
\newblock Recovery {RL:} safe reinforcement learning with learned recovery
  zones.
\newblock In {\em ICRA}.

\bibitem[Thomas et~al., 2021]{thomas2021safe}
Thomas, G., Luo, Y., and Ma, T. (2021).
\newblock Safe reinforcement learning by imagining the near future.
\newblock In {\em NeurIPS}.

\bibitem[Turchetta et~al., 2020]{turchetta2020safe}
Turchetta, M., Kolobov, A., Shah, S., Krause, A., and Agarwal, A. (2020).
\newblock Safe reinforcement learning via curriculum induction.
\newblock In {\em NeurIPS}.

\bibitem[Van~Moffaert et~al., 2013]{Moffaert2013}
Van~Moffaert, K., Drugan, M.~M., and Nowé, A. (2013).
\newblock Scalarized multi-objective reinforcement learning: Novel design
  techniques.
\newblock In {\em 2013 IEEE Symposium on Adaptive Dynamic Programming and
  Reinforcement Learning (ADPRL)}.

\bibitem[Vinyals et~al., 2019]{Vinyals2019GrandmasterLI}
Vinyals, O., Babuschkin, I., Czarnecki, W.~M., Mathieu, M., Dudzik, A., Chung,
  J., Choi, D.~H., Powell, R., Ewalds, T., Georgiev, P., Oh, J., Horgan, D.,
  Kroiss, M., Danihelka, I., Huang, A., Sifre, L., Cai, T., Agapiou, J.~P.,
  Jaderberg, M., Vezhnevets, A.~S., Leblond, R., Pohlen, T., Dalibard, V.,
  Budden, D., Sulsky, Y., Molloy, J., Paine, T.~L., Gulcehre, C., Wang, Z.,
  Pfaff, T., Wu, Y., Ring, R., Yogatama, D., W{\"u}nsch, D., McKinney, K.,
  Smith, O., Schaul, T., Lillicrap, T.~P., Kavukcuoglu, K., Hassabis, D., Apps,
  C., and Silver, D. (2019).
\newblock Grandmaster level in starcraft ii using multi-agent reinforcement
  learning.
\newblock {\em Nature}, pages 1--5.

\bibitem[Yang et~al., 2020]{Yang2020Projection-Based}
Yang, T.-Y., Rosca, J., Narasimhan, K., and Ramadge, P.~J. (2020).
\newblock Projection-based constrained policy optimization.
\newblock In {\em ICLR}.

\bibitem[Zhang et~al., 2020]{zhang2020order}
Zhang, Y., Vuong, Q., and Ross, K.~W. (2020).
\newblock First order constrained optimization in policy space.
\newblock In {\em NeurIPS}.

\bibitem[Zhao et~al., 2020]{Zhao2020}
Zhao, W., Queralta, J.~P., and Westerlund, T. (2020).
\newblock Sim-to-real transfer in deep reinforcement learning for robotics: a
  survey.
\newblock In {\em 2020 IEEE Symposium Series on Computational Intelligence
  (SSCI)}.

\end{thebibliography}
\bibliographystyle{apalike}

\section*{NeurIPS Checklist}

\begin{enumerate}

\item For all authors...
\begin{enumerate}
  \item Do the main claims made in the abstract and introduction accurately reflect the paper's contributions and scope?
    \answerYes{}
  \item Did you describe the limitations of your work?
    \answerYes{Section~\ref{sec:conclusion}}
  \item Did you discuss any potential negative societal impacts of your work?
    \answerNA{The paper is about RL safety which strives to provide positive societal impacts.}
  \item Have you read the ethics review guidelines and ensured that your paper conforms to them?
    \answerYes{}
\end{enumerate}

\item If you are including theoretical results...
\begin{enumerate}
  \item Did you state the full set of assumptions of all theoretical results?
    \answerNA{No theoretical results.}
        \item Did you include complete proofs of all theoretical results?
    \answerNA{}
\end{enumerate}

\item If you ran experiments...
\begin{enumerate}
  \item Did you include the code, data, and instructions needed to reproduce the main experimental results (either in the supplemental material or as a URL)?
    \answerYes{}
  \item Did you specify all the training details (e.g., data splits, hyperparameters, how they were chosen)?
    \answerYes{Appendix~\ref{app:hyperparameters}}
        \item Did you report error bars (e.g., with respect to the random seed after running experiments multiple times)?
    \answerYes{For example, Figure~\ref{fig:safety_gym_curves} and Figure~\ref{fig:safe_car_racing_curves}}
        \item Did you include the total amount of compute and the type of resources used (e.g., type of GPUs, internal cluster, or cloud provider)?
    \answerYes{Appendix~\ref{app:hyperparameters}}
\end{enumerate}

\item If you are using existing assets (e.g., code, data, models) or curating/releasing new assets...
\begin{enumerate}
  \item If your work uses existing assets, did you cite the creators?
    \answerYes{See Section~\ref{sec:experiments}, Appendix~\ref{app:safety_gym}, and Appendix~\ref{app:safe_racing}.}
  \item Did you mention the license of the assets?
    \answerNA{}
  \item Did you include any new assets either in the supplemental material or as a URL?
    \answerYes{}
  \item Did you discuss whether and how consent was obtained from people whose data you're using/curating?
    \answerNA{The data are open-sourced.}
  \item Did you discuss whether the data you are using/curating contains personally identifiable information or offensive content?
    \answerNA{We only use virtual simulators.}
\end{enumerate}

\item If you used crowdsourcing or conducted research with human subjects...
\begin{enumerate}
  \item Did you include the full text of instructions given to participants and screenshots, if applicable?
    \answerNA{}
  \item Did you describe any potential participant risks, with links to Institutional Review Board (IRB) approvals, if applicable?
    \answerNA{}
  \item Did you include the estimated hourly wage paid to participants and the total amount spent on participant compensation?
    \answerNA{}
\end{enumerate}

\end{enumerate}


\clearpage
\appendix

\begin{figure}[!t]
\centering
\resizebox{\columnwidth}{!}{
\begin{tabular}{@{}c|c@{}}
    \begin{tabular}{@{}c@{}c@{}c@{}}
        \includegraphics[width=0.4\columnwidth]{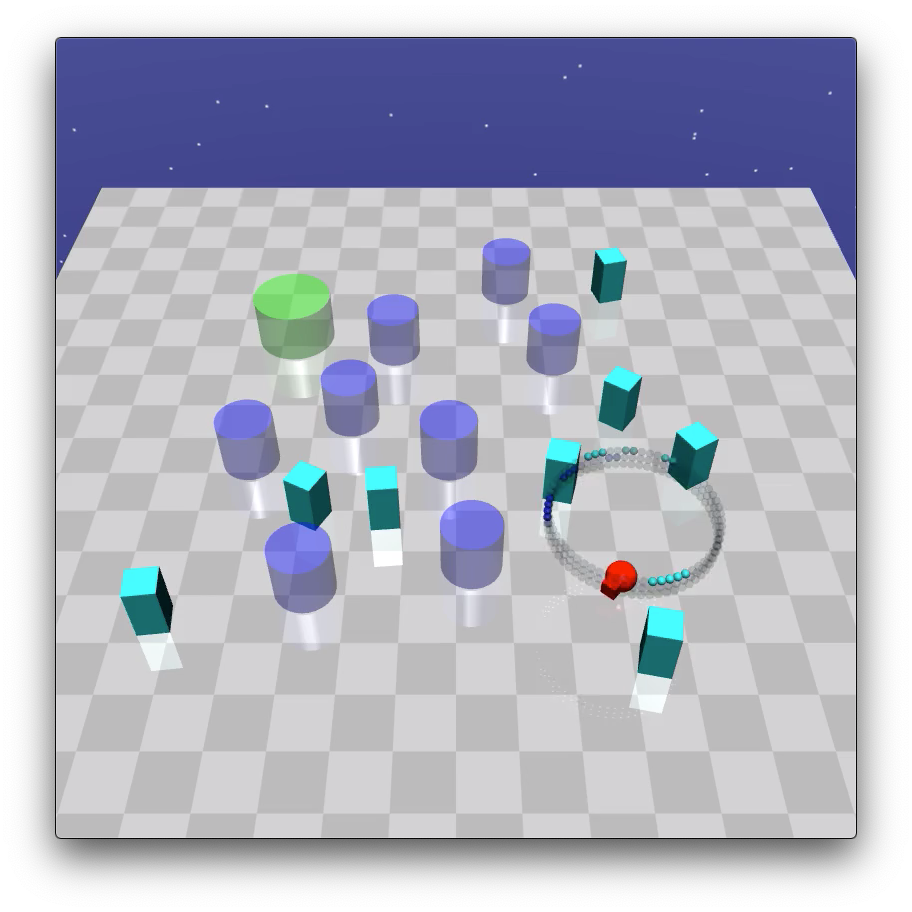}
        &\includegraphics[width=0.4\columnwidth]{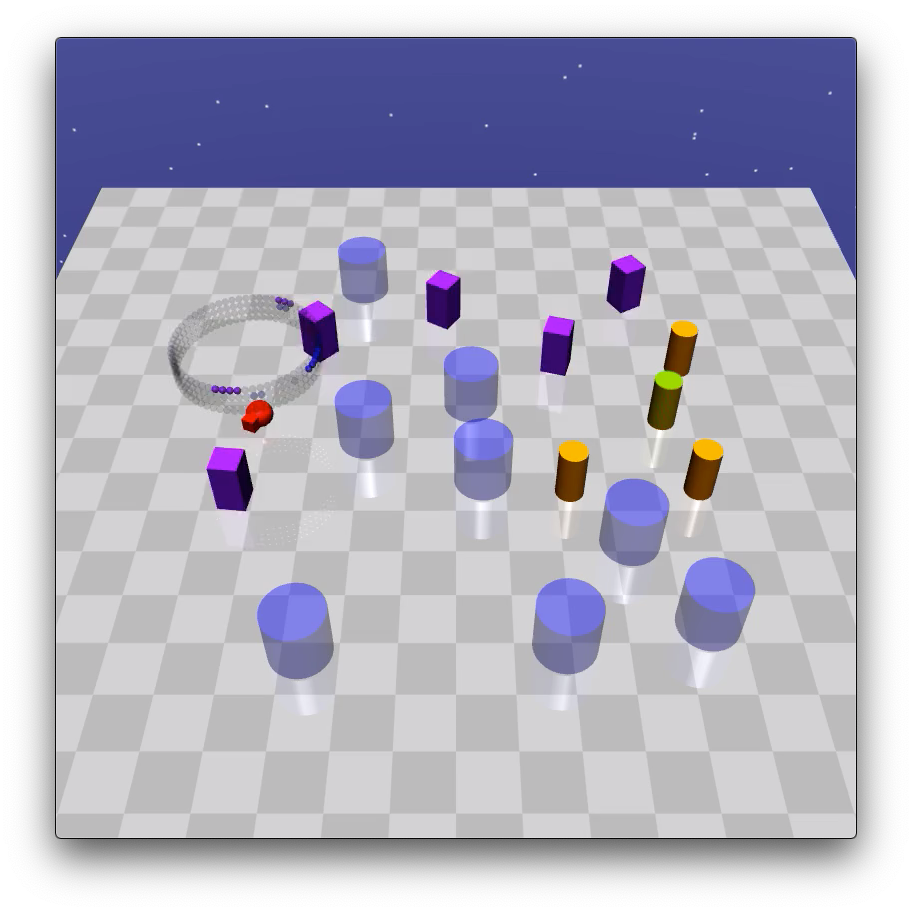}
        &\includegraphics[width=0.4\columnwidth]{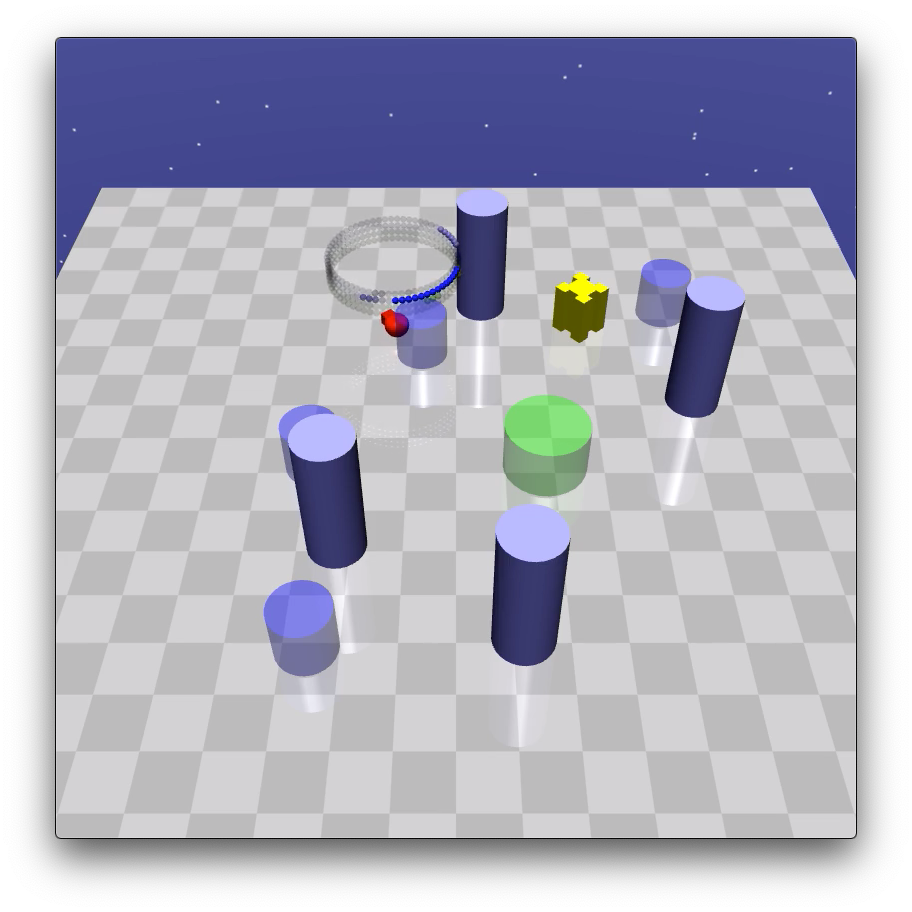}\\
        \textsc{PointGoal2} & \textsc{PointButton2} & \textsc{PointPush2}\\\\
    \end{tabular}&
    \begin{tabular}{@{}c@{}}
        \includegraphics[width=0.12\columnwidth]{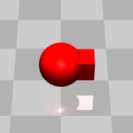}\\
        \includegraphics[width=0.12\columnwidth]{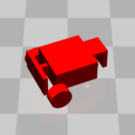}\\
    \end{tabular}\\
\end{tabular}
}
\caption{Left: The three level-2 Safety Gym tasks with the \textsc{Point} robot.
Level-1 tasks have less crowded maps.
\textcolor{LimeGreen}{Green} zones: goal locations;
\textcolor{Violet}{blue} zones: hazard zones;
\textcolor{Cyan}{cyan} cubes: vases;
\textcolor{Purple}{purple} cubes: gremlins;
\textcolor{BurntOrange}{orange} cylinders: buttons;
\textcolor{Violet}{blue} cylinders: pillars.
Hazard zones are penetrable areas. 
Vases are lightweight and can be moved by the robot.
Unlike other static obstacles, gremlins have circular movements. 
Right: the \textsc{Point} and \textsc{Car} robots. 
}
\label{fig:safety_gym_tasks}
\end{figure}

\section{Task details}
\subsection{Safety Gym}
\label{app:safety_gym}
In Safety Gym~\citep{Ray2019} environments, a robot with lidar sensors navigates through cluttered environments to achieve tasks.
There are three types of tasks (Figure~\ref{fig:safety_gym_tasks}) for a robot:
\begin{compactenum}[1)]
    \item \textsc{Goal}: reaching a goal location while avoiding hazard zones and vases.
    \item \textsc{Button}: hitting one goal button out of several buttons while avoiding gremlins and hazard zones.
    \item \textsc{Push}: pushing a box to a goal location while avoiding pillars and hazard zones.
\end{compactenum}
Each task has two levels, where level 2 has more obstacles and a larger map size than level 1.
In total there are $3\times 2=6$ tasks for a robot.
We use the \textsc{Point} and \textsc{Car} robots in our experiments.

\begin{table}[!t]
\centering
\resizebox{0.4\textwidth}{!}{
    \begin{tabular}{c|ccc}
                        & \textsc{Goal} & \textsc{Push} & \textsc{Button} \\
        \hline 
        \textsc{Point}  
        & 204
        & 268
        & 268\\
        \textsc{Car} 
        & 216
        & 280
        & 280\\
    \end{tabular}}
    \caption{The observation dimensions of our custom Safety Gym tasks.
    For each combination, level 1 and 2 have the same observation space.
    All action spaces are $[-1,1]^2$.
    }
    \label{tab:safety_gym_spaces}
\end{table}

We customized the environment so that the robot has a natural lidar of 64 bins.
The natural lidar contains more information of object shapes in the environment than the default pseudo lidar. 
We found that rich shape information is necessary for the agent to achieve a harsh constraint threshold.
A separate lidar vector of length 64 is produced for each obstacle type or goal.
All lidar vectors and the robot status vector (\eg\, acceleration, velocity, rotations) are concatenated together to produce a flattened observation vector.
A summary of the observation dimensions is in Table~\ref{tab:safety_gym_spaces}.
Whenever an obstacle is in contact with the robot, a constraint reward of $-1$ is given.
The utility reward is calculated as the decrement of the distance between the robot (\textsc{Goal} and \textsc{Button}) or box (\textsc{Push}) and the goal at every step.
An episode terminates when the goal is achieved, or after $1000$ time steps.
We define a success as achieving the goal before timeout.
The map layout is \emph{randomized} at the beginning of each episode.
We emphasize that the agent has no prior knowledge of which states are unsafe, thus path planning with known obstacles does not apply here.
\ifdefined\isaccepted
Our customized Safety Gym is available at \url{https://github.com/hnyu/safety-gym}.
\else\fi

\begin{figure}[!t]
\centering
\resizebox{0.7\columnwidth}{!}{
\begin{tabular}{@{}cc@{}}
\includegraphics[width=0.45\columnwidth]{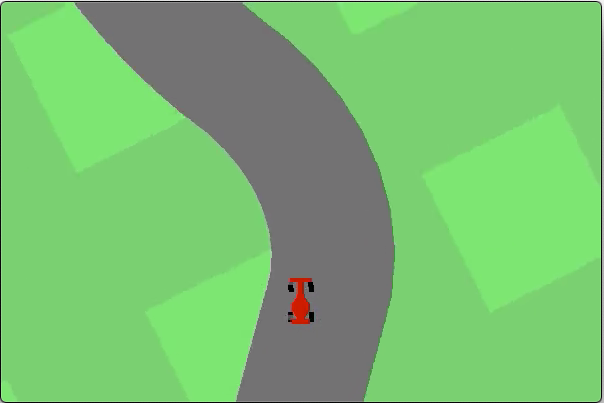}
&\includegraphics[width=0.45\columnwidth]{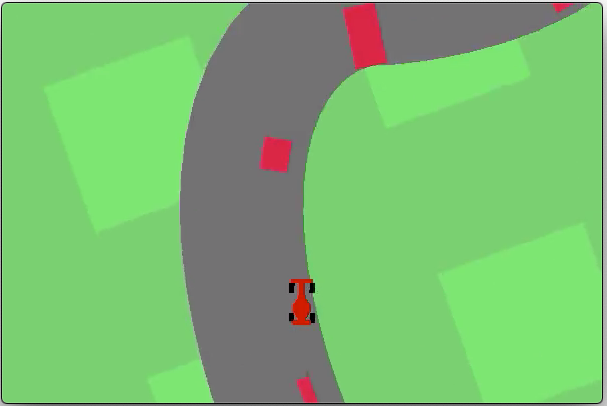}\\
\end{tabular}}
\caption{The \textsc{SafeRacing} (left) and \textsc{SafeRacingObstacle} (right) tasks.
In the former task, the car needs to keep itself within the track to avoid penalties.
In the latter task, the car gets a collision cost once hitting an obstacle (red blocks).
However, it can drive outside of the track without being penalized.
For each episode, the track layout and obstacles are randomly generated.
}
\label{fig:safe_car_racing}
\end{figure}

\subsection{Safe Racing}
\label{app:safe_racing}
The agent's observation includes a bird's-eye view image ($96\times96$) and a car status vector (length 11) consisting of ABS sensor, wheel angles, speed, angular velocity, and the remaining tile portion.
The action space is $[-1,1]^3$.
\ifdefined\isaccepted
Based on the code~\url{https://github.com/NotAnyMike/gym/blob/master/gym/envs/box2d/car_racing.py}, we modify the original unconstrained car racing task~\citep{Brockman2016} to add obstacles on the track.
\else 
We modify the original unconstrained car racing task~\citep{Brockman2016} to add obstacles on the track.
\fi
A bird's-eye view of the two safe racing tasks is illustrated in Figure~\ref{fig:safe_car_racing}.
We set the obstacle density ($\frac{\text{\#Obstacles}}{\text{\#TrackTiles}}$) to $10\%$ for \textsc{SafeRacingObstacle}.

\section{Training with 10M Environment Steps}
In Figure~\ref{fig:safety_gym_curves}, one might wonder if \name{} only improves sample efficiency of the Lagrangian SAC, but doesn't really improve the final performance on the 12 Safety Gym tasks. 
To answer this question, we report here the SWU score comparison between SAC-actor2x-Lag and \name{} with 10M steps on Safety Gym. 
The SWU scores are calculated based on the performance of unconstrained SAC at 10M steps (and thus are not directly comparable to those in Table~\ref{tab:SWU}).
\begin{table}[!h]
    \centering
    \resizebox{\columnwidth}{!}{
    \begin{tabular}{@{}r|cccccccccccc|cc@{}}
         &\textsc{CP1}&\textsc{CG1}&\textsc{CB1}&\textsc{CP2}&\textsc{CG2}&\textsc{CB2}&\textsc{PP1}&\textsc{PG1}&\textsc{PB1}&\textsc{PP2}&\textsc{PG2}&\textsc{PB2}
         &Overall&Improvement\\
         \hline
        SAC-actor2x-Lag&0.91&0.84&\textbf{1.00}&0.82&0.84&0.80&1.02&1.00&0.74&0.75&1.00&1.00&\cellcolor{blue!15}0.89&\cellcolor{blue!15}17\%\\
        \name{}&1.01&\textbf{1.00}&0.85&\textbf{1.28}&\textbf{1.00}&\textbf{0.98}&0.94&1.00&0.97&\textbf{1.42}&1.00&1.00&\cellcolor{blue!15}\textbf{1.04}&\cellcolor{blue!15}-\\
    \end{tabular}
    }
    \caption{
    The SWU scores of SAC-actor2x-Lag and \name{} at 10M environment steps.
    Task name abbreviations: \textsc{CP} - \textsc{CarPush}, \textsc{CG} - \textsc{CarGoal}, \textsc{CB} - \textsc{CarButton}, 
    \textsc{PP} - \textsc{PointPush}, \textsc{PG} - \textsc{PointGoal}, and \textsc{PB} - \textsc{PointButton}.}
    \label{tab:SWU-10M}
\end{table}

We observe that both methods have saturated at 10M steps.
Training more steps somewhat decreases but not closes the gap between SAC-actor2x-Lag and \name{} regarding the final performance.

\section{Experiment on the Unmodified \textsc{PointGoal1}}
Since we have modified the Safety Gym environments to pursue a much (98\%) lower constraint violation threshold, one might be curious to see if \name{} also performs well on the original unmodified tasks.
As a representative experiment, we compare \name{} (averaged over 4 random seeds) with the results reported in \citet{Ray2019} and \citet{stooke2020responsive} on the unmodified \textsc{PointGoal1}.
We observe that all methods can satisfy the constraint threshold well; the difference residues in their utility performance. 
We list their (rough) utility scores at different environment steps below:
\begin{table}[!h]
    \centering
    \resizebox{0.9\columnwidth}{!}{
    \begin{tabular}{@{}r|cccc@{}}
        Steps&\citet{Ray2019} (PPO-Lag)&\citet{Ray2019}(TRPO-Lag)&\citet{stooke2020responsive}&\name{}\\
        \hline
        $2.5\times10^7$ & - & - & 26 & 29\\
        $1\times 10^7$ & 13 & 17 & 23 & 27\\
        $5\times 10^6$ & 14 & 16 & 22 & 24\\
    \end{tabular}
    }
    \caption{The utility performance on the original \textsc{PointGoal1}. All methods are able to satisfy the constraint threshold of $0.025$.
    }
    \label{tab:unmodified-pointgoal1}
\end{table}

It's unsurprising that \name{} did pretty well under such a much higher cost limit.
We also observe that without P-control, \name{}'s cost curve is similar to the ones of $K_P=0$ in \citet{stooke2020responsive}, which is expected: the initial cost was high and then quickly dropped to the limit. 
Our cost stabilized at about 3M steps while \citet{stooke2020responsive} stabilized at about 10M steps (with $K_I=1\times10^{-2}$).

\section{Ablation Study Results}
\label{app:ablation}
\begin{figure*}[!t]
\centering
\resizebox{\textwidth}{!}{
\begin{tabular}{@{}c@{}}
\includegraphics[width=\textwidth]{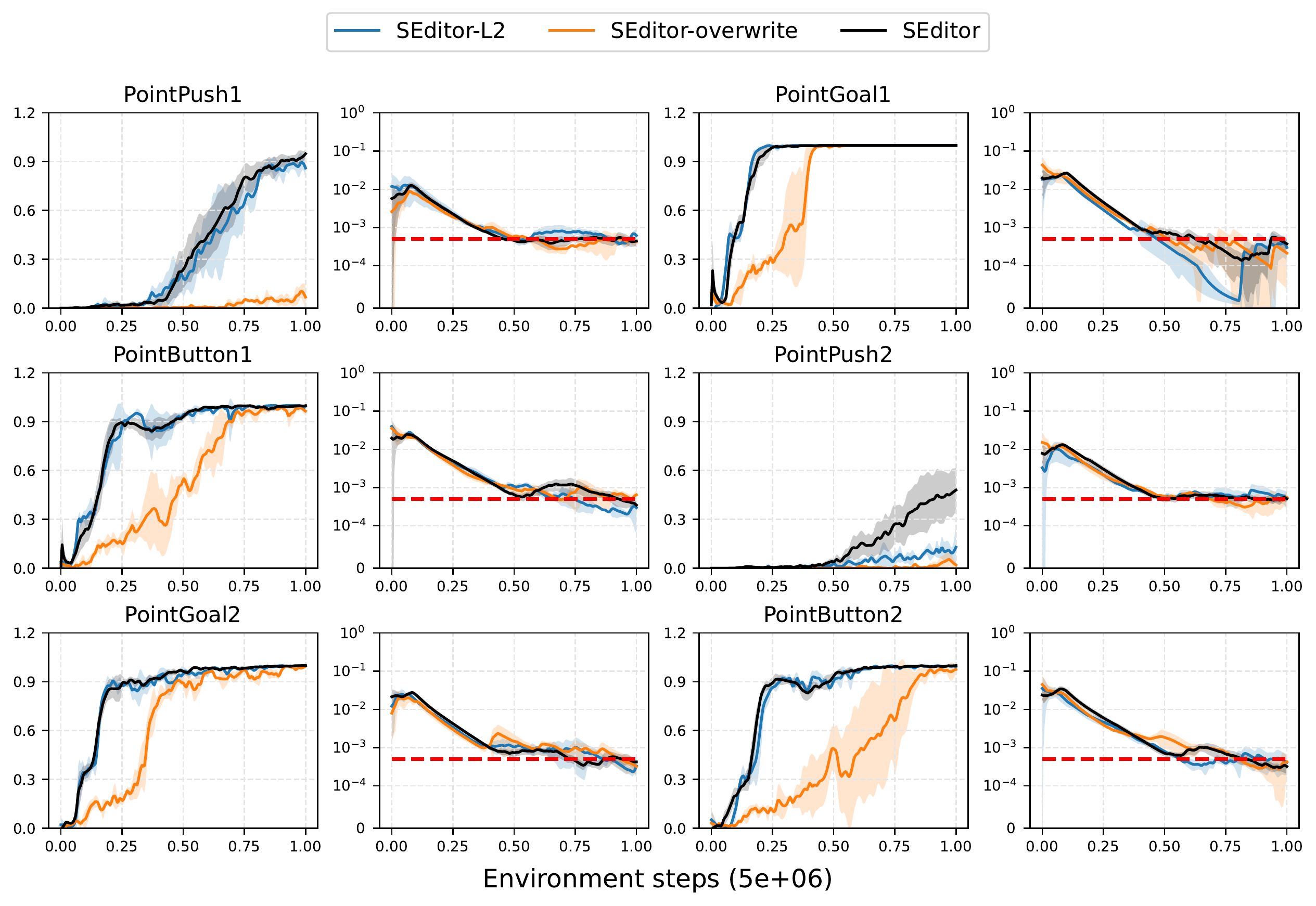}\\
\includegraphics[width=\textwidth]{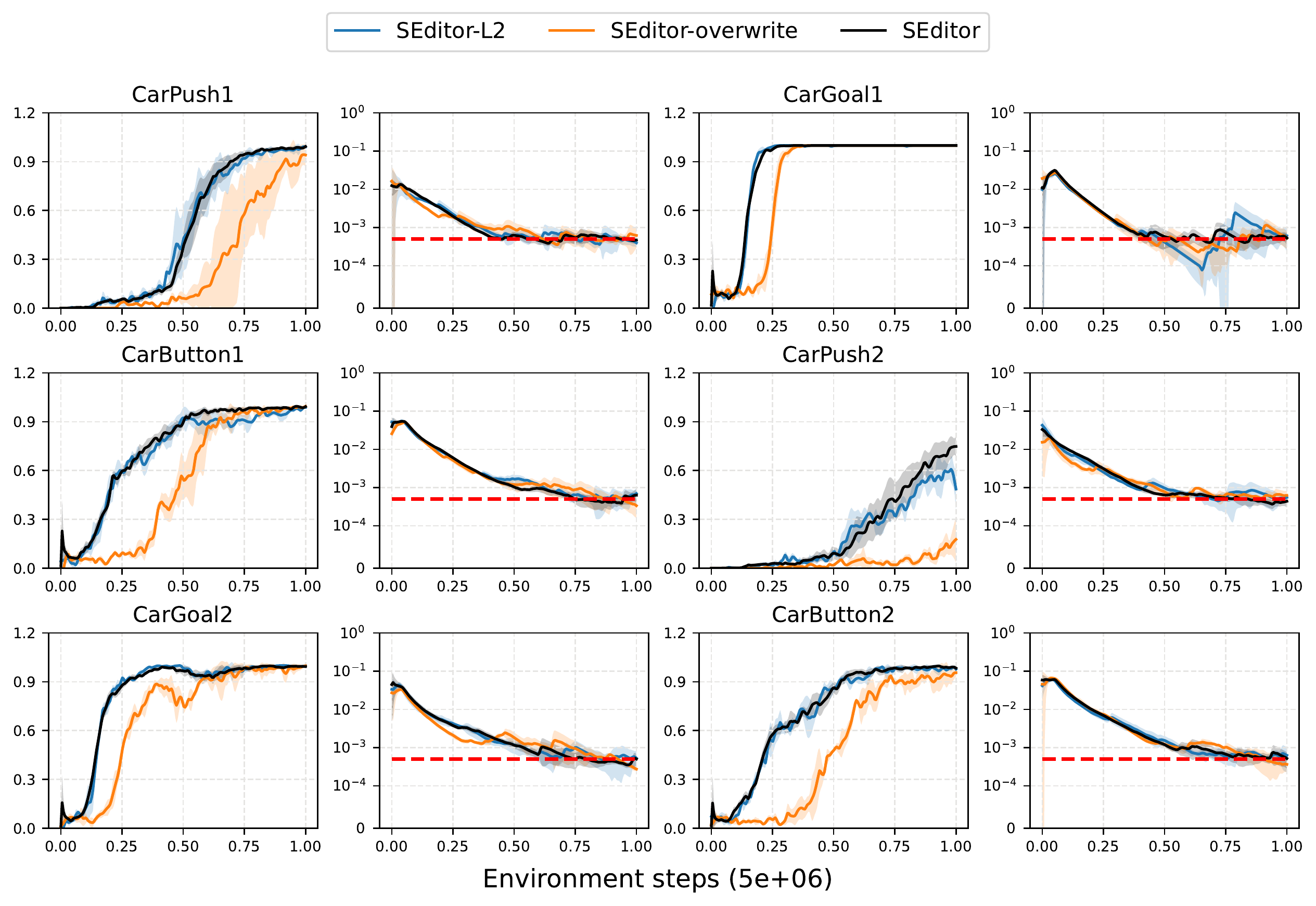}
\end{tabular}
}
\caption{
The ablation study results on the 12 Safety Gym tasks. 
Odd columns: $\uparrow$ success rate. Even columns: $\downarrow$ constraint violation rate (log scale). Red dashed horizontal lines: violation rate target $c=5\times10^{-4}$. Shaded areas: 95\% confidence interval (CI).
}
\label{fig:safety_gym_ablation}
\end{figure*}

\begin{figure*}[!t]
\centering
\resizebox{\textwidth}{!}{
\begin{tabular}{@{}c@{}c@{}}
\multicolumn{2}{c}{\includegraphics[width=0.5\textwidth]{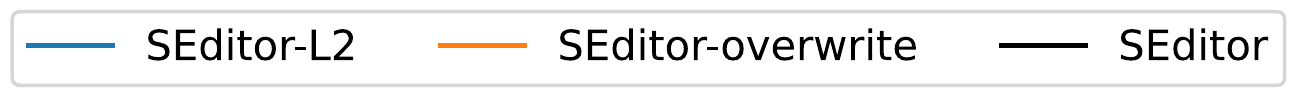}}\\
\includegraphics[width=0.5\textwidth]{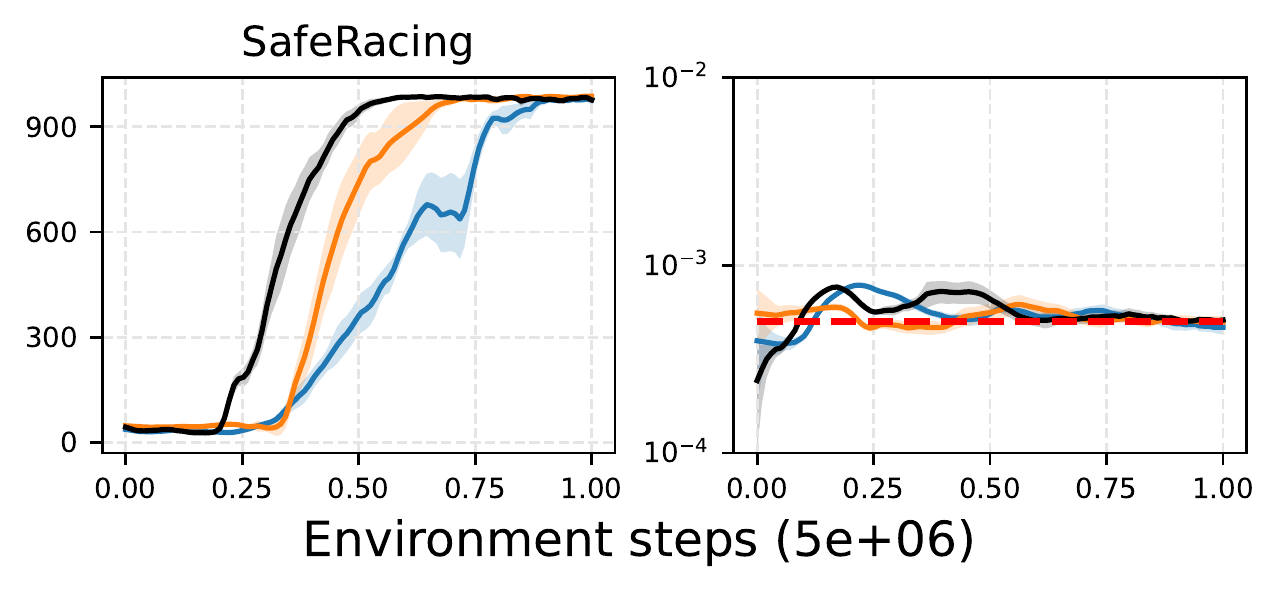}&
\includegraphics[width=0.5\textwidth]{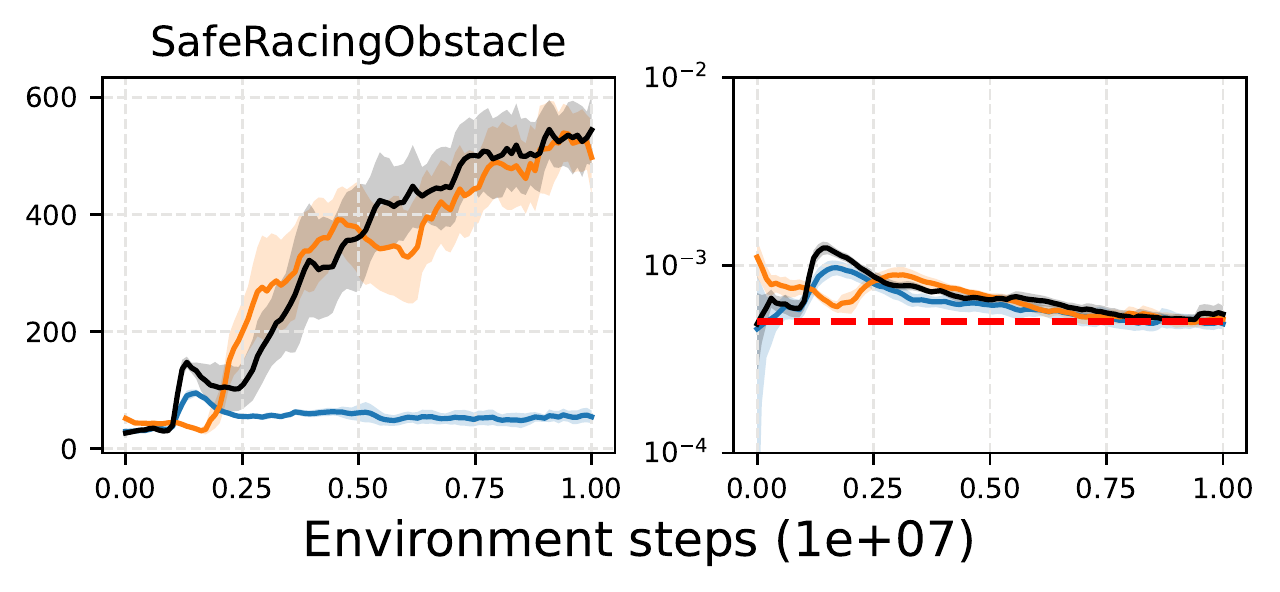}\\
\end{tabular}
}
\caption{The ablation study results on the safe racing tasks. 
Odd columns: $\uparrow$ undiscounted episode return. Even columns: $\downarrow$ constraint violation rate (log scale). Red dashed horizontal lines: violation rate target $c=5\times10^{-4}$. Shaded areas: 95\% confidence interval (CI).
}
\label{fig:safe_racing_ablation}
\end{figure*}

Figure~\ref{fig:safety_gym_ablation} and~\ref{fig:safe_racing_ablation} show the comparison results between \name{} and its two variants \name{}-L2 and \name{}-overwrite, as introduced in Section~\ref{sec:experiments}.
All three approaches share a common training setting except the changes to the action distance function $d(a,\hat{a})$ or the editing function $h(\hat{a},\Delta a)$.

\begin{figure*}[!t]
\centering
\resizebox{0.9\textwidth}{!}{
\begin{tabular}{@{}c@{}}
\includegraphics[width=\textwidth]{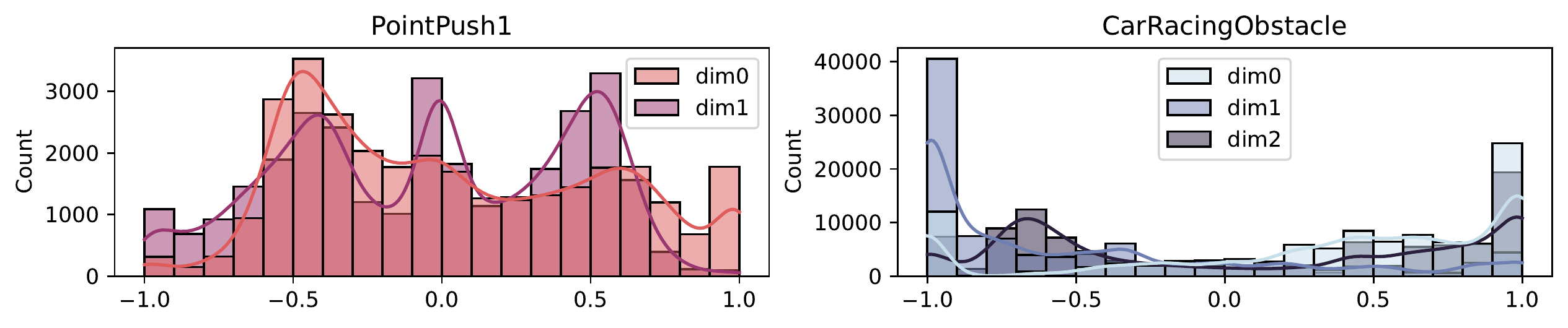}\\
\end{tabular}
}
\caption{The empirical distributions of $\Delta a$ over 100 episodes of \textsc{PointPush1} and \textsc{SafeRacingObstacle}, by evaluating trained models of \name{}.
Recall that their action dimensions are 2 and 3, respectively.
}
\label{fig:da_distribution}
\end{figure*}

We notice that Figure~\ref{fig:safety_gym_ablation} and \ref{fig:safe_racing_ablation} have opposite results, where \name{}-L2 is comparable to \name{} in the former while \name{}-overwrite is comparable to \name{} in the latter.
This suggests a hypothesis that the magnitude of $\Delta a$ output by SE is usually small in Safety Gym but larger in the safe racing tasks, 
because \name{}-overwrite removes the inductive bias of $\hat{a}$ being close to $h(\hat{a},\Delta a)$.
To verify this hypothesis, we record the output $\Delta a$ when evaluating the trained models of \name{} on two representative tasks \textsc{PointPush1} and \textsc{SafeRacingObstacle}.
For either task, we plot the empirical distribution of $\Delta a$ over $100$ episodes (each episode has $1000$ steps).
The plotted distributions are in Figure~\ref{fig:da_distribution}.
It is clear that on \textsc{PointPush1}, the population of $\Delta a$ is more centered towards $0$.
On \textsc{SafeRacingObstacle}, the population tends to distribute on the two extremes of $\pm 1$.
This somewhat explains why the L2 distance can be a good proxy for the utility Q closeness on Safety Gym but not on the safe racing tasks.

\begin{figure*}[hbt!]
\centering
\resizebox{\textwidth}{!}{
\begin{tabular}{@{}c@{}}
\includegraphics[width=\textwidth]{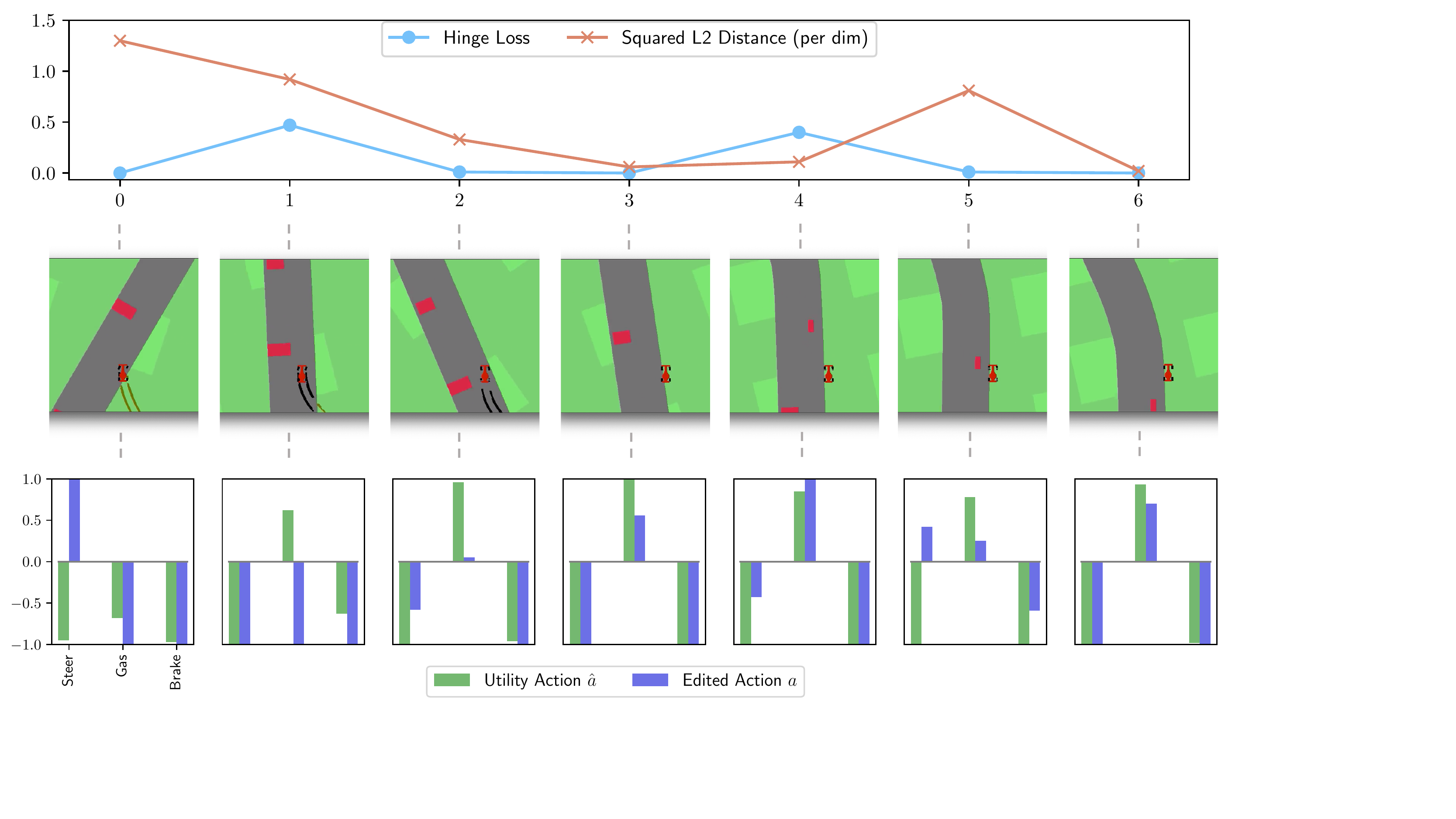}
\end{tabular}
}
\caption{Inference results during an example episode of \textsc{SafeRacingObstacle}, by evaluating a trained model of \name{}.
Top: The changing hinge loss of the utility state-action values of $\hat{a}$ and $a$, and their changing squared L2 distance (per dimension).
Note that the absolute magnitudes of the two quantities are not comparable.
Instead, only the relative trend within either curve is meaningful.
Middle: 7 key frames of the episode (zoom in for a better view).
Bottom: bar plots of the utility action $\hat{a}$ by UM and the edited action $a$ by SE.
Each plot corresponds to a key frame. 
}
\label{fig:episode}
\end{figure*}

\clearpage 

\section{Hinge Loss \vs\ L2 distance}
\label{app:hinge_vs_l2}

\begin{table}[!t]
\centering
\resizebox{0.6\columnwidth}{!}{
    \begin{tabular}{c|cc}
        Action dimension & $-1$ & $+1$ \\
        \hline
        $0$ & Steer left & Steer right\\
        $1$ & No acceleration & Full acceleration\\
        $2$ & No brake & Full brake\\
    \end{tabular}}
    \caption{The semantics of the action space of \textsc{SafeRacingObstacle}.
    Since the action space is continuous, numbers between the two extremes of $\pm 1$ represent a smooth transition.
    }
    \label{tab:action_semantics}
\end{table}

In Figure~\ref{fig:safe_racing_ablation}, \name{}-L2 is especially bad compared to \name{}, indicating that the L2 distance is not a good choice for measuring the difference between the utility action $\hat{a}$ and the edited action $a$, when we actually attempt to compare their state-action values.
For further analysis, we evaluate a trained \name{} model on \textsc{SafeRacingObstacle}, and inspect the following inference results during an episode:
\begin{compactenum}[a)]
    \item The action proposed by UM $\piu$, also known as the utility action $\hat{a}\in[-1,1]^3$;
    \item The edited action $a\in[-1,1]^3$ by SE $\pis$ as the output to the environment;
    \item The hinge loss of their utility state-action values (Eq.~\ref{eq:hinge_loss});
    \item The squared L2 distance (per dimension) of the two actions $\frac{1}{3}\Vert \hat{a} - a\Vert^2$
\end{compactenum}
We select 7 key frames of the episode and visualize their corresponding inference results in Figure~\ref{fig:episode}.
The semantics of the action space is listed in Table~\ref{tab:action_semantics}.

In the first frame, the car just gets back on track from outside and there is an obstacle in front of it.
The utility action $\hat{a}$ steers left while the edited action $a$ steers right due to safety concern. 
This causes their squared L2 distance to be quite large.
However, $Q(s,a;\theta)$ is no worse than $Q(s,\hat{a};\theta)$, and thus in this case $\pis$ of \name{} only needs to focus on maximizing the constraint reward, while $\pis$ of \name{}-L2 has to make compromises.
The second frame is where the hinge loss is positive because $\hat{a}$ commands acceleration while $a$ does not, resulting in a potential decrease of the utility return.
(The front tires of the car are already steered all the way to the right, thus both actions turn left.)
Overall, SE is more cautious and wants to slow down when passing the obstacle.
For frames 2 and 3, $\hat{a}$ and $a$ are similar, as the car is temporarily free from constraint violation.
Frame 4 is an example where a subtle difference in the L2 distance results in a large hinge loss.
The car is driving near the border of the track, and at any time it could go off-track and miss the next utility reward (a reward is given if the car touches a track tile).
Thus $\hat{a}$ turns all the way to the left to make sure that the off-track scenario will not happen.
However, because there is an obstacle in front, $a$ makes the steering less extreme.
Since the car is at the critical point of being on-track, even a small difference in steering results in a large hinge loss.
In comparison, the left-front and left-rear tires are already on the track in frame 5, and even though $a$ wants to turn right a little bit to avoid the obstacle, the utility return is not affected and the hinge loss is still zero. 
Frame 6 is an example where both the L2 distance and hinge loss are small.

\begin{figure*}[hbt!]
\centering
\resizebox{\textwidth}{!}{
\begin{tabular}{@{}c|c|c@{}}
\includegraphics[width=0.2\textwidth]{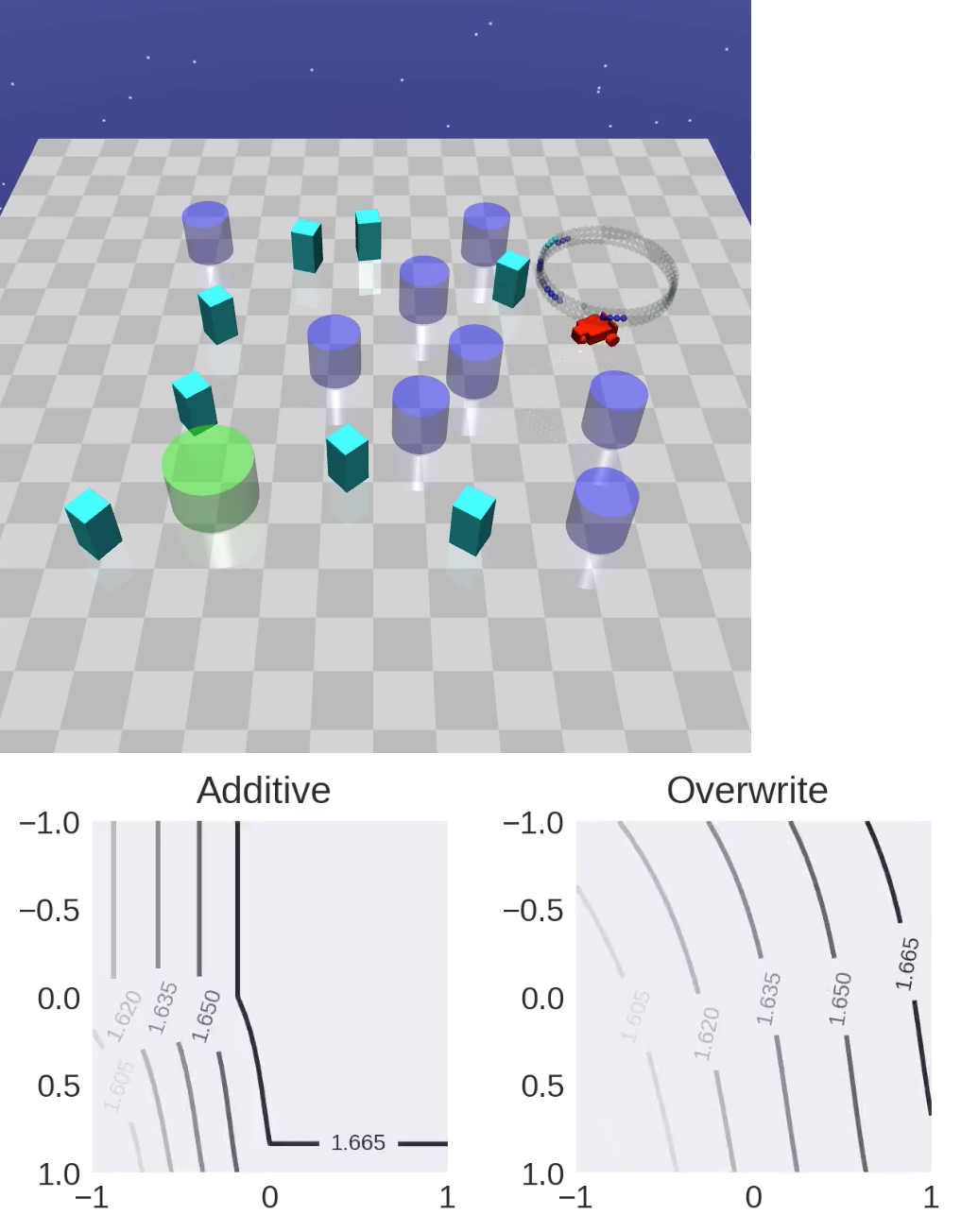}&
\includegraphics[width=0.2\textwidth]{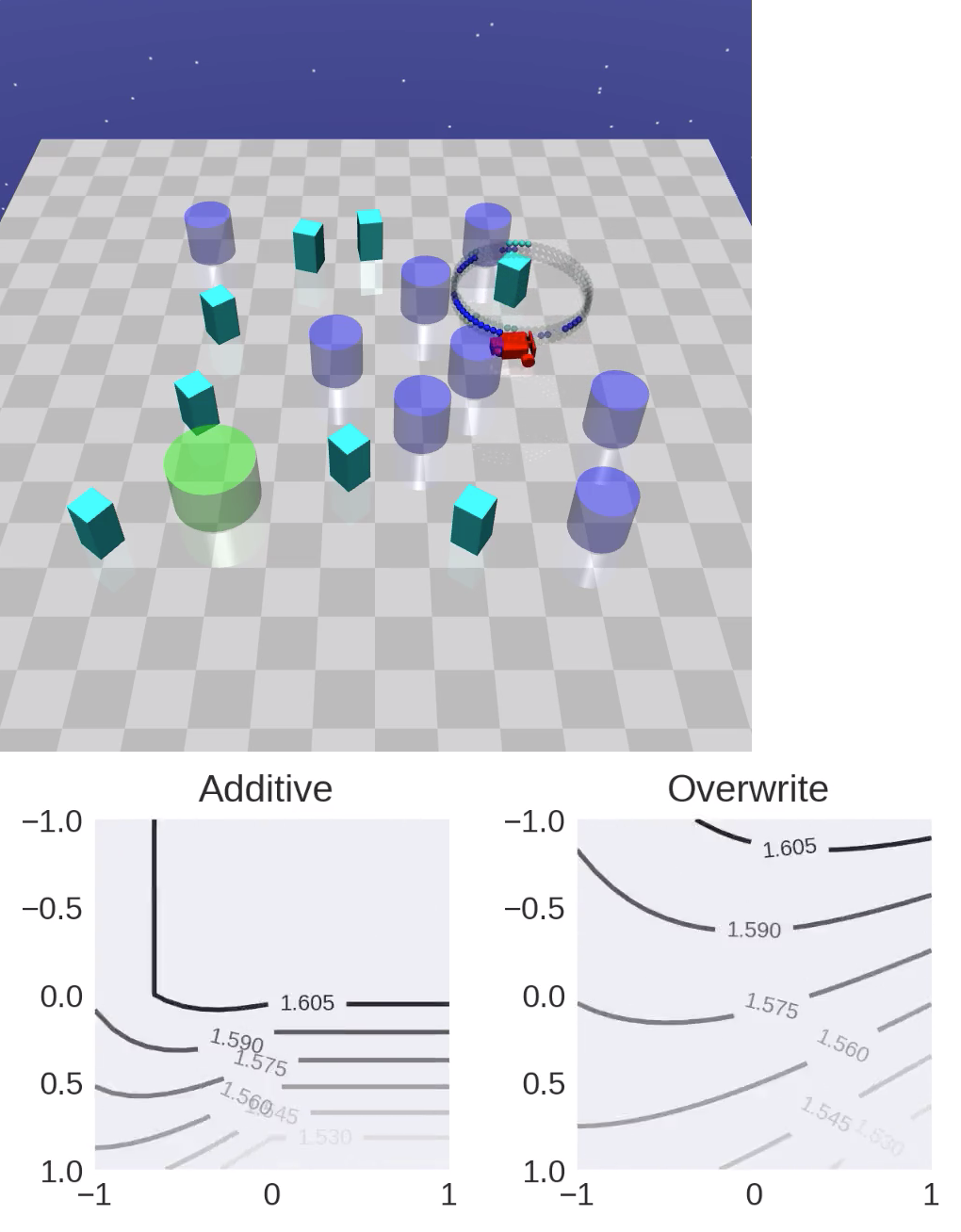}&
\includegraphics[width=0.2\textwidth]{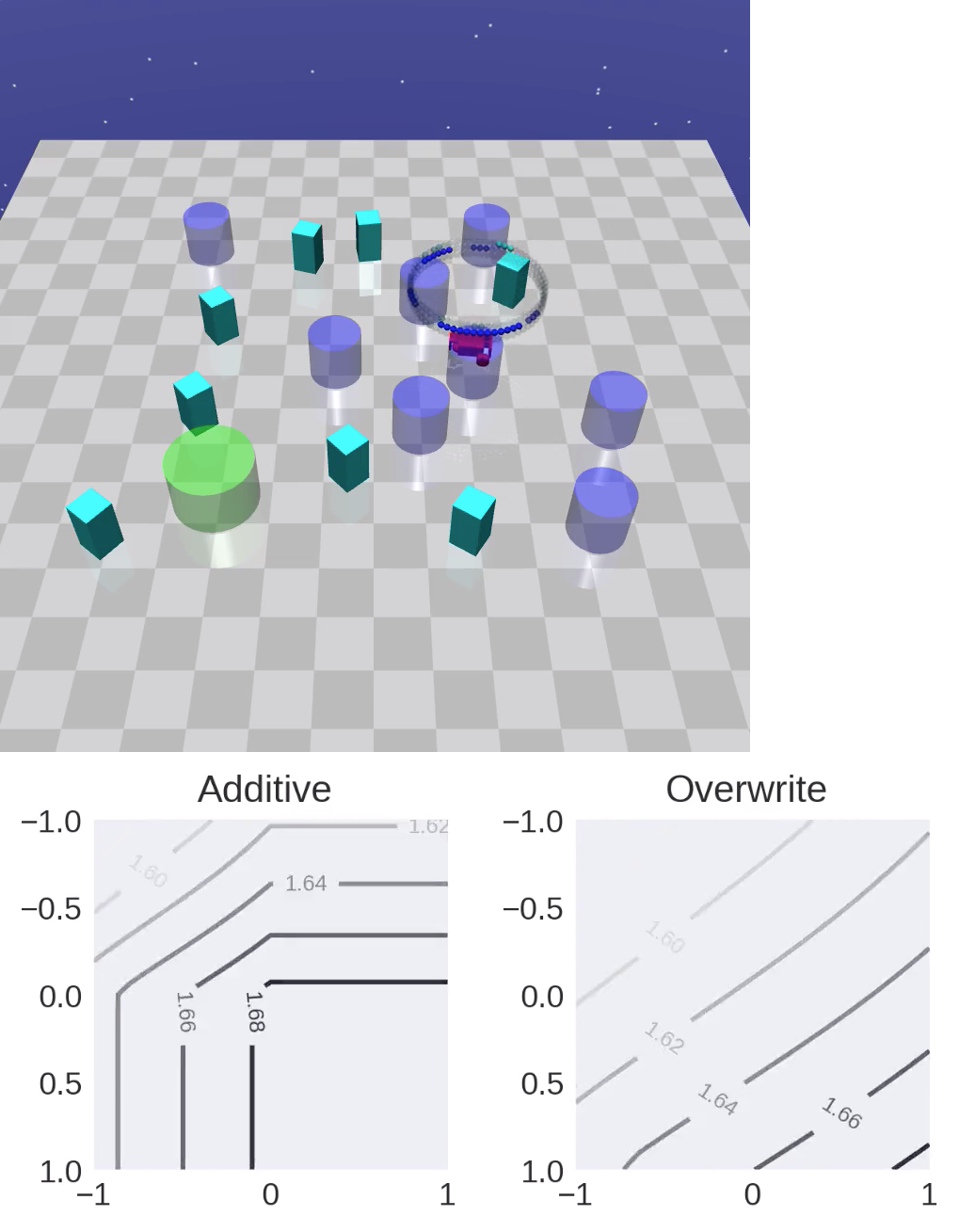}\\
\hline
\includegraphics[width=0.2\textwidth]{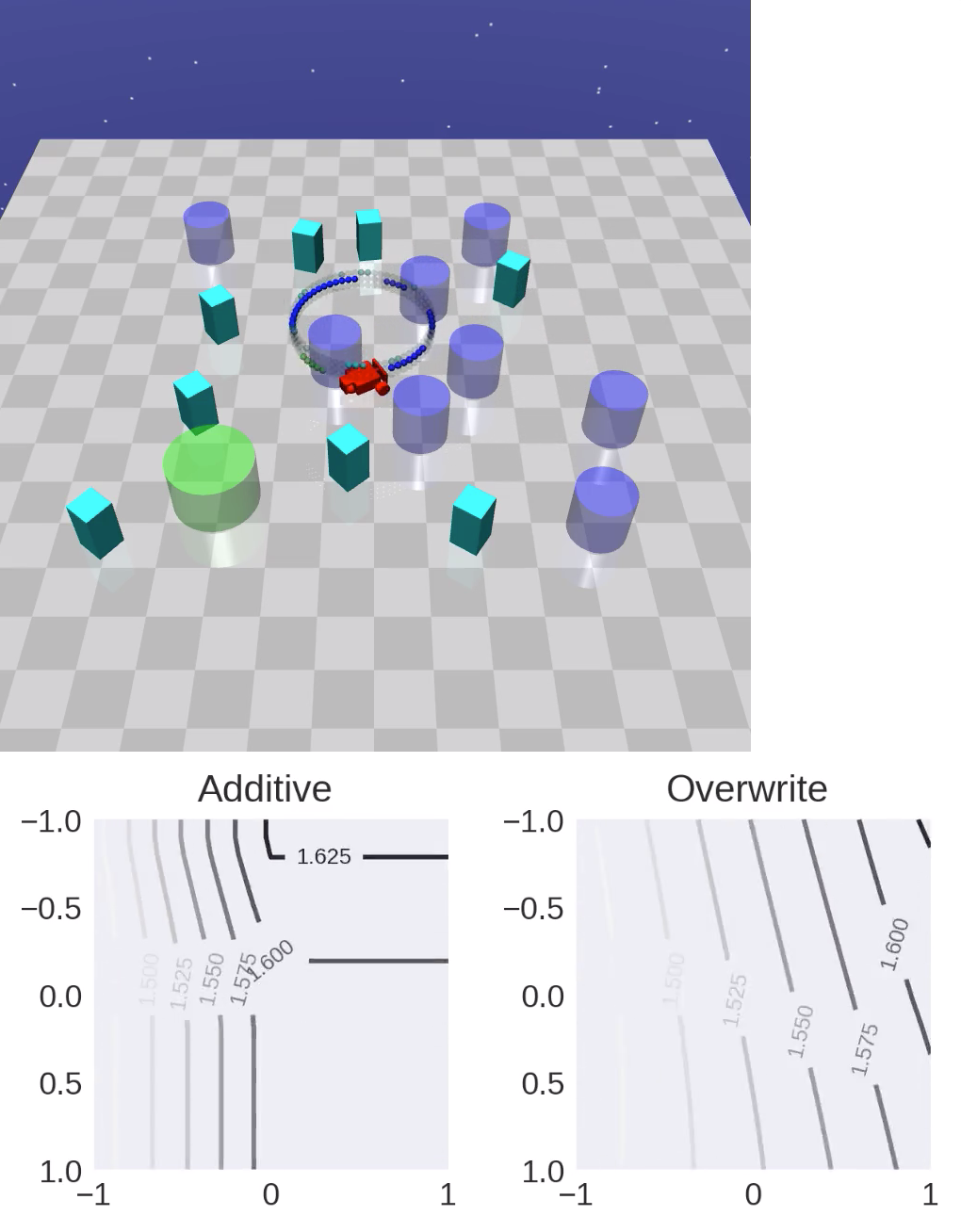}&
\includegraphics[width=0.2\textwidth]{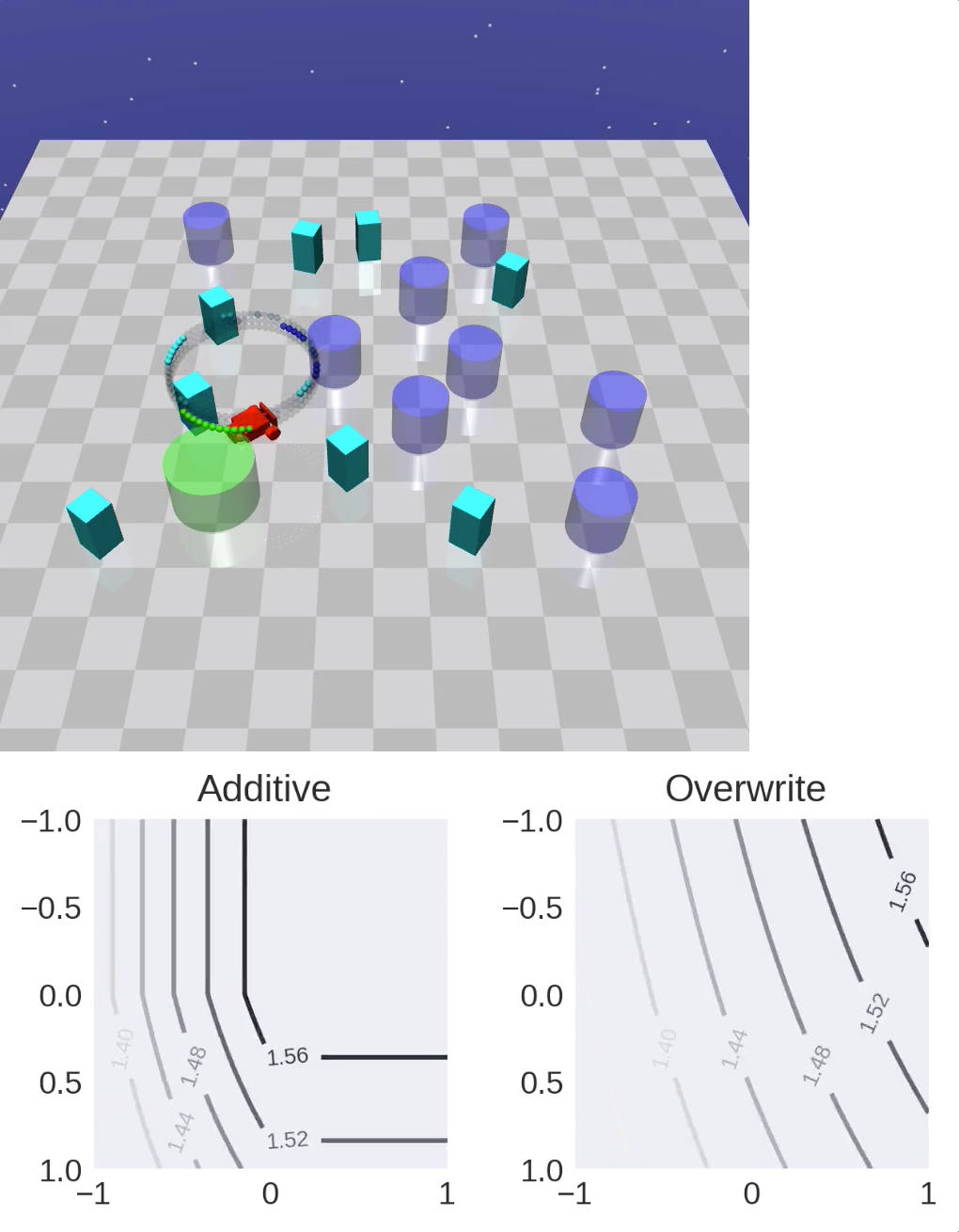}&
\includegraphics[width=0.2\textwidth]{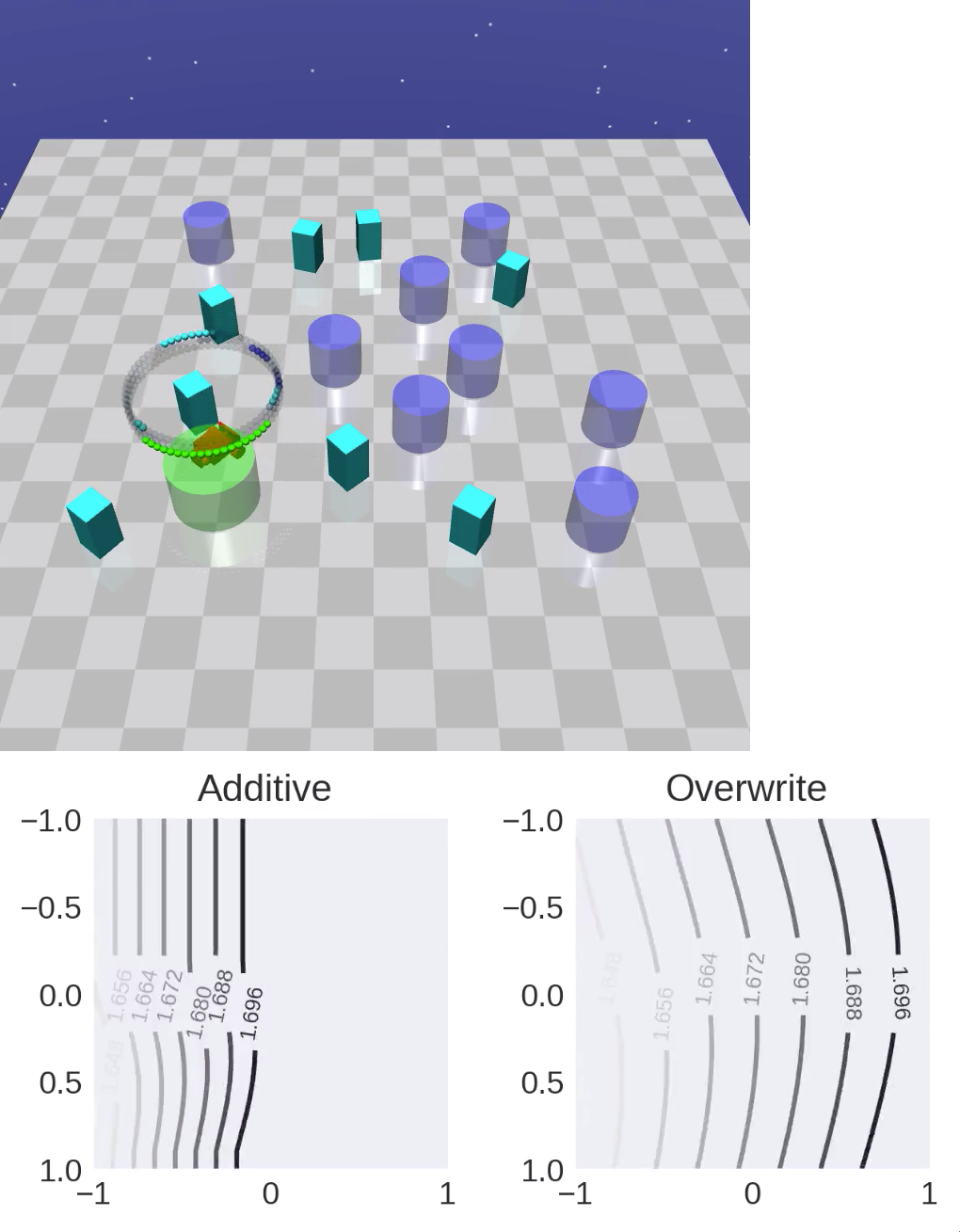}\\
\end{tabular}
}
\caption{Several key frames of an evaluation episode of \name{}. 
Each frame is paired with two visualized optimization landscapes, one using an additive action editing function while the other directly overwriting the action proposal $\hat{a}$ with $\Delta a$.
Each contour map is generated by enumerating $\Delta a$ given a sampled action proposal $\hat{a}$ from UM $\piu$, and evaluating the value of of Eq.~\ref{eq:objectives2} (b).
The map axes correspond to the agent's action dimensions (2D). 
Note that because we normalize the constraint reward by its moving average mean and standard deviation (Table~\ref{tab:hyperparameters_safety_gym}), the values on the map can be positive.
}
\label{fig:editing_function_frames}
\end{figure*}

\section{Action Editing Function}
\label{app:editing_function}
In Section~\ref{sec:approach}, our motivation for an additive action editing function is to ensure an easier optimization landscape for SE $\pis$.
We hypothesize that if each proposed $\hat{a}$ is ``mostly'' safe, then there is an inductive bias for $\pis$ to output $\Delta a \rightarrow 0$.
In Section~\ref{sec:experiments} on the Safety Gym tasks, we did observe that \name{}-overwrite (directly using $\Delta a$ as the final action) is worse than \name{}.
To further show why an additive editing function is beneficial, when evaluating \name{} we visualize the function surface of Eq.~\ref{eq:objectives2} (b) \emph{w.r.t.} $\Delta a$ given a sampled action proposal $\hat{a}$ in Figure~\ref{fig:editing_function_frames}.
It is clear that compared to an overwriting editing function, an additive editing function always has a much larger set of optimal $\Delta a$.
Furthermore, this set almost always covers those $\Delta a$ that are close to 0.
Thus the additive editing function does provide a very good inductive bias for SE $\pis$.

\section{Parameterization}
\label{app:parameterization}
We set $\lambda=\text{softplus}(\lambda_0)$ to enforce the Lagrangian multiplier $\lambda\ge 0$, where $\lambda_0$ is a real-valued variable.
Thus Eq.~\ref{eq:surrogate_obj} (b) becomes 
\begin{equation}
\label{eq:lambda0_obj}
\min_{\lambda_0} (\text{softplus}(\lambda_0)\Lambda_{\pia})
\end{equation}
as unconstrained optimization solved by typical SGD.

We parameterize both policies $\piu$ and $\pis$ as Beta distribution policies~\citep{chou2017improving}.
The advantage of Beta over Normal~\citep{Haarnoja2018} is that it natively has a bounded support of $[0,1]$ for a continuous action space. 
It avoids using squashing functions like tanh which could have numerical issues when computing the inverse mapping. 
With PyTorch, we can use the reparameterization trick for the Beta distribution to enable gradient computation in Eq.~\ref{eq:objectives2}.
Generally speaking, Eq.~\ref{eq:objectives2} is re-written as 
\begin{equation*}
\begin{array}{@{}ll@{}}
\text{(a)} & \displaystyle\max_{\phi}\expect_{\substack{s\sim\mathcal{D},\epsilon_1,\epsilon_2,\\a=h(f_{\phi}(s,\epsilon_1),f_{\psi}(s,f_{\phi}(s,\epsilon_1),\epsilon_2))}} \Big[Q(s,a;\theta)\Big], \\
\text{(b)} & \displaystyle\max_{\psi}\expect_{\substack{s\sim\mathcal{D},\epsilon_1,\epsilon_2,\\a=h(f_{\phi}(s,\epsilon_1),f_{\psi}(s,f_{\phi}(s,\epsilon_1),\epsilon_2))}} \Big[-d(a,f_{\phi}(s,\epsilon_1))+\lambda Q_c(s,a;\theta)\Big],\\
\end{array}
\end{equation*}
where $\epsilon_1$ and $\epsilon_2$ are sampled from two fixed noise distributions.
Then gradients can be easily computed for $\phi$ and $\psi$.

\section{Hyperparameters and Compute}
\label{app:hyperparameters}
In this section, we list the key hyperparameters used by the baselines and \name{}. 
%
%
A summary for the Safety Gym experiments is in Table~\ref{tab:hyperparameters_safety_gym}. 
For the safe racing experiments, we only list the differences with Table~\ref{tab:hyperparameters_safety_gym} in Table~\ref{tab:hyperparameters_safe_racing}.
For the remaining implementation details, we refer the reader to the source code 
\ifdefined\isaccepted
\url{https://github.com/hnyu/seditor}.
\else 
(to be released).
\fi

With the model hyperparameters and training configurations above, a single job (one random seed) of each compared method in Section~\ref{sec:experiments} takes up to 6 hours training on any of the Safety Gym tasks and up to 20 hours training on either safe racing task, on a single machine of Intel(R) Core(TM) i9-7960X CPU@2.80GHz with 32 CPU cores and one RTX 2080Ti GPU. 
In practice, we use our internal cluster with similar hardware to launch multiple jobs in parallel.

\begin{table*}[!h]
\centering
\resizebox{\textwidth}{!}{
    \begin{tabular}{@{}r|ccccc@{}}
    \textbf{Hyperparameter} & \textbf{PPO-Lag} & \textbf{FOCOPS} & \textbf{SAC} & \textbf{SAC-actor2x-Lag} & \textbf{\name{}}\\
    \hline\\
    Number of parallel environments & 32 & $\leftarrow$ & $\leftarrow$ & $\leftarrow$ & $\leftarrow$\\
    Initial rollout steps before training & N/A & N/A & $10000$ & $\leftarrow$ & $\leftarrow$\\
    Number of hidden $\text{layers}^{*}$ & 3 & $\leftarrow$ & $\leftarrow$ & $\leftarrow$ & $\leftarrow$\\
    Number of hidden units of each $\text{layer}^{*}$ & 256 & $\leftarrow$ & $\leftarrow$ & $\leftarrow$ & $\leftarrow$\\
    Beta distribution min concentration & $1.0$ & $\leftarrow$ & $\leftarrow$ & $\leftarrow$ & $\leftarrow$\\
    Frame stacking & 4 & $\leftarrow$ & $\leftarrow$ & $\leftarrow$ & $\leftarrow$\\
    Reward normalizer $\text{clipping}^{\circ}$ & $10.0$ & $\leftarrow$ & $\leftarrow$ & $\leftarrow$ & $\leftarrow$ \\
    Hidden activation & tanh & $\leftarrow$ & $\leftarrow$ & $\leftarrow$ & $\leftarrow$ \\
    Entropy regularization weight & $10^{-3}$ & N/A & N/A & N/A & N/A \\
    Entropy target per dimension & N/A & N/A & $-1.609^{\dagger}$ & $\leftarrow$ & $(-1.609,-1.609)$ \\
    KLD $\text{weight}^{\ddagger}$ & N/A & $1.5$ & N/A & N/A & N/A\\
    Trust region $\text{bound}^{\ddagger}$ & N/A & $0.02$ & N/A & N/A & N/A\\
    Initial Lagrangian multiplier $\lambda$ & $1.0$ & $\leftarrow$ & $\leftarrow$ & $\leftarrow$ & $\leftarrow$\\
    Learning rate of $\lambda$ & $0.01$ & $\leftarrow$ & $\leftarrow$ & $\leftarrow$ & $\leftarrow$\\
    Learning $\text{rate}^{\triangleright}$ & $10^{-4}$ & $\leftarrow$ & $3\times10^{-4}$ & $\leftarrow$ & $\leftarrow$\\
    Training interval (action steps per environment) & $8$ & $\leftarrow$ & $5$ & $\leftarrow$ & $\leftarrow$\\
    Mini-batch size & $256$ & $\leftarrow$ & $1024$ & $\leftarrow$ & $\leftarrow$ \\
    Mini-batch length for n-TD or GAE & $8$ & $\leftarrow$ & $\leftarrow$ & $\leftarrow$ & $\leftarrow$\\
    TD($\lambda$) for n-TD or GAE & $0.95$ & $\leftarrow$ & $\leftarrow$ & $\leftarrow$ & $\leftarrow$\\
    Discount $\gamma$ for both rewards & $0.99$ & $\leftarrow$ & $\leftarrow$ & $\leftarrow$ & $\leftarrow$\\
    Number of updates per training iteration & $10$ & $\leftarrow$ & $1$ & $\leftarrow$ & $\leftarrow$\\
    Target critic network update rate $\tau$ & N/A & N/A & $5\times10^{-3}$ & $\leftarrow$ & $\leftarrow$\\
    Target critic network update period & N/A & N/A & $1$ & $\leftarrow$ & $\leftarrow$\\
    Replay buffer size & N/A & N/A & $1.6\times10^6$ & $\leftarrow$ & $\leftarrow$\\ 
    \end{tabular}}
    \caption{Hyperparameters used in our experiments of Safety Gym for different approaches. 
    The symbol ``$\leftarrow$'' means the same value with the column on the left.
    $^{*}$Both for the policy and value/critic networks. SAC-actor2x-Lag has a double-size policy network.
    $^{\circ}$We normalize each dimension of the reward vector by its moving average mean and standard deviation, and the clipping is performed on normalized values.
    $^{\dagger}$This roughly assumes that the target action distribution has a probability mass concentrated on $\frac{1}{10}$ of the support $[-1,1]$.
    $^{\ddagger}$Following the FOCOPS paper~\citep{zhang2020order}.
    $^{\triangleright}$We explored both $10^{-4}$ and $3\times10^{-4}$ for PPO/FOCOPS, and the former was selected.
    }
    \label{tab:hyperparameters_safety_gym}
\end{table*}

\begin{table*}[!h]
\centering
\resizebox{\textwidth}{!}{
    \begin{tabular}{@{}r|c@{}ccc@{}c@{}}
    \textbf{Hyperparameter} & \textbf{PPO-Lag} & \textbf{FOCOPS} & \textbf{SAC} & \textbf{SAC-actor2x-Lag} & \textbf{\name{}}\\
    \hline\\
    Number of parallel environments & 16 & $\leftarrow$ & $\leftarrow$ & $\leftarrow$ & $\leftarrow$\\
    Initial rollout steps before training & N/A & N/A & $50000$ & $\leftarrow$ & $\leftarrow$\\
    CNN layers $(channels, kernel\ size, stride)^{*}$ & $(32,8,4),(64,4,2),(64,3,1)$ & $\leftarrow$ & $\leftarrow$ & $\leftarrow$ & $\leftarrow$\\
    Number of hidden layers after $\text{CNN}^{*}$ & 2 & $\leftarrow$ & $\leftarrow$ & $\leftarrow$ & $\leftarrow$\\
    Number of hidden units of each layer after $\text{CNN}^*$ & $256$ & $\leftarrow$ & $\leftarrow$ & $\leftarrow$ & $\leftarrow$\\
    Frame stacking & 1 & $\leftarrow$ & $\leftarrow$ & $\leftarrow$ & $\leftarrow$\\
    Hidden activation for CNN & relu & $\leftarrow$ & $\leftarrow$ & $\leftarrow$ & $\leftarrow$ \\
    Entropy regularization weight & $10^{-2}$ & N/A & N/A & N/A & N/A \\
    Entropy target per dimension & N/A & N/A & $-1.609^{\dagger}$ & $\leftarrow$ & $(-1.609,-0.916^{\dagger})$ \\
    Learning $\text{rate}$ & $3\times10^{-4}$ & $\leftarrow$ & $\leftarrow$ & $\leftarrow$ & $\leftarrow$\\
    Mini-batch size & $128$ & $\leftarrow$ & $256$ & $\leftarrow$ & $\leftarrow$ \\
    \end{tabular}}
    \caption{Hyperparameters used in our experiments of safe racing for different approaches. 
    The symbol ``$\leftarrow$'' means the same value with the column on the left.
    $^{*}$Both for the policy and value/critic networks. SAC-actor2x-Lag has a double-size policy network.
    $^{\dagger}$This roughly assumes that the target action distribution has a probability mass concentrated on $\frac{1}{5}$ of the support $[-1,1]$.
    }
    \label{tab:hyperparameters_safe_racing}
\end{table*}

\clearpage 

\section{Success and Failure Modes}
\label{app:example_frames}
\begin{figure*}[hbt!]
\centering
\resizebox{\textwidth}{!}{
\begin{tabular}{@{}c@{}}
\begin{tabular}{@{}l@{}c@{}c@{}c@{}c@{}c@{}}
\textsc{CarGoal2} & & & & &\\
\includegraphics[width=0.2\textwidth]{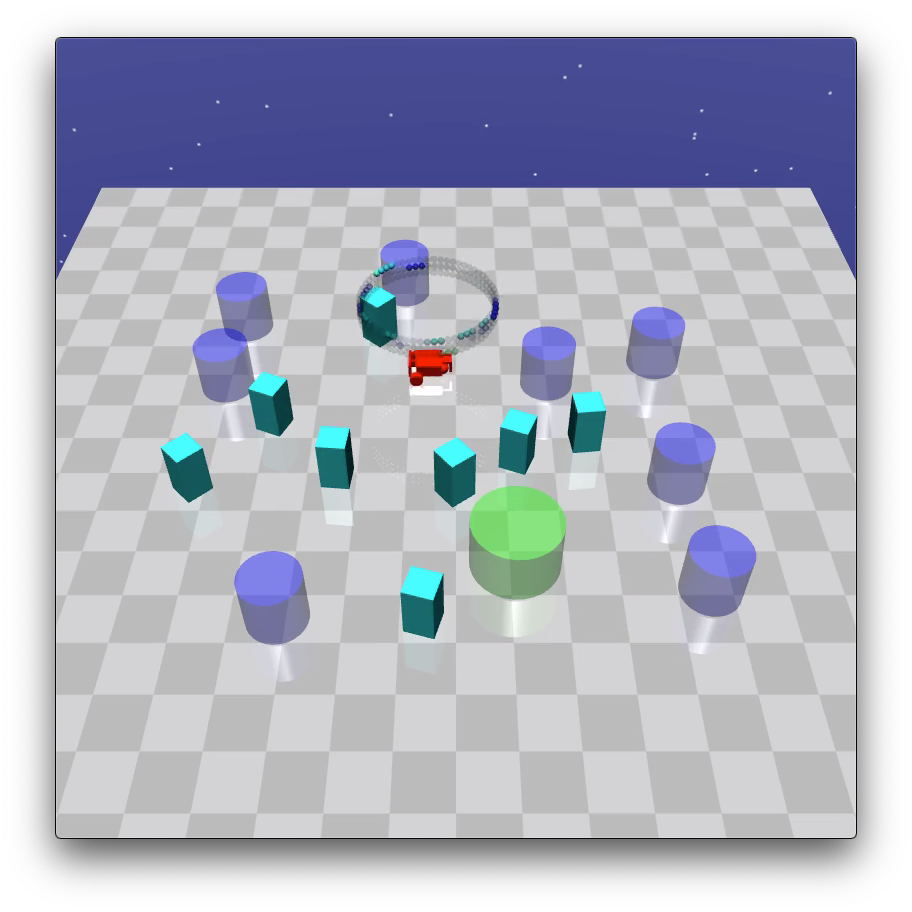}
&\includegraphics[width=0.2\textwidth]{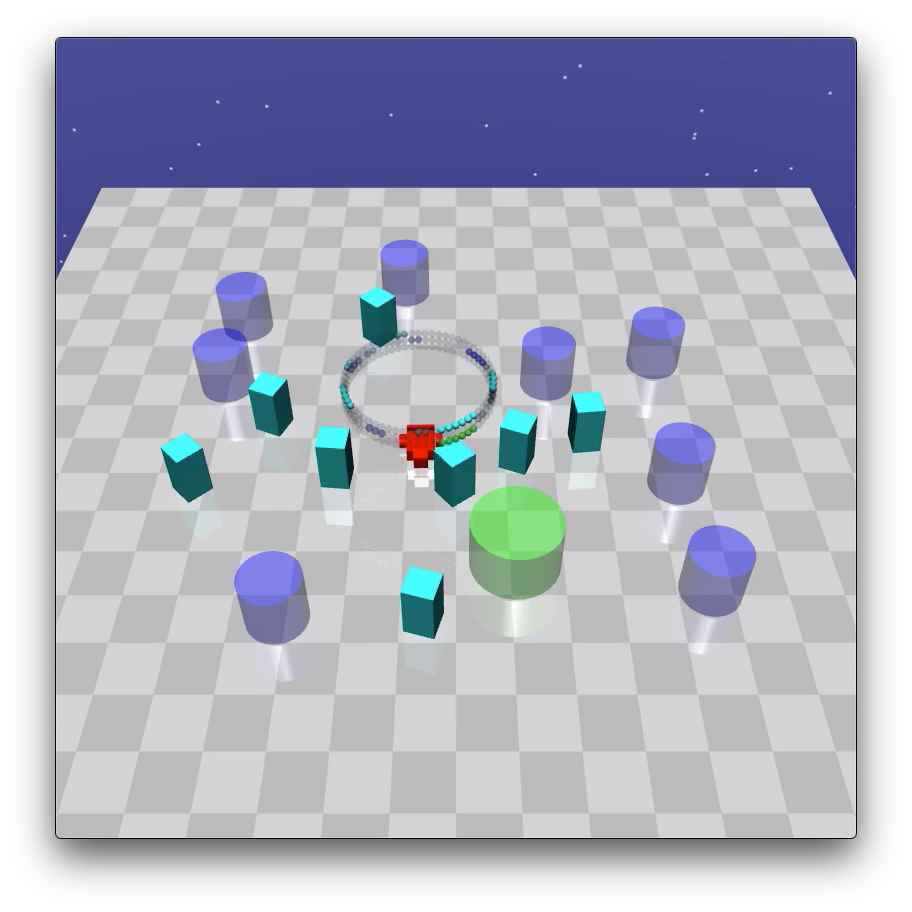}
&\includegraphics[width=0.2\textwidth]{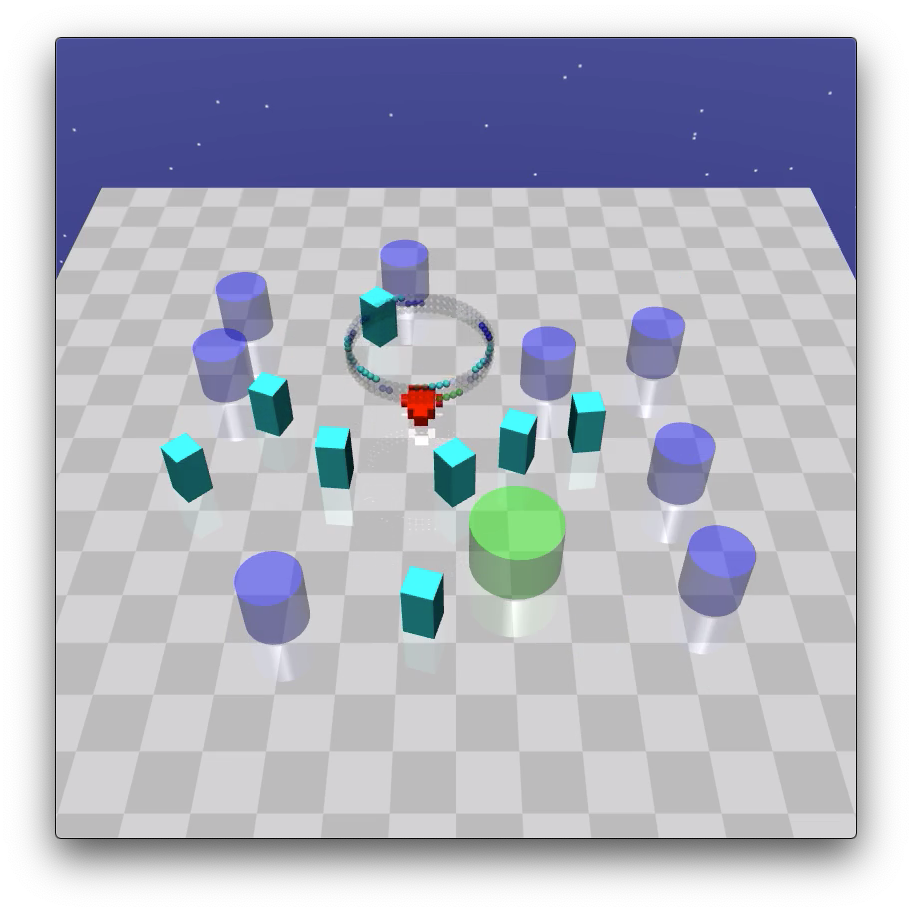}
&\includegraphics[width=0.2\textwidth]{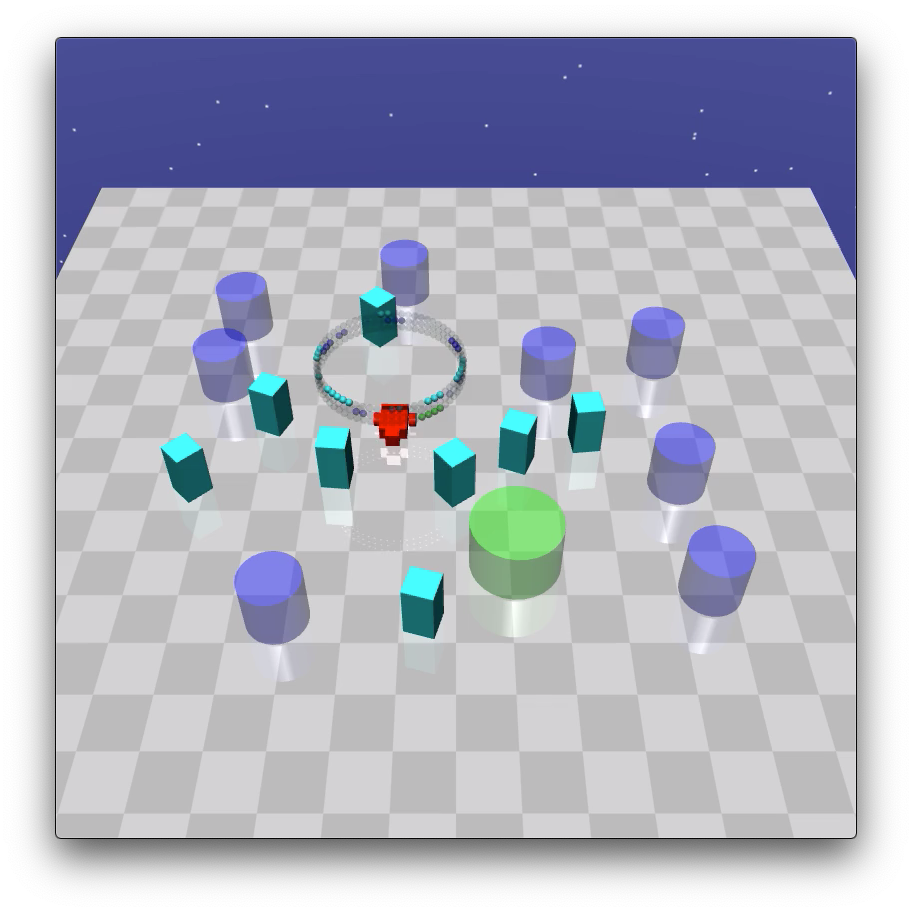}
&\includegraphics[width=0.2\textwidth]{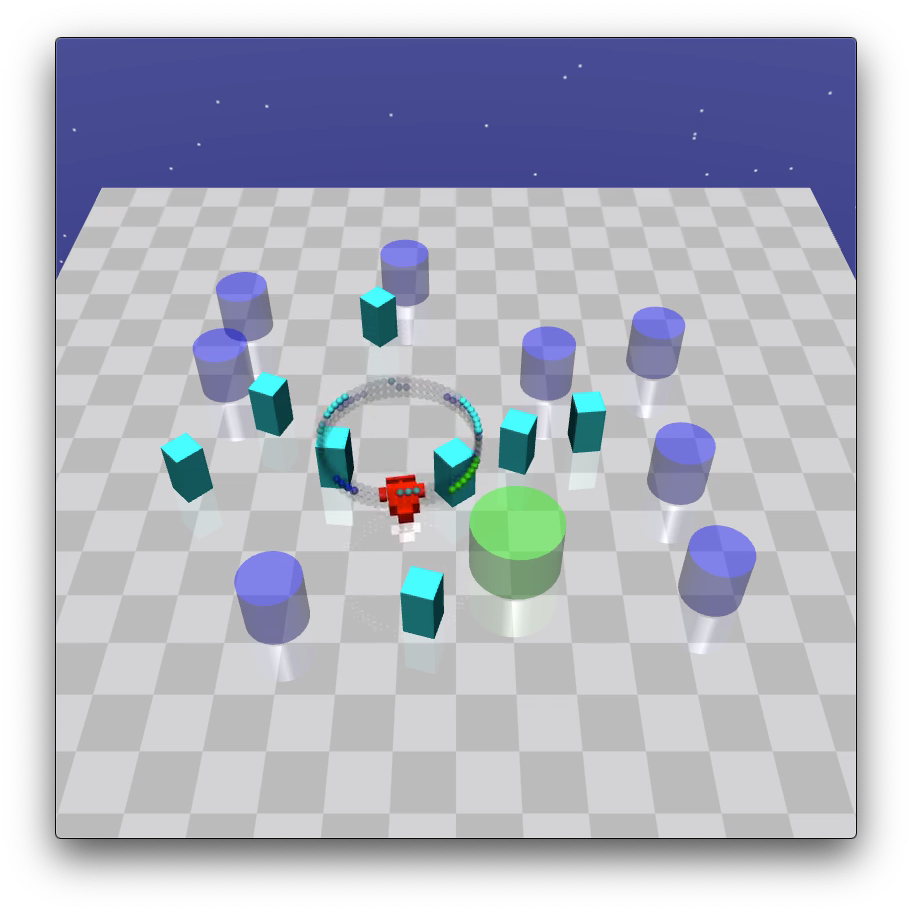}
&\includegraphics[width=0.2\textwidth]{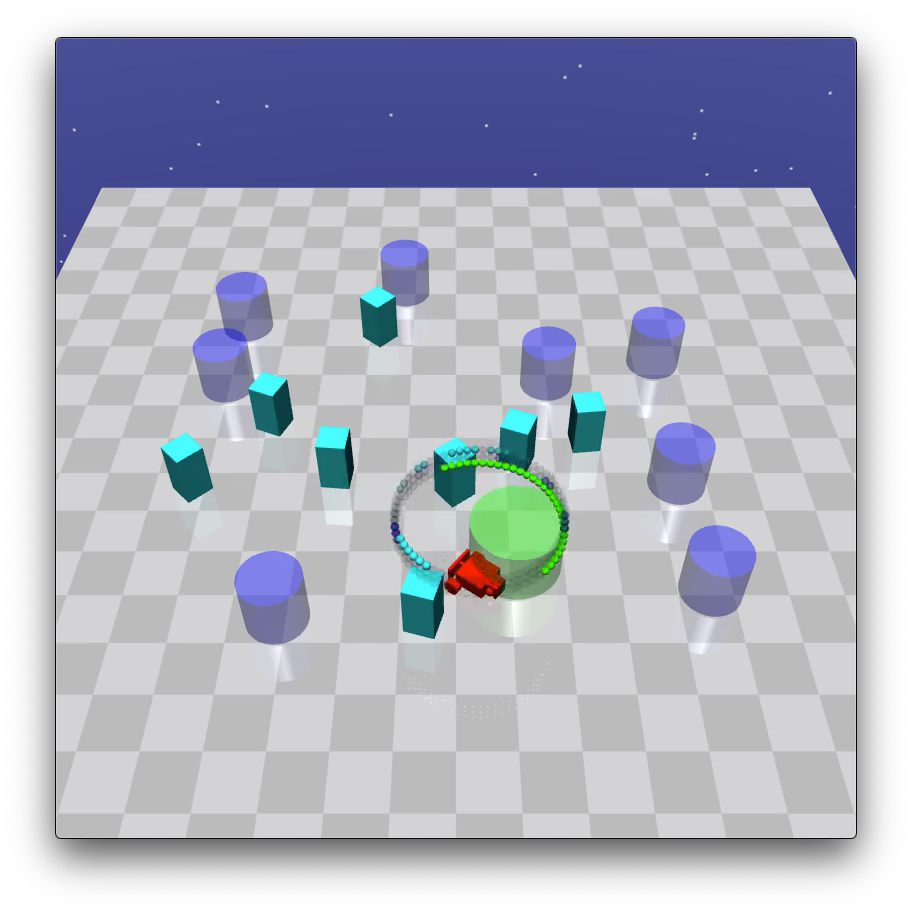}\\
\textsc{CarButton2} & & & & &\\
\includegraphics[width=0.2\textwidth]{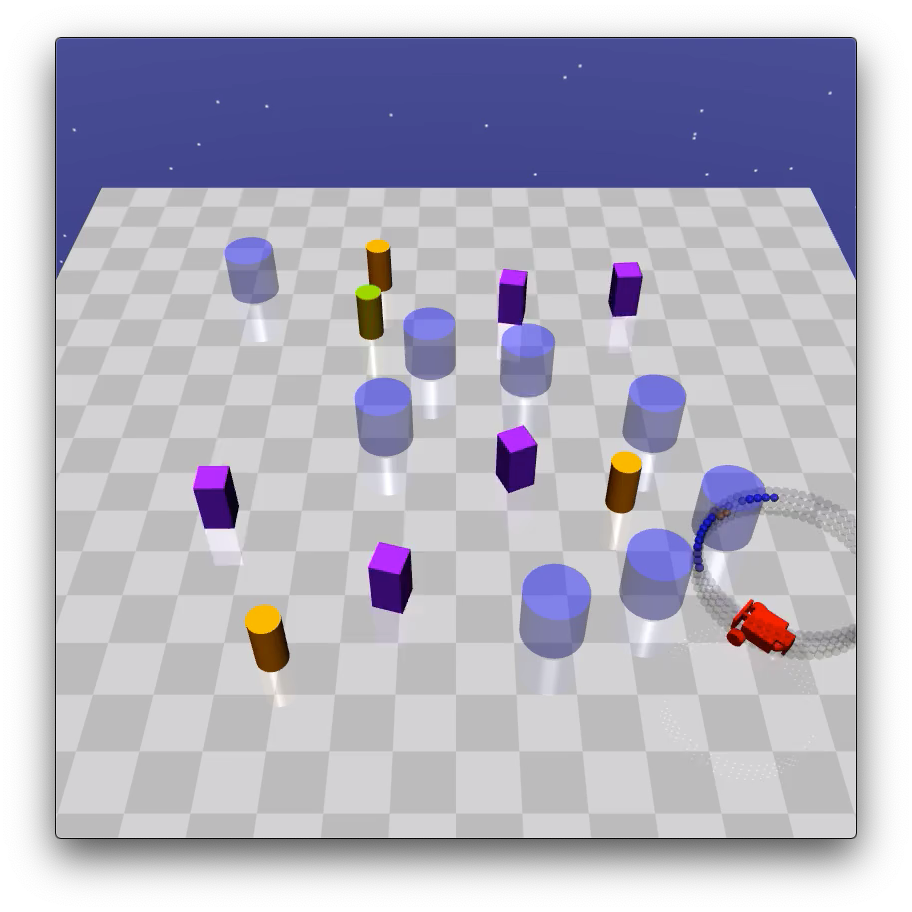}
&\includegraphics[width=0.2\textwidth]{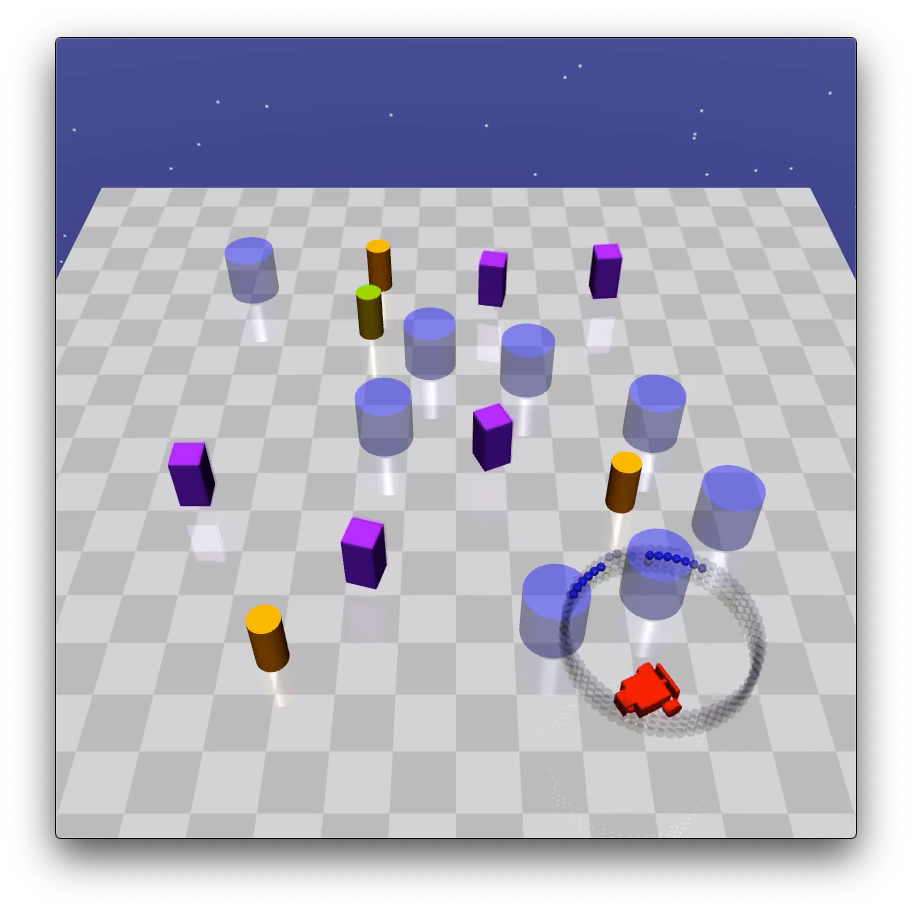}
&\includegraphics[width=0.2\textwidth]{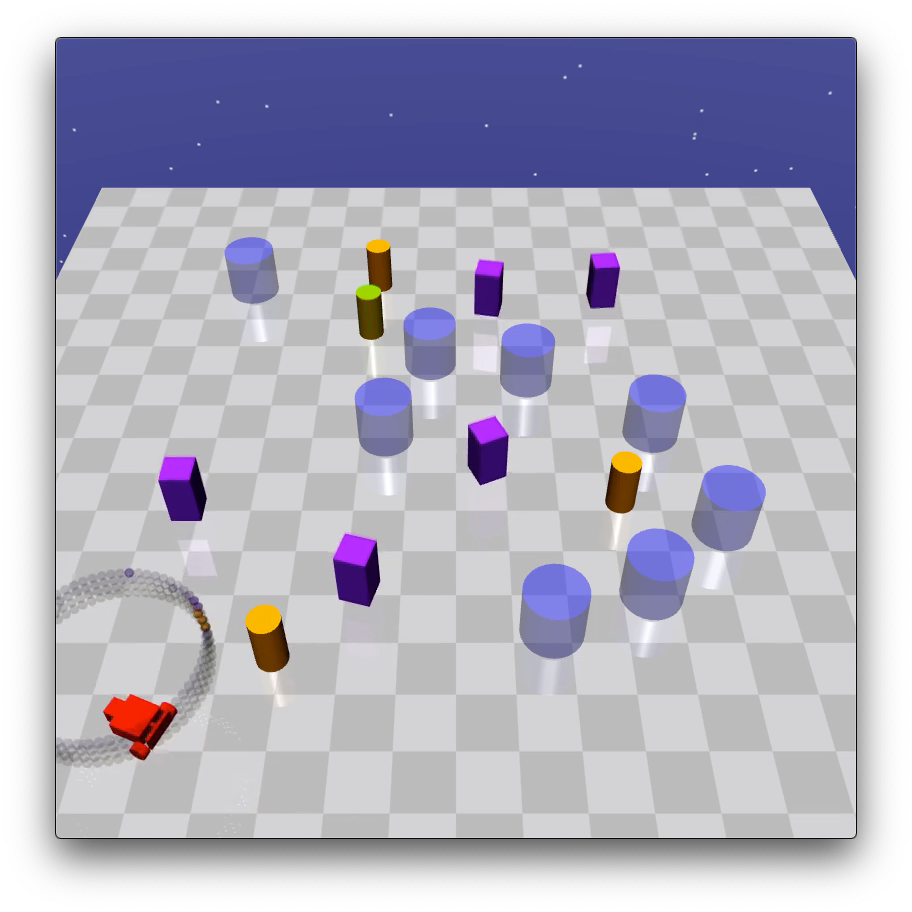}
&\includegraphics[width=0.2\textwidth]{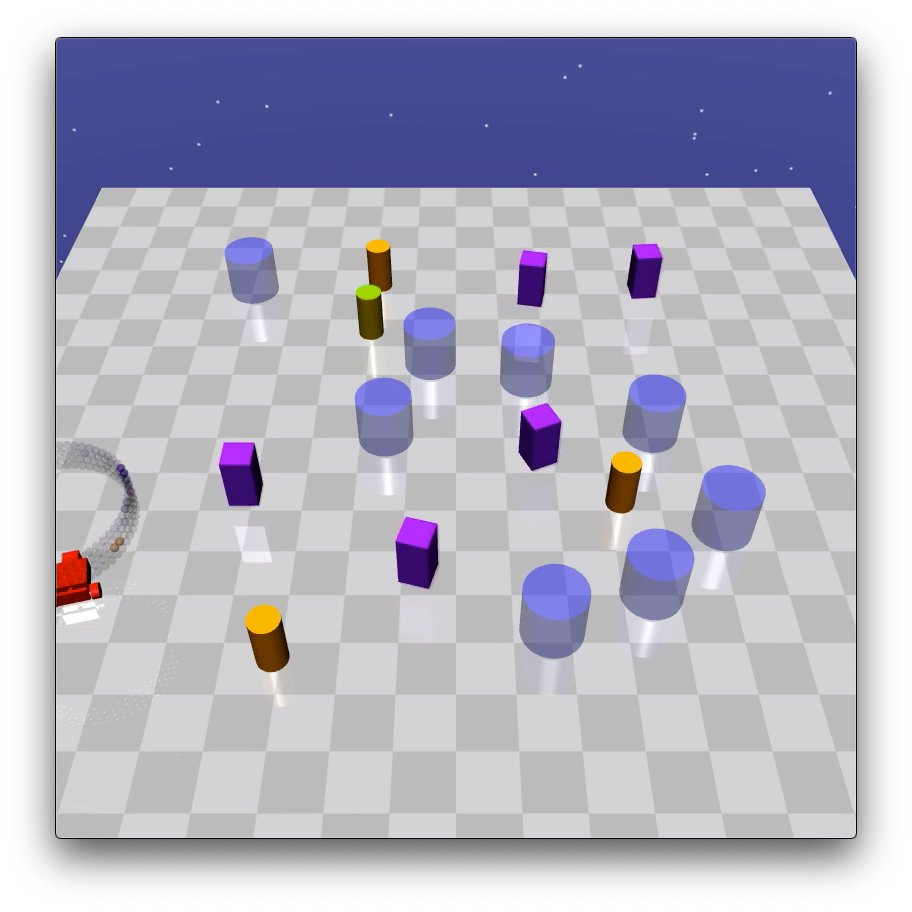}
&\includegraphics[width=0.2\textwidth]{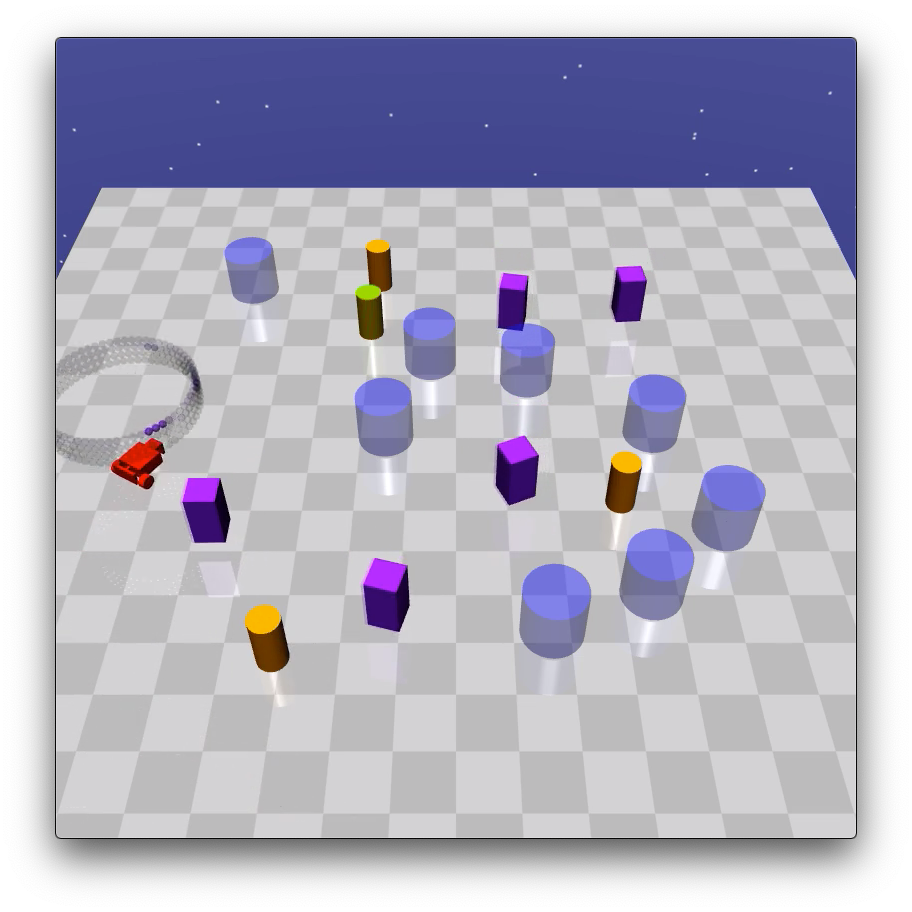}
&\includegraphics[width=0.2\textwidth]{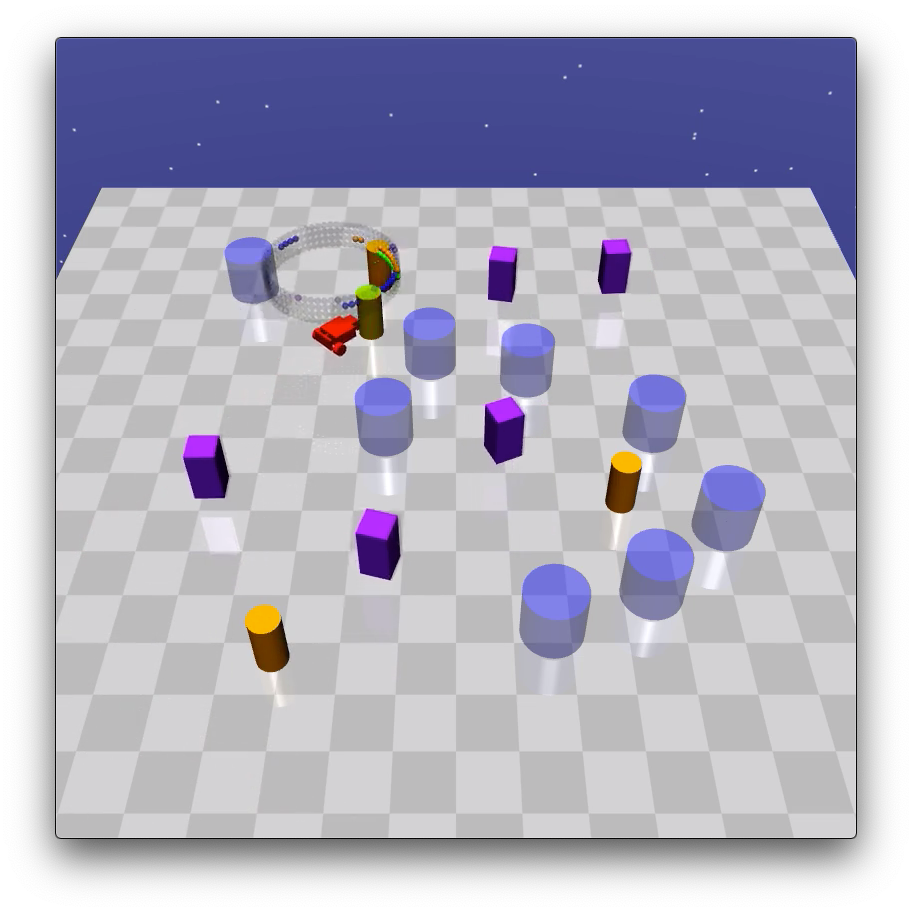}\\
\textsc{CarPush2} & & & & &\\
\includegraphics[width=0.2\textwidth]{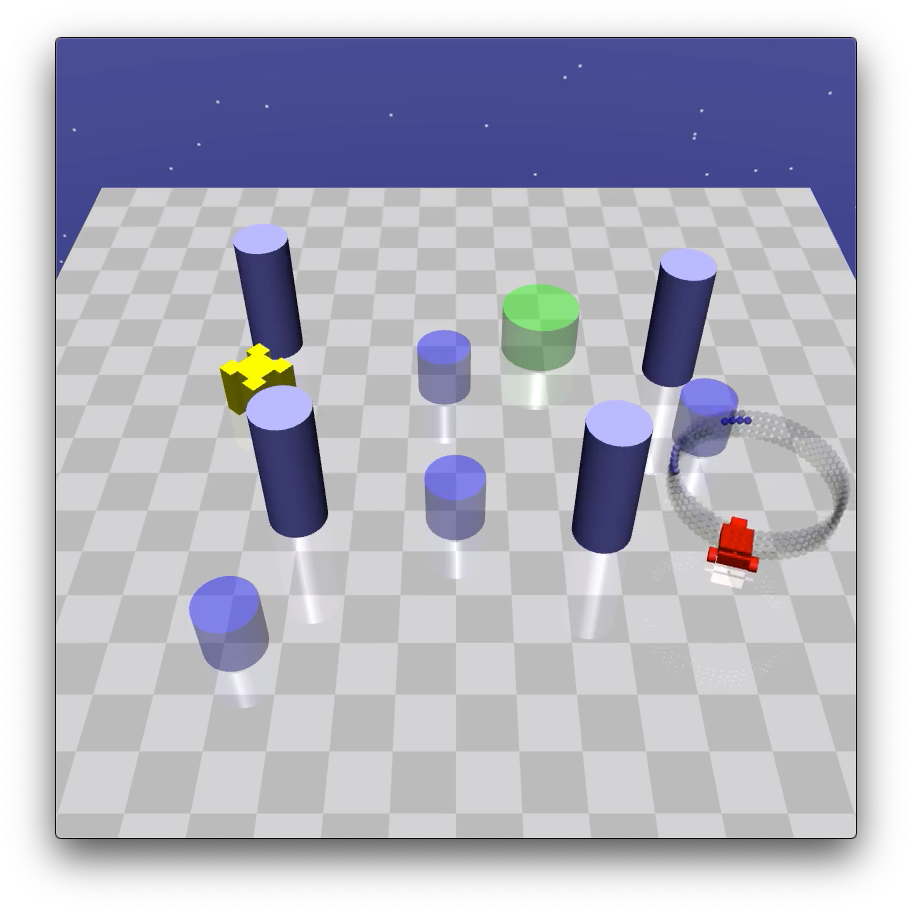}
&\includegraphics[width=0.2\textwidth]{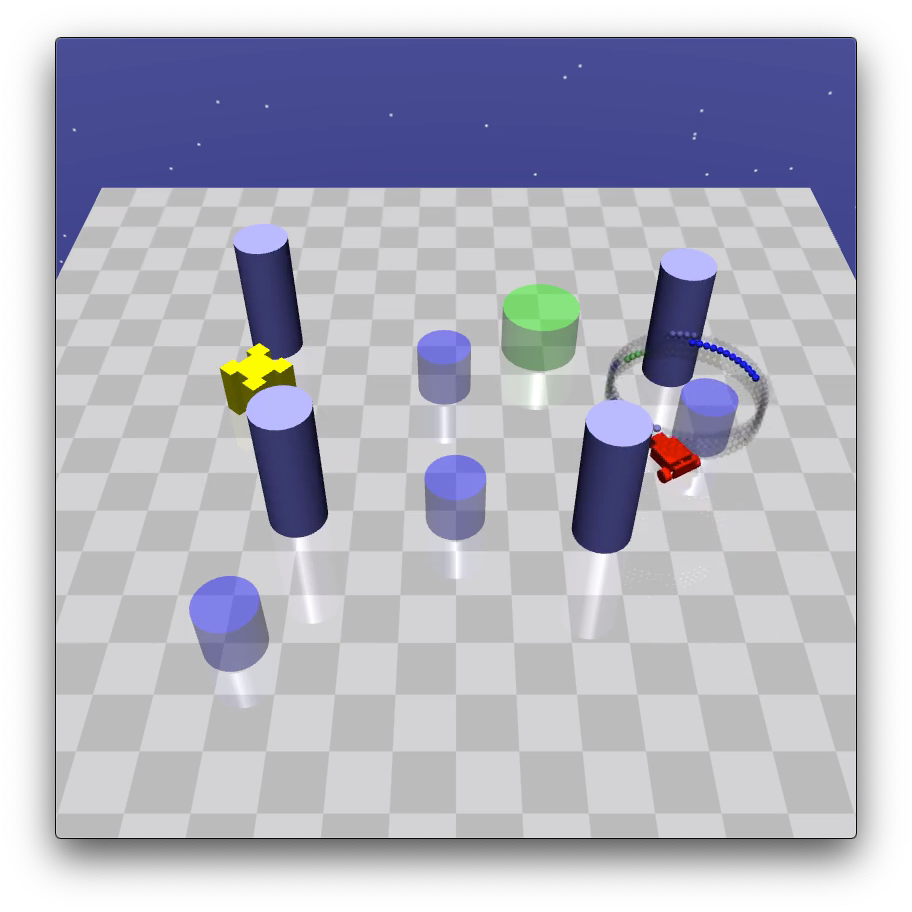}
&\includegraphics[width=0.2\textwidth]{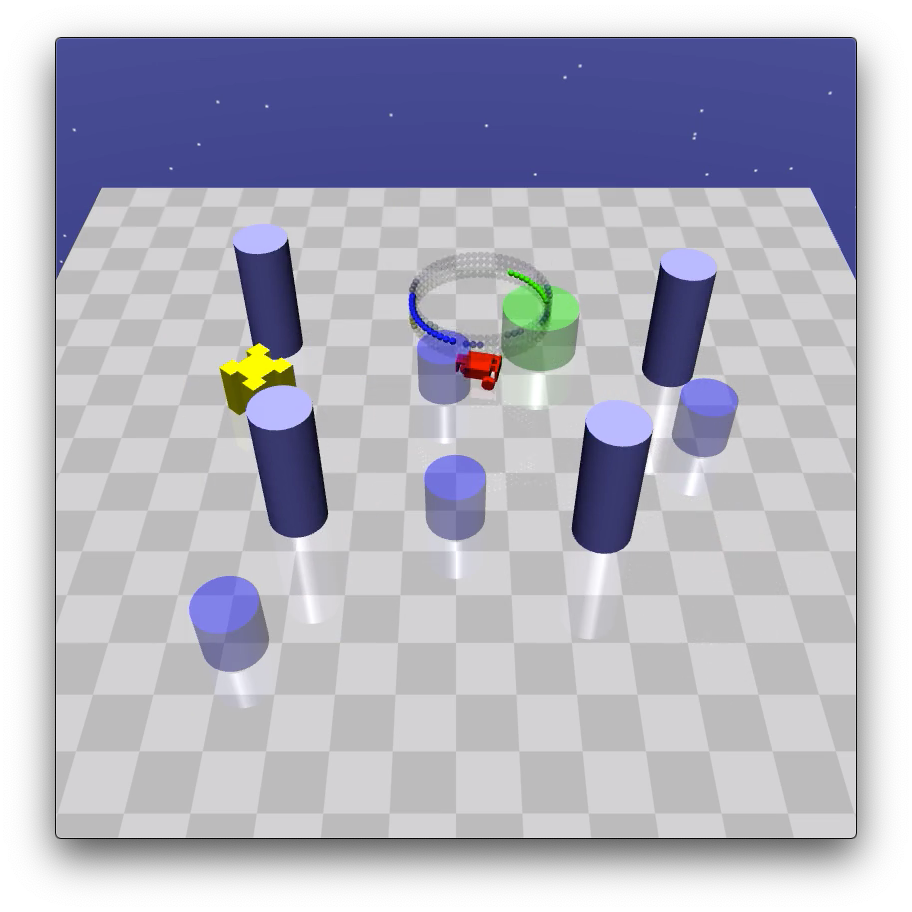}
&\includegraphics[width=0.2\textwidth]{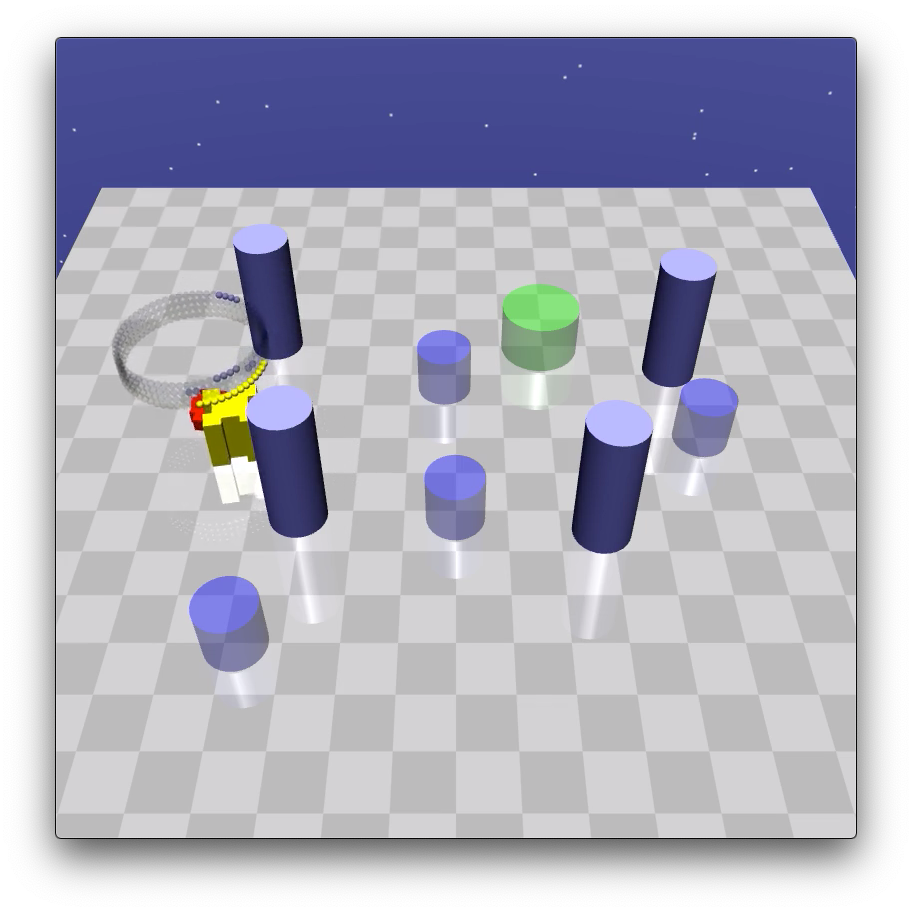}
&\includegraphics[width=0.2\textwidth]{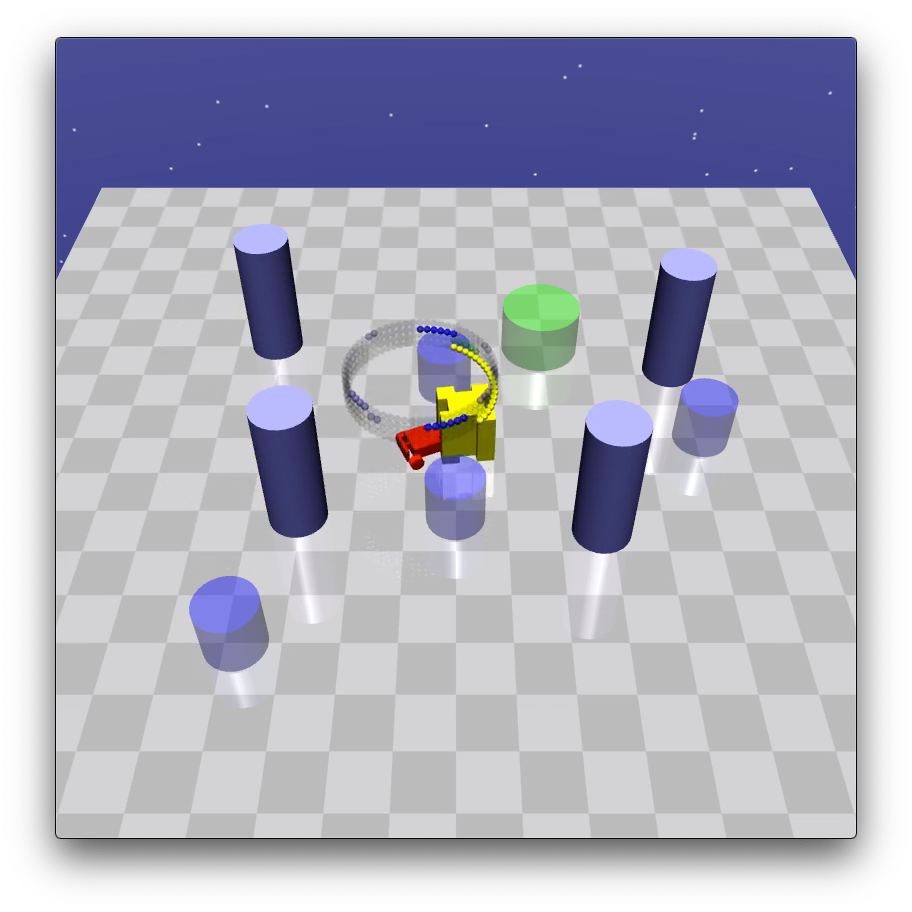}
&\includegraphics[width=0.2\textwidth]{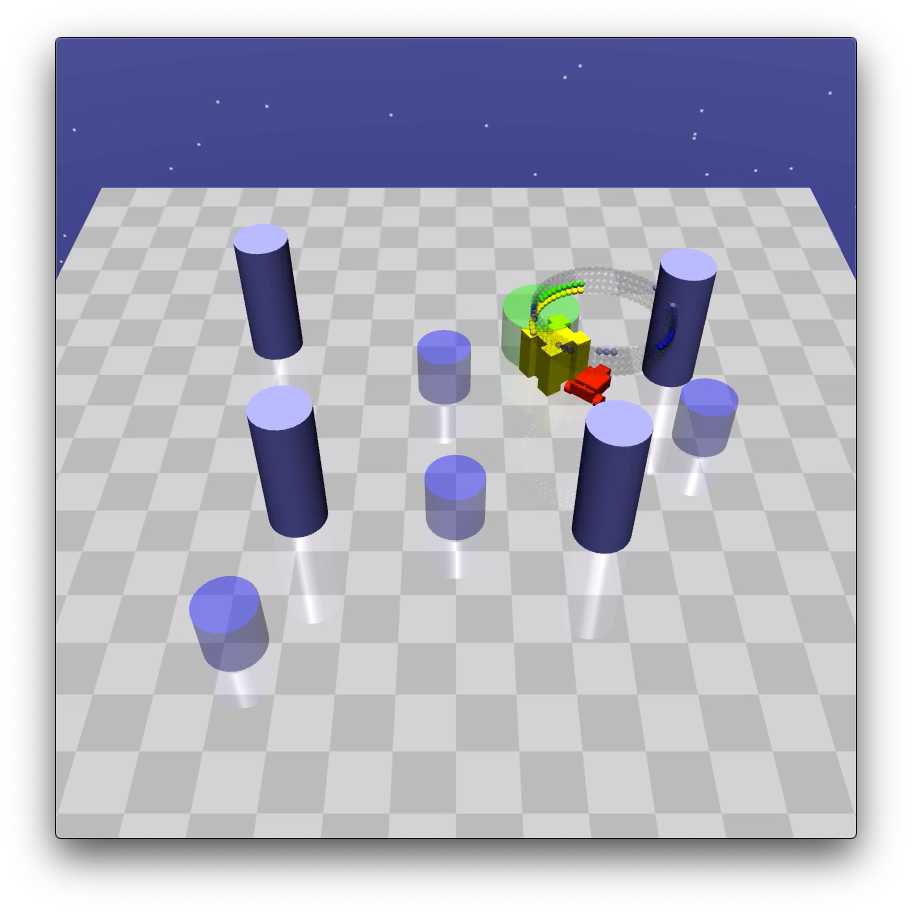}\\
\end{tabular}\\
\begin{tabular}{@{}l@{}c@{}c@{}c@{}}
\textsc{SafeRacing} & & & \\
\includegraphics[width=0.3\textwidth]{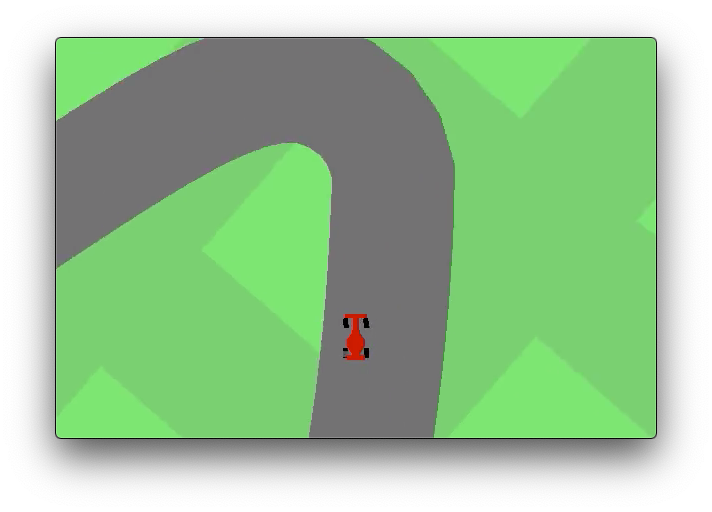}
&\includegraphics[width=0.3\textwidth]{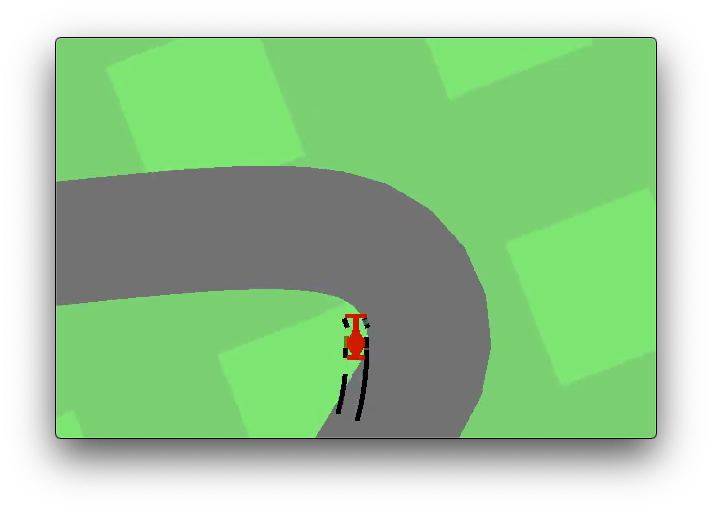}
&\includegraphics[width=0.3\textwidth]{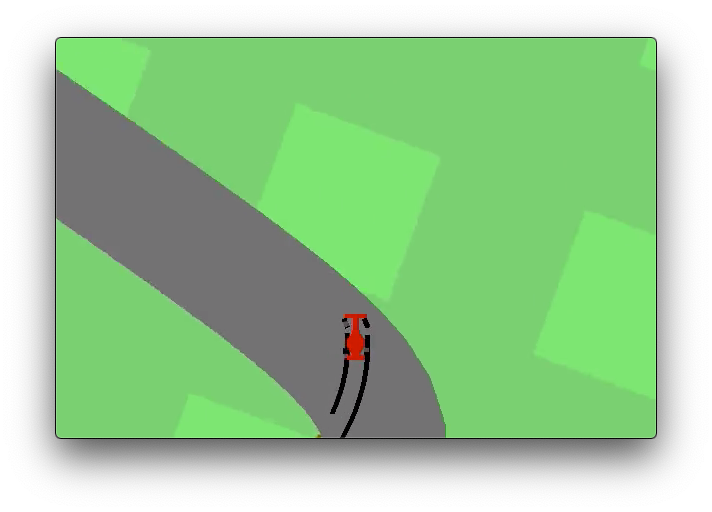}
&\includegraphics[width=0.3\textwidth]{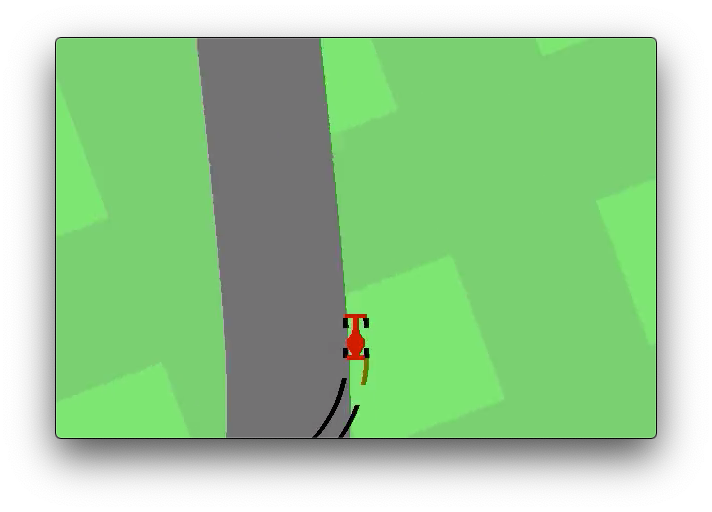}\\
\textsc{SafeRacingObstacle} & & & \\
\includegraphics[width=0.3\textwidth]{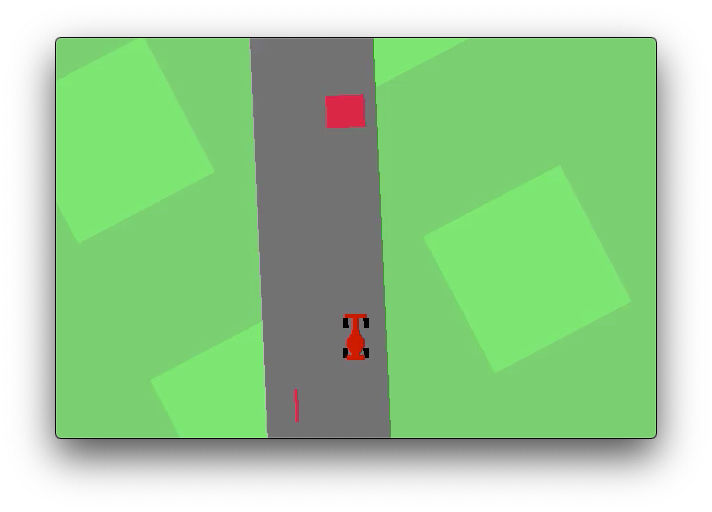}
&\includegraphics[width=0.3\textwidth]{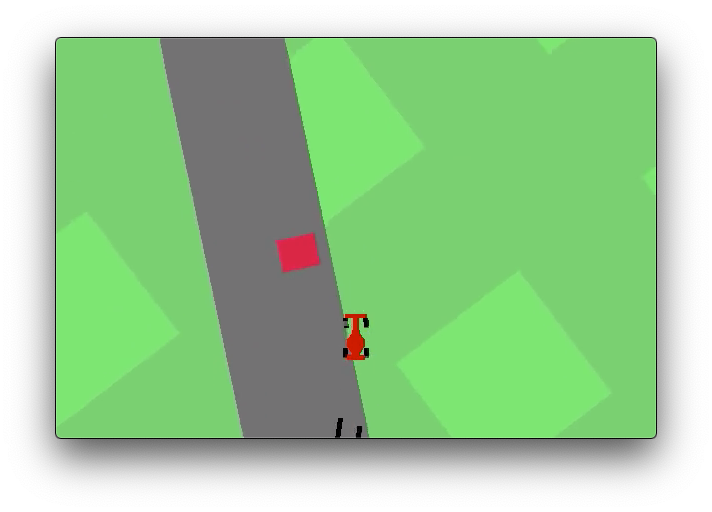}
&\includegraphics[width=0.3\textwidth]{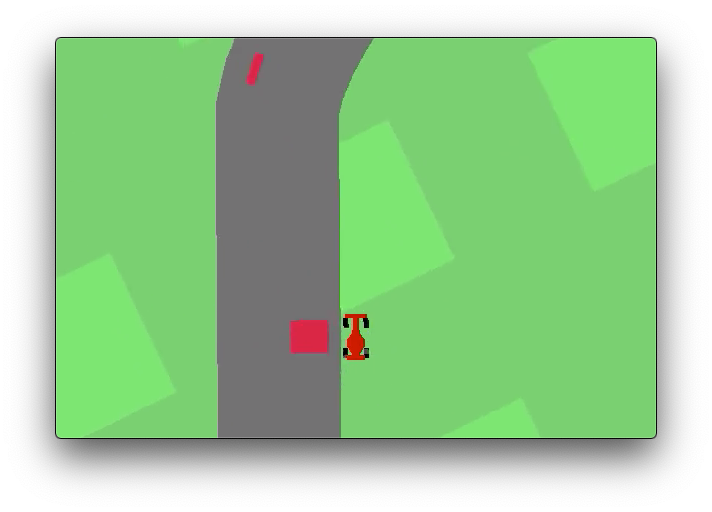}
&\includegraphics[width=0.3\textwidth]{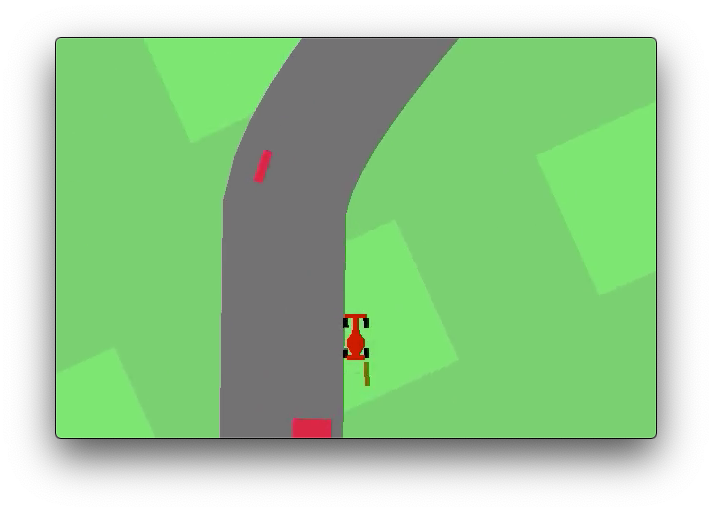}\\
\end{tabular}
\end{tabular}
}
\caption{Key frames of several successful episodes of our approach.
The robot in each episode finishes the task without violating any constraint.
For the safe racing tasks, we only show one representative segment of the track due to space limit. 
}
\label{fig:success_episodes}
\end{figure*}

\begin{figure*}[hbt!]
\centering
\resizebox{\textwidth}{!}{
\begin{tabular}{@{}c@{}}
\begin{tabular}{@{}l@{}c@{}c@{}c@{}c@{}c@{}}
\textsc{CarGoal2} & & & & &\\
\includegraphics[width=0.2\textwidth]{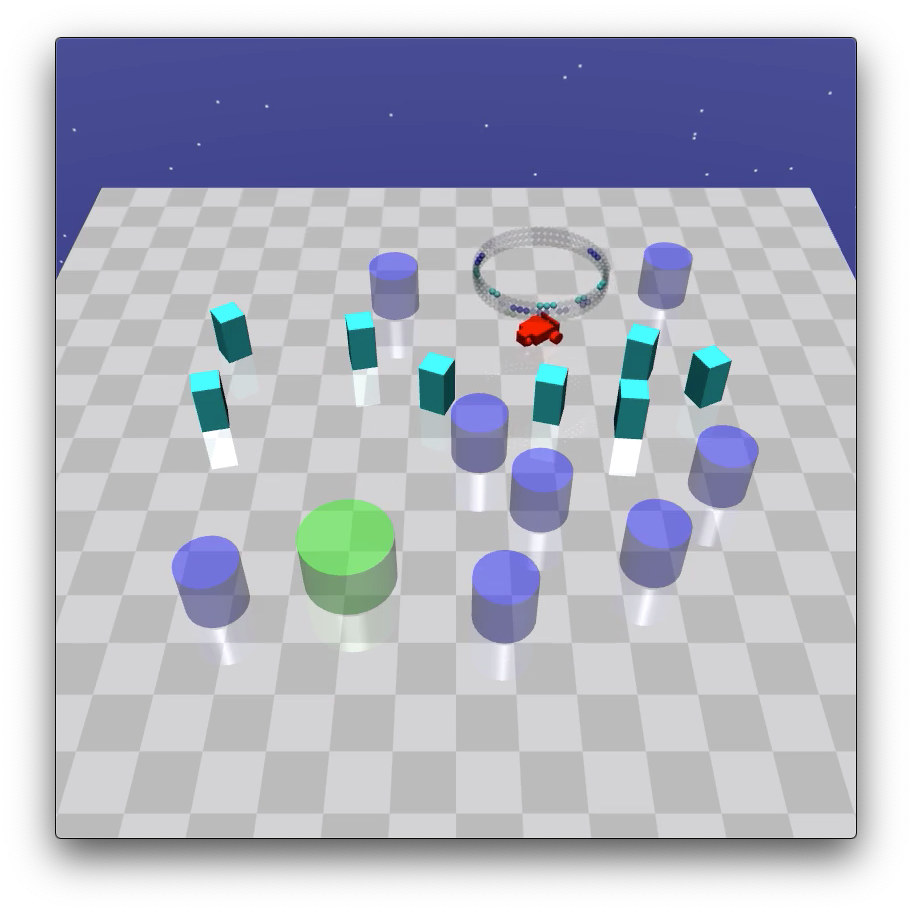}
&\includegraphics[width=0.2\textwidth]{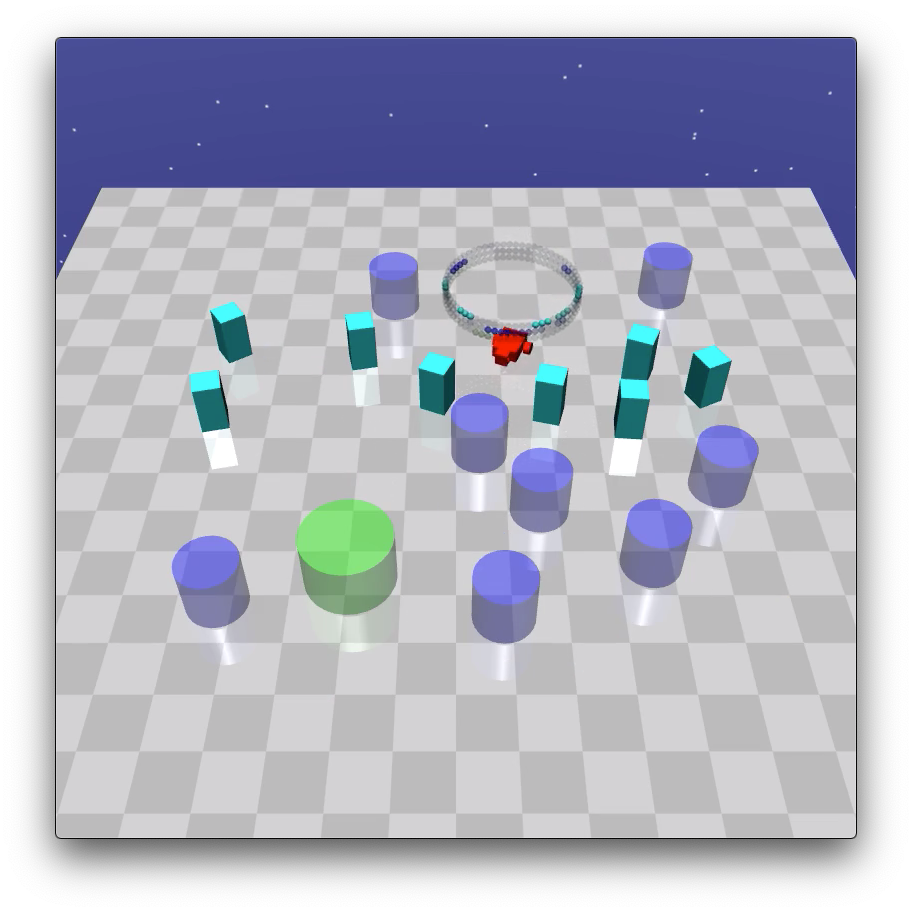}
&\includegraphics[width=0.2\textwidth]{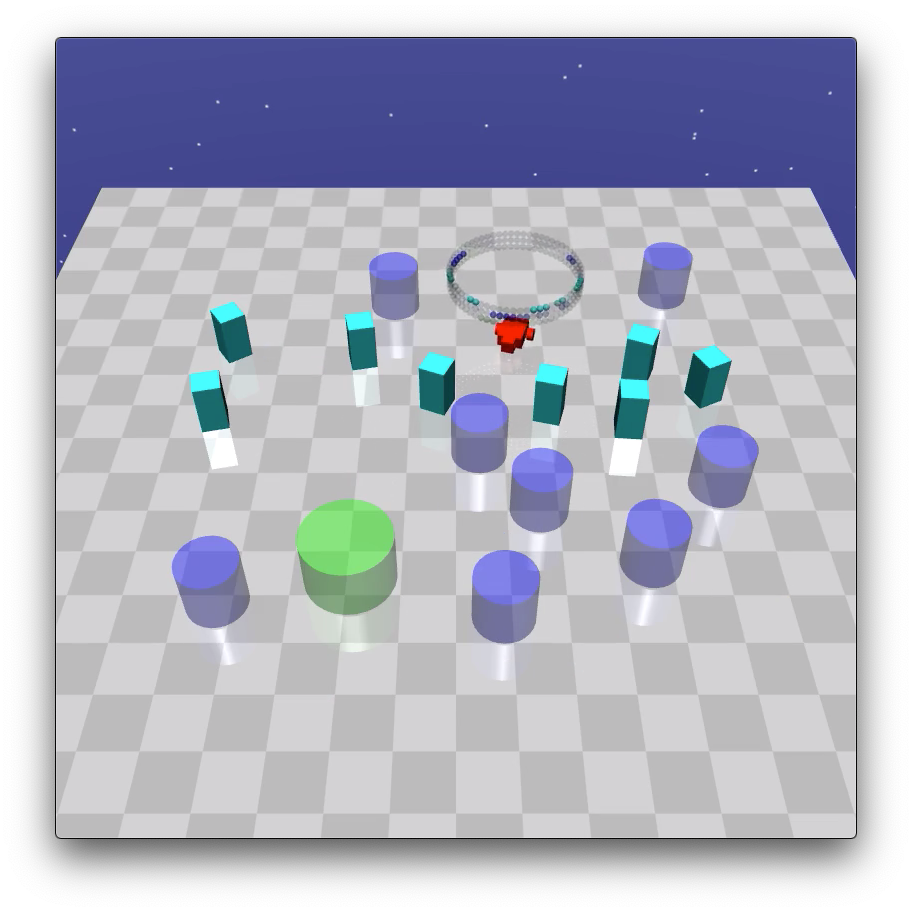}
&\includegraphics[width=0.2\textwidth]{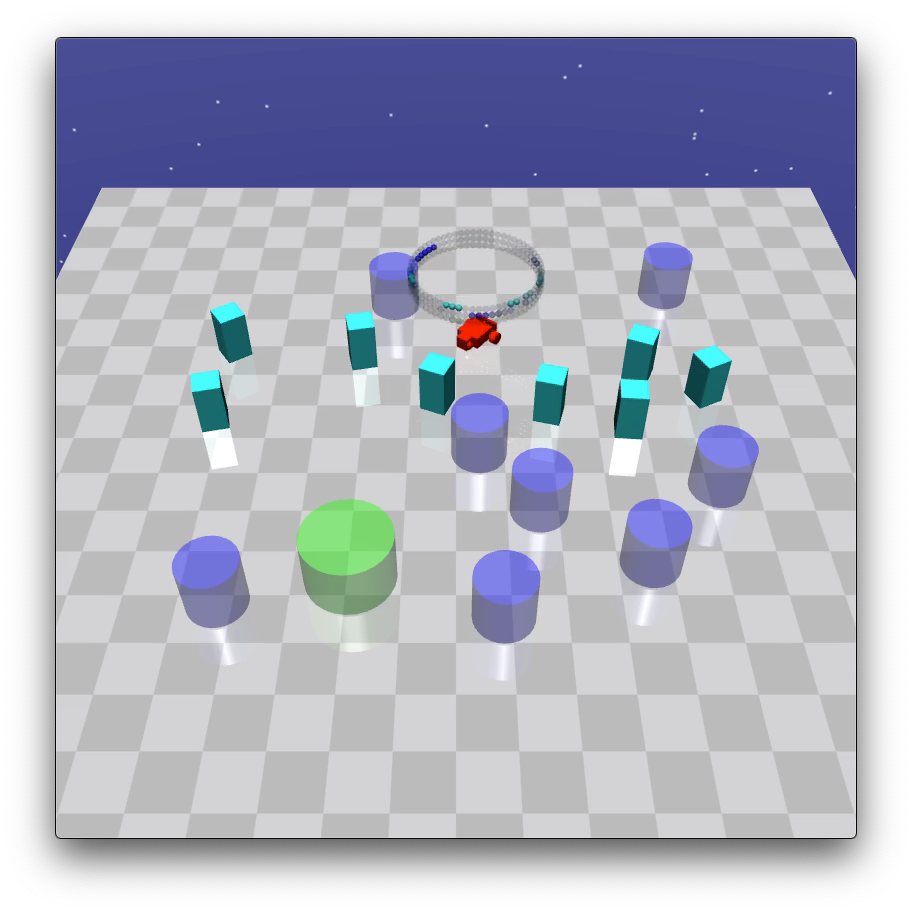}
&\includegraphics[width=0.2\textwidth]{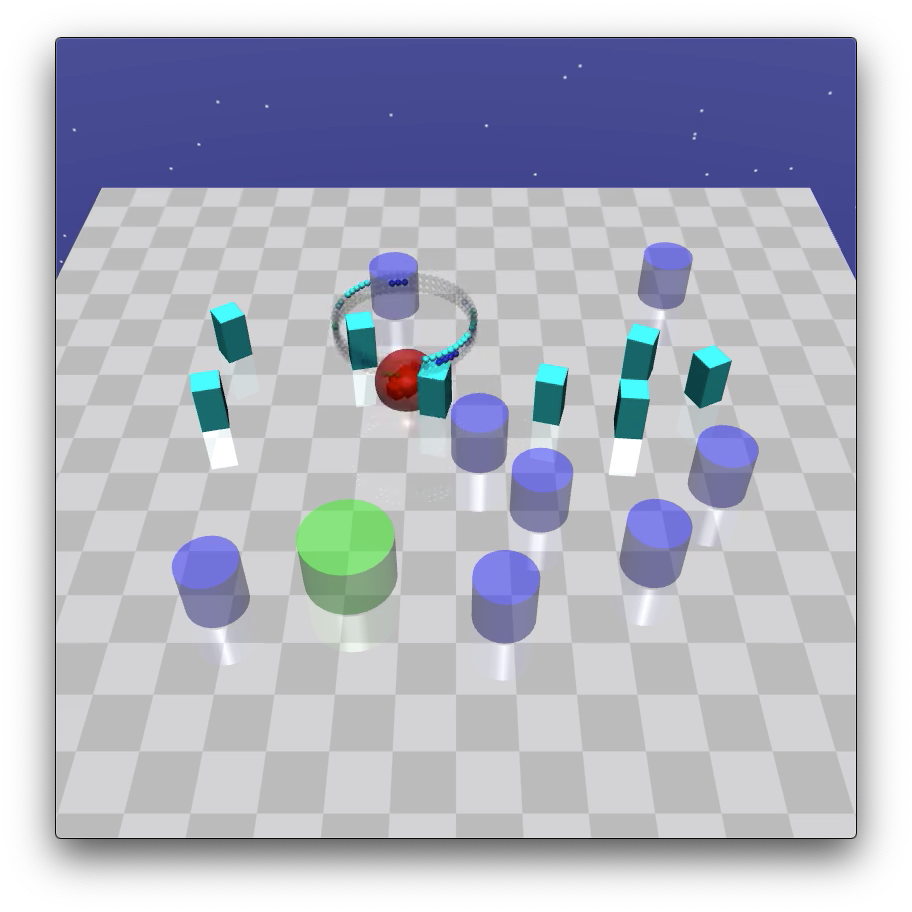}
&\includegraphics[width=0.2\textwidth]{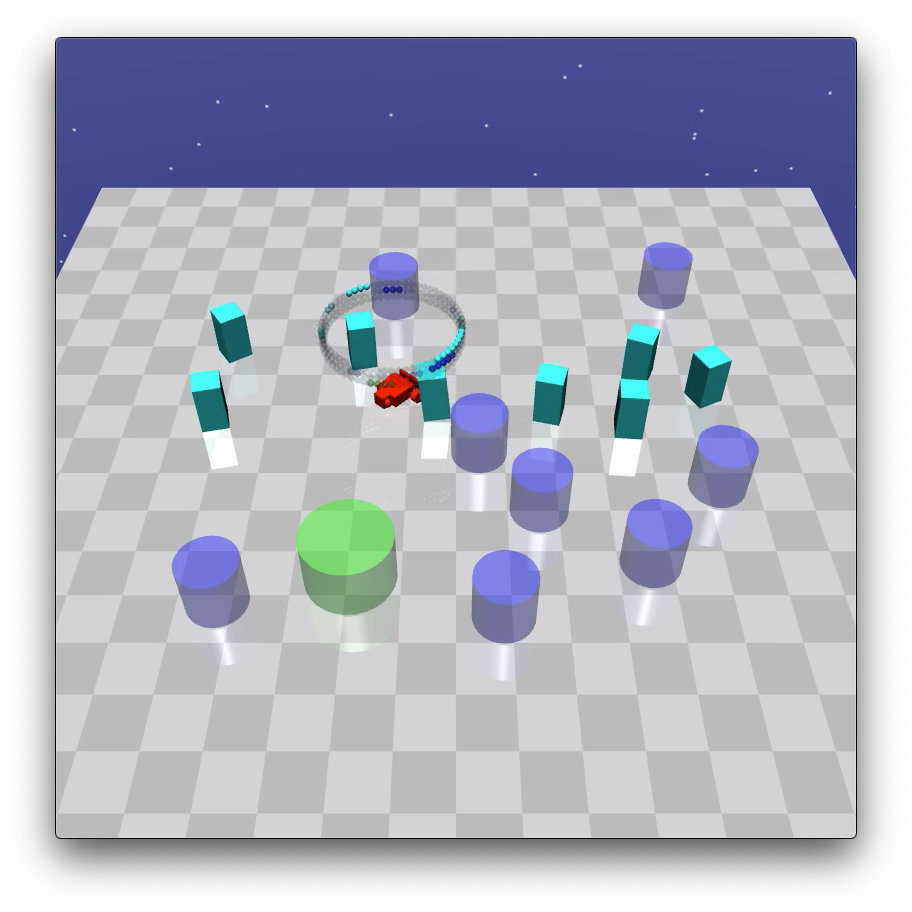}\\
\textsc{PointButton2} & & & & &\\
\includegraphics[width=0.2\textwidth]{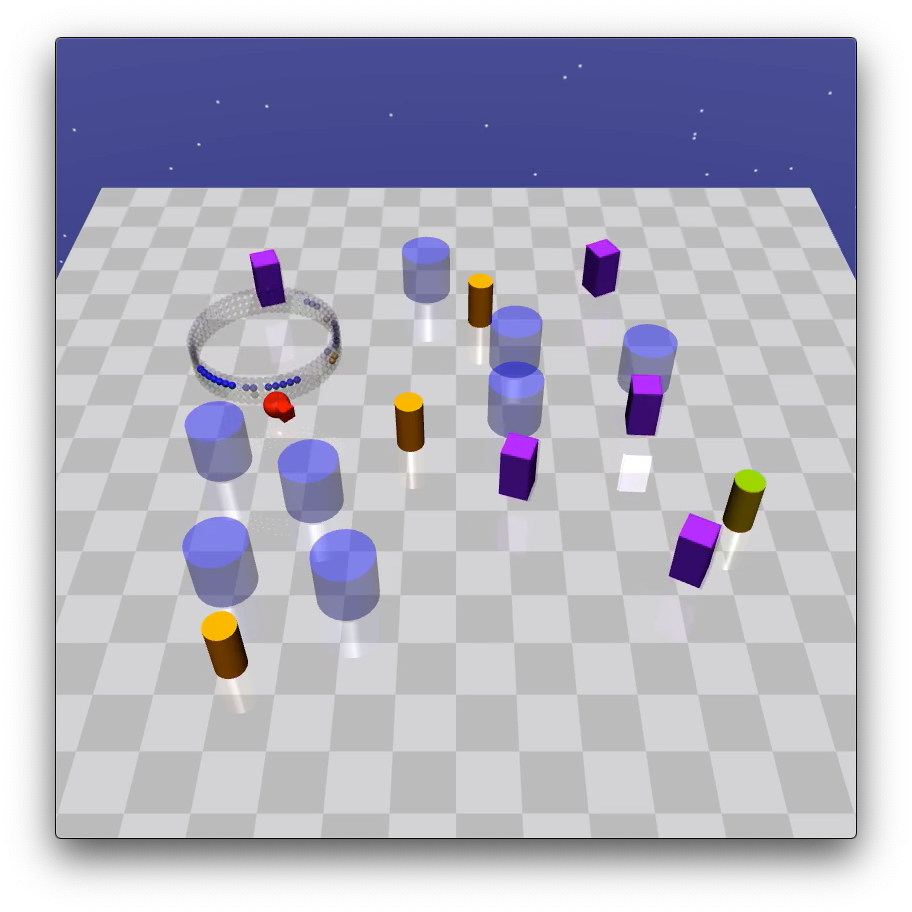}
&\includegraphics[width=0.2\textwidth]{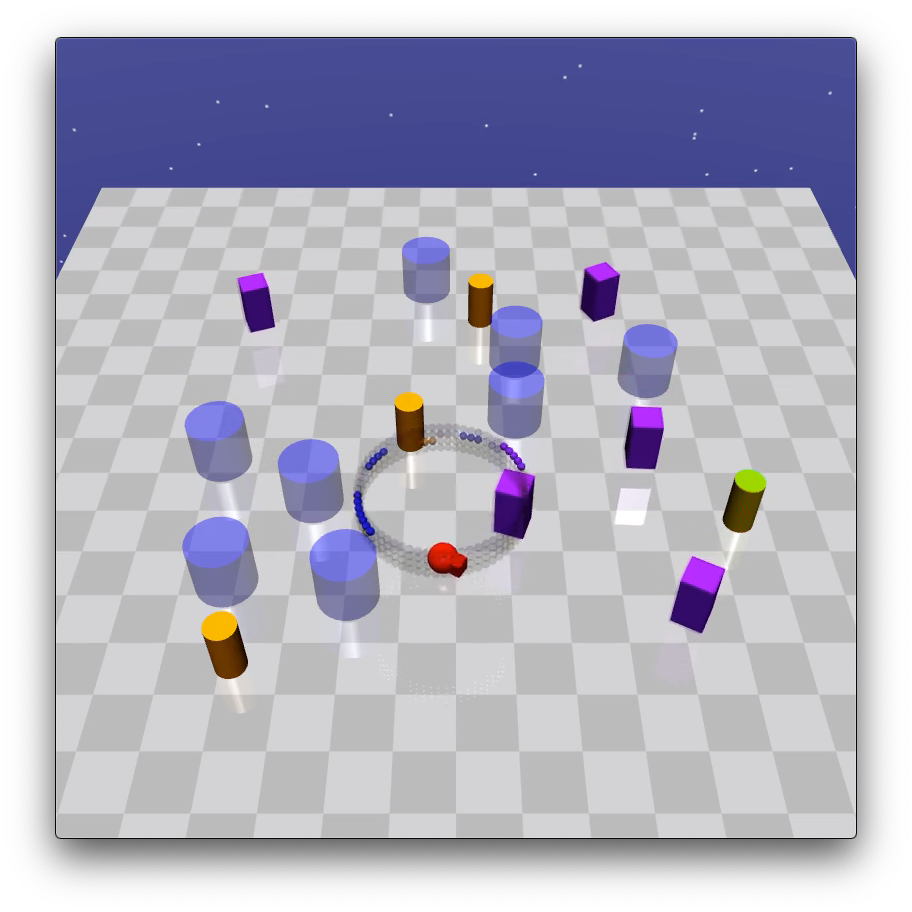}
&\includegraphics[width=0.2\textwidth]{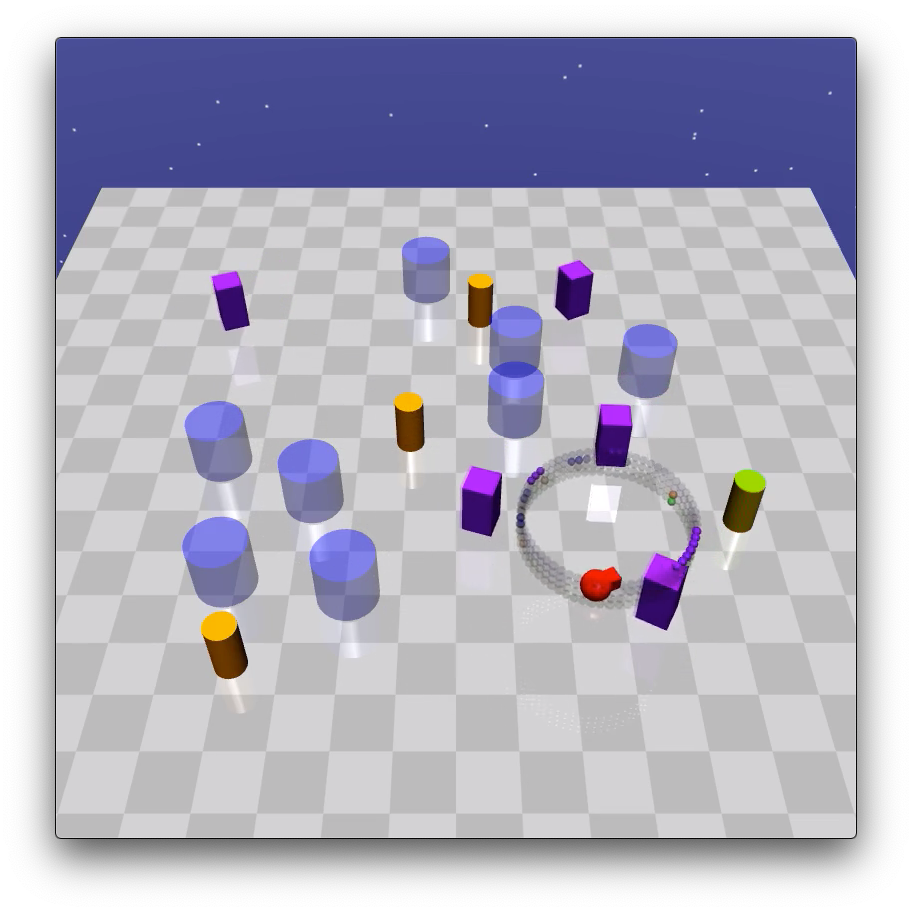}
&\includegraphics[width=0.2\textwidth]{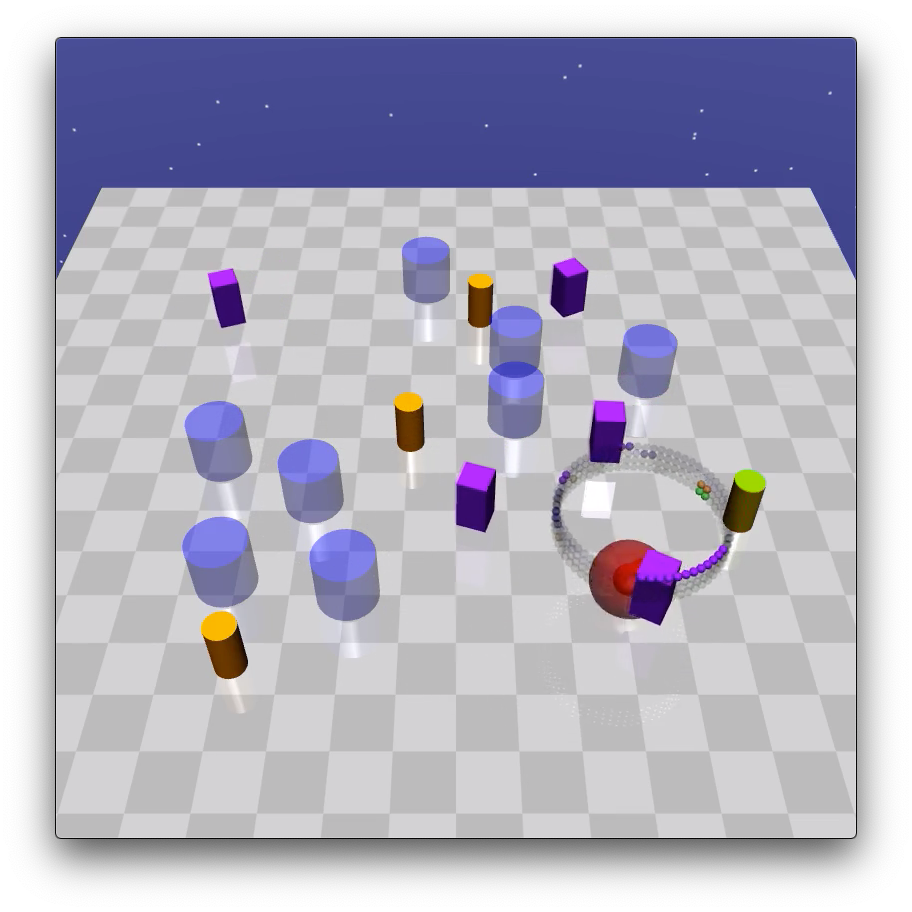}
&\includegraphics[width=0.2\textwidth]{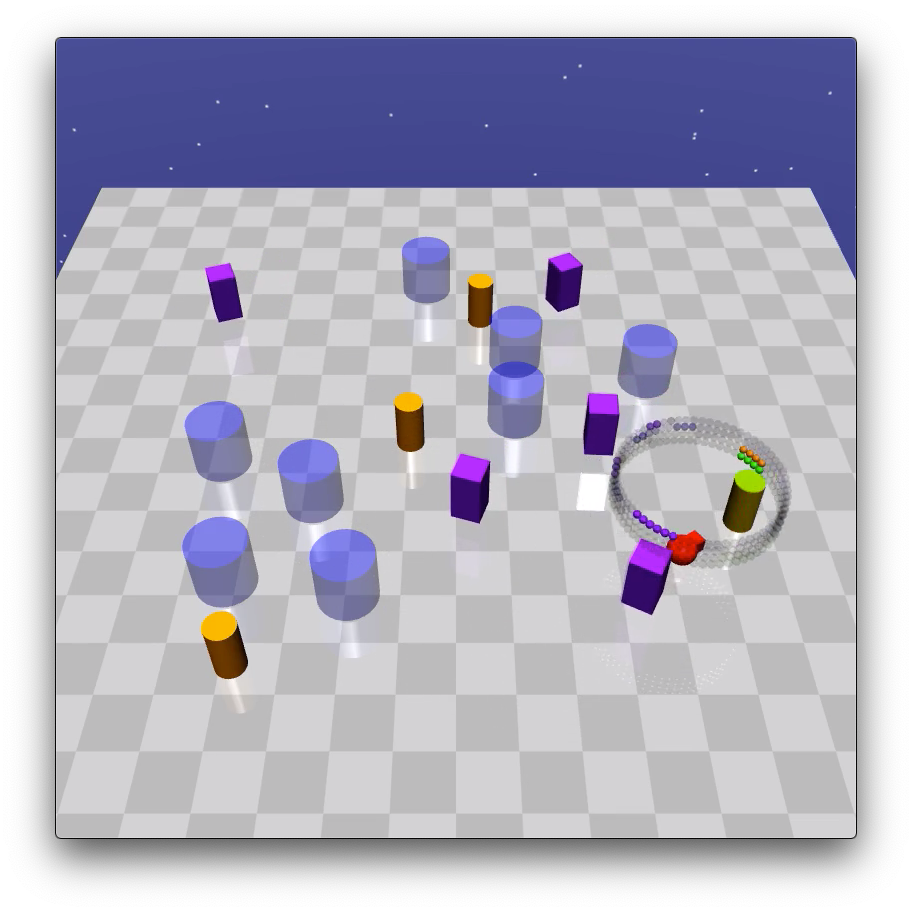}
&\includegraphics[width=0.2\textwidth]{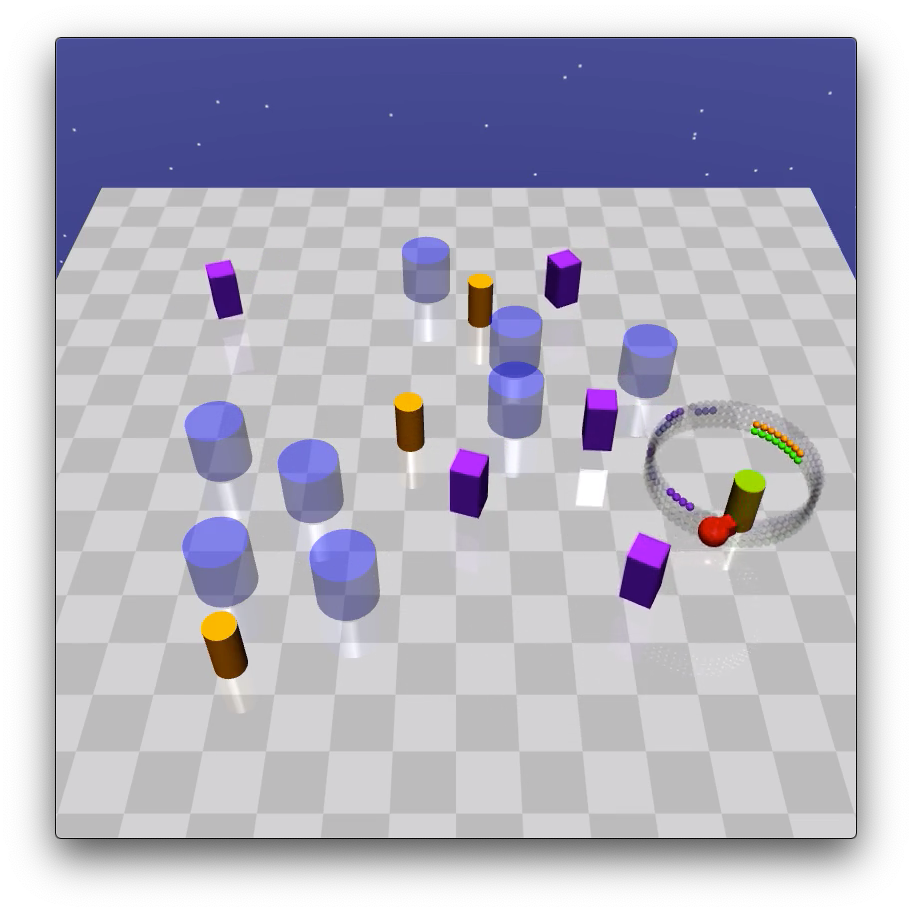}\\
\textsc{CarPush2} & & & & &\\
\includegraphics[width=0.2\textwidth]{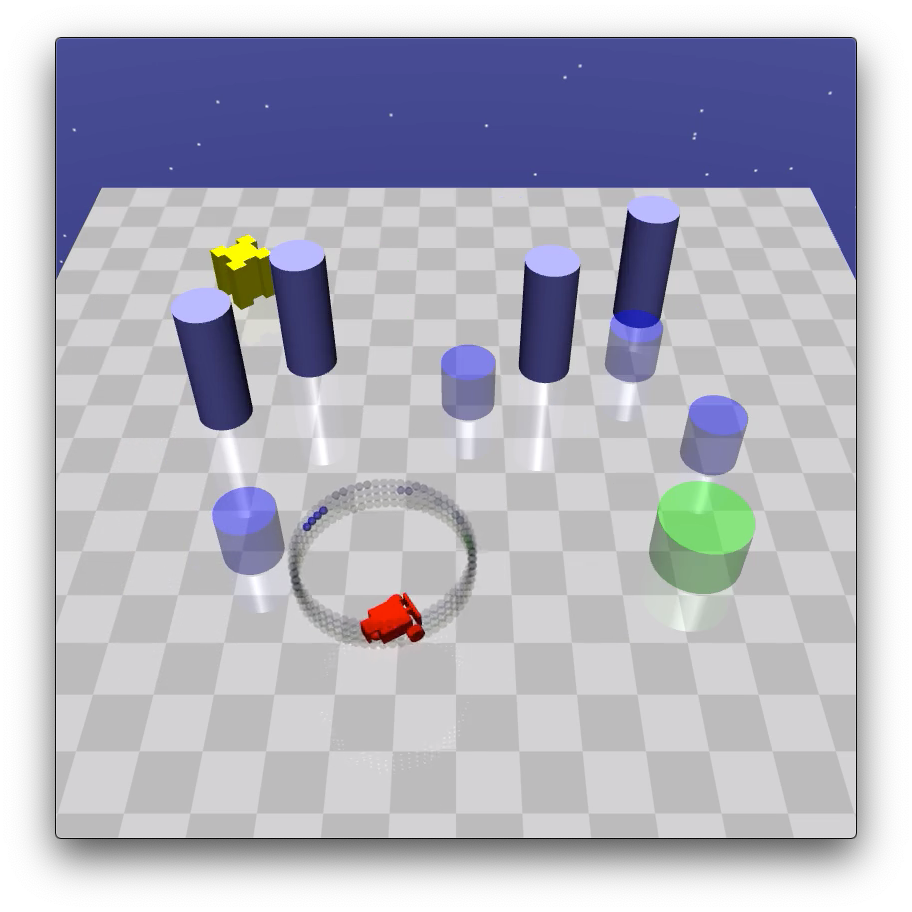}
&\includegraphics[width=0.2\textwidth]{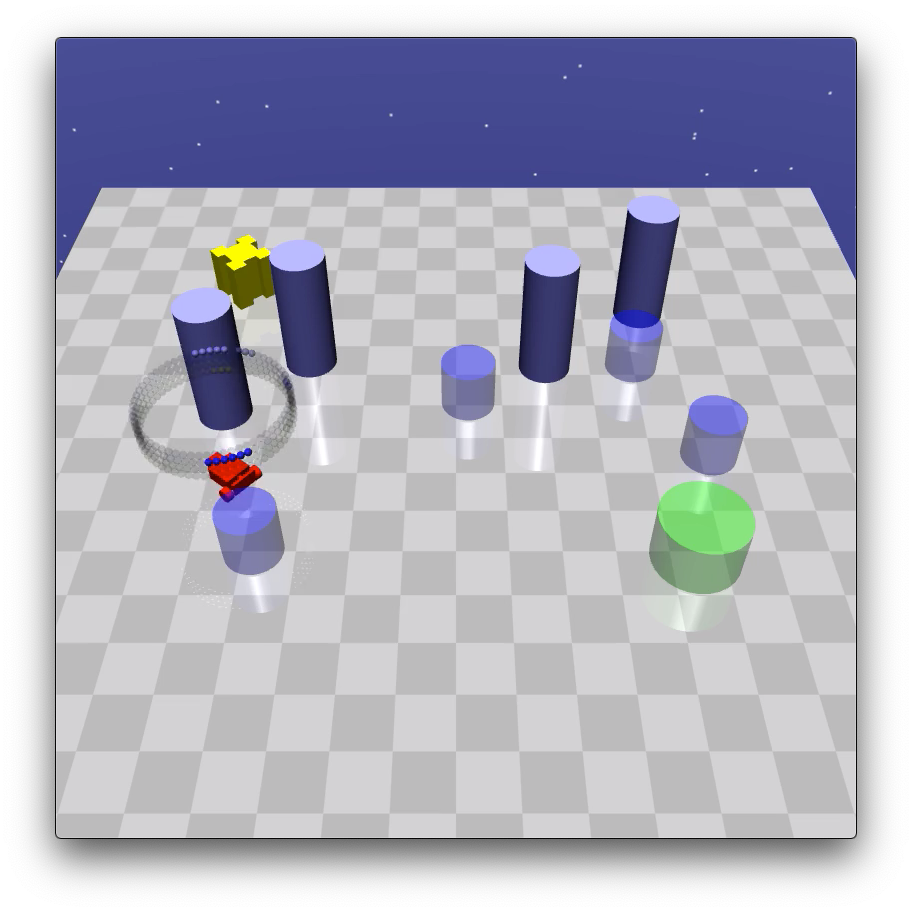}
&\includegraphics[width=0.2\textwidth]{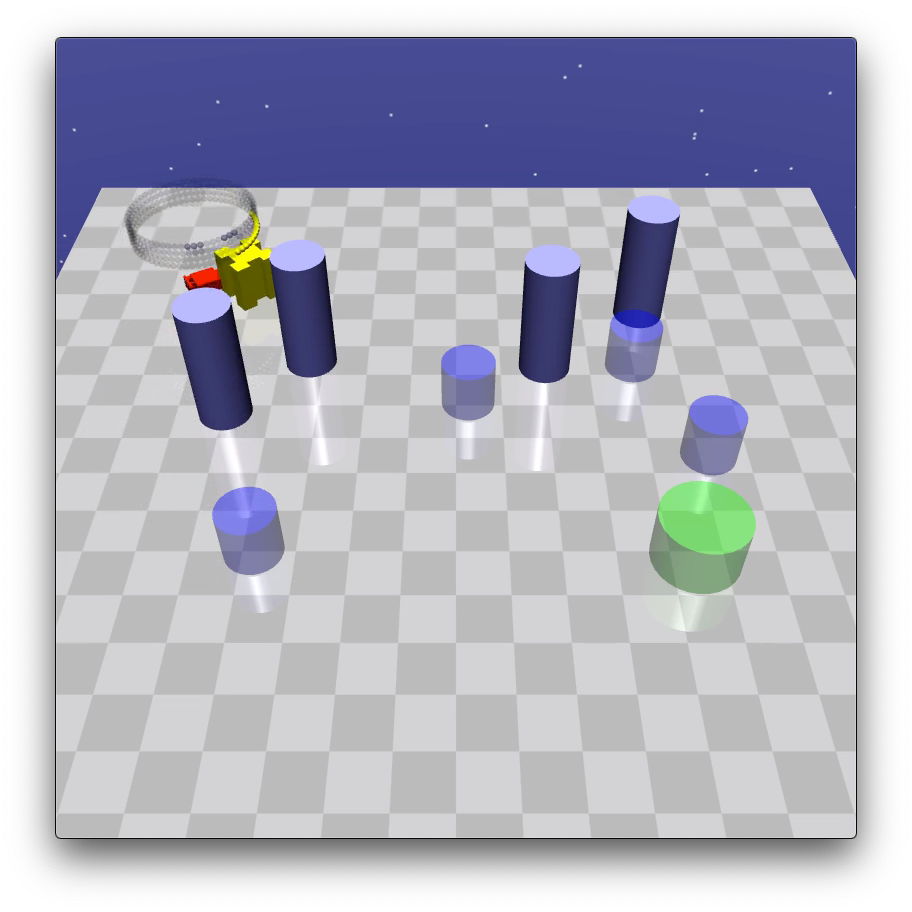}
&\includegraphics[width=0.2\textwidth]{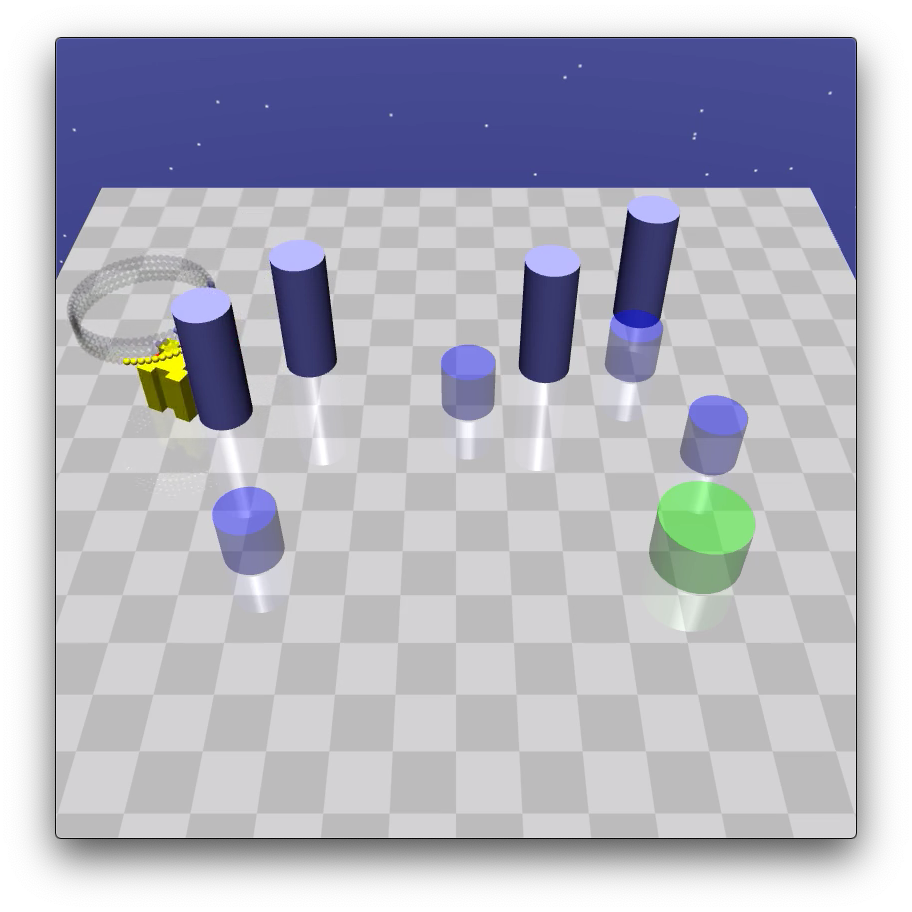}
&\includegraphics[width=0.2\textwidth]{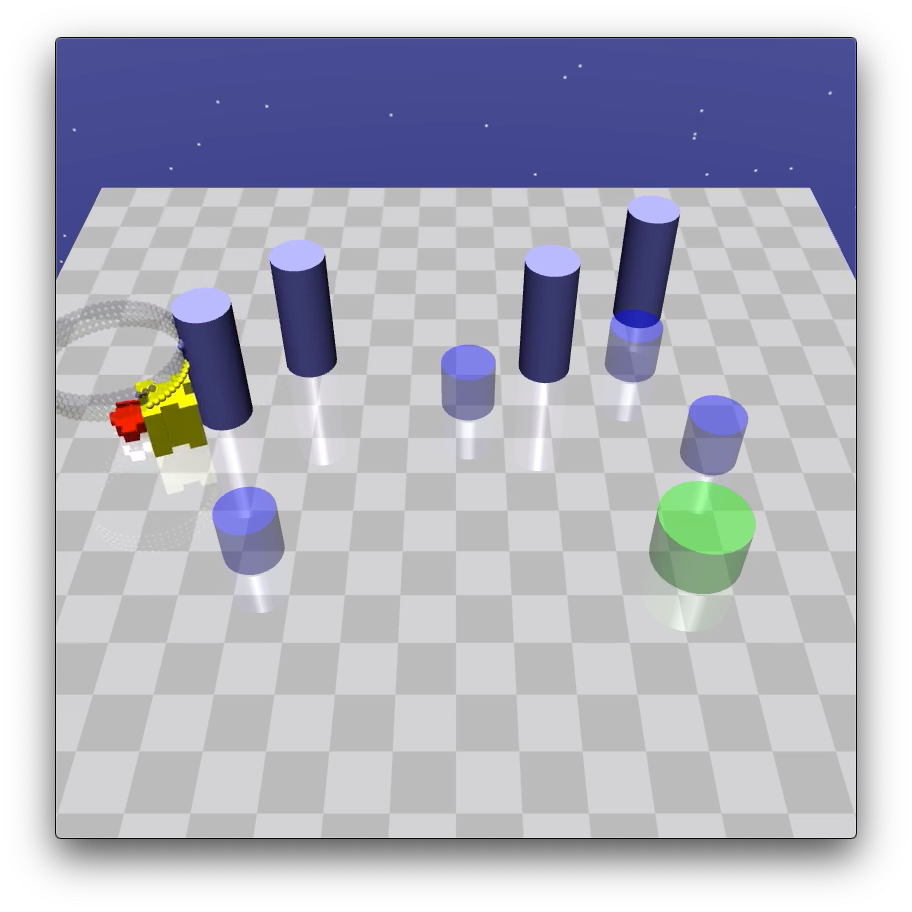}
&\includegraphics[width=0.2\textwidth]{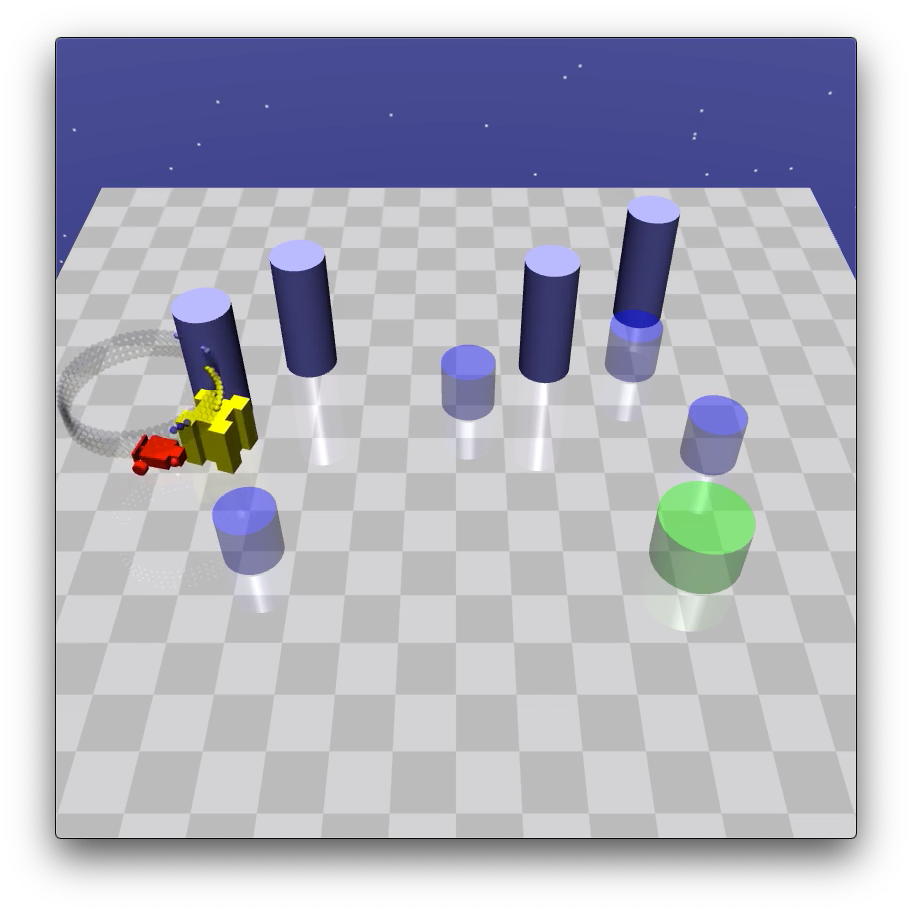}\\
\end{tabular}\\
\begin{tabular}{@{}l@{}c@{}c@{}c@{}}
\textsc{SafeRacingObstacle} & & & \\
\includegraphics[width=0.3\textwidth]{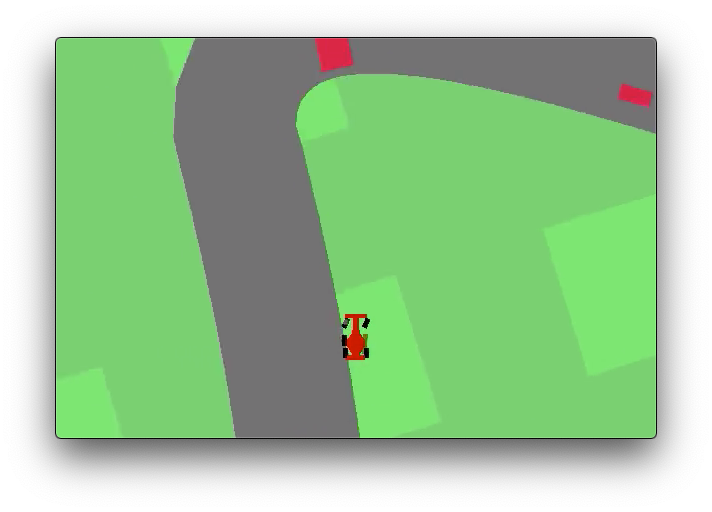}
&\includegraphics[width=0.3\textwidth]{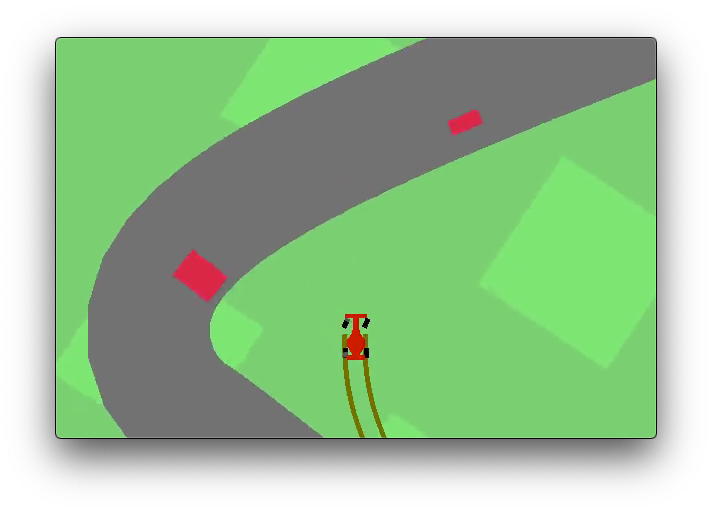}
&\includegraphics[width=0.3\textwidth]{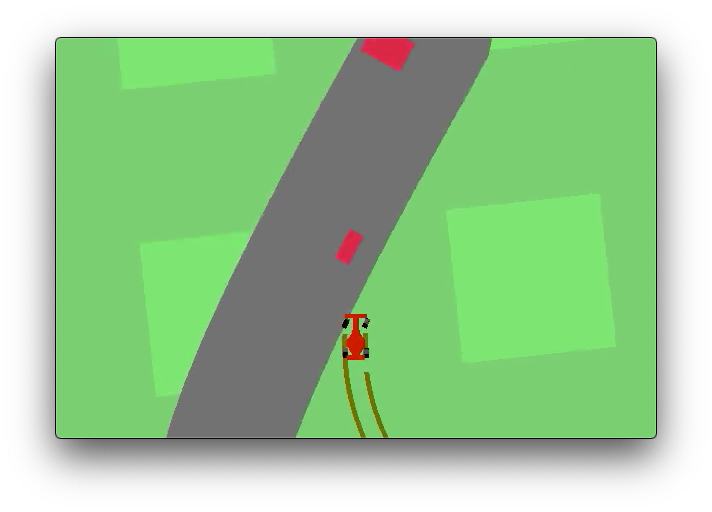}
&\includegraphics[width=0.3\textwidth]{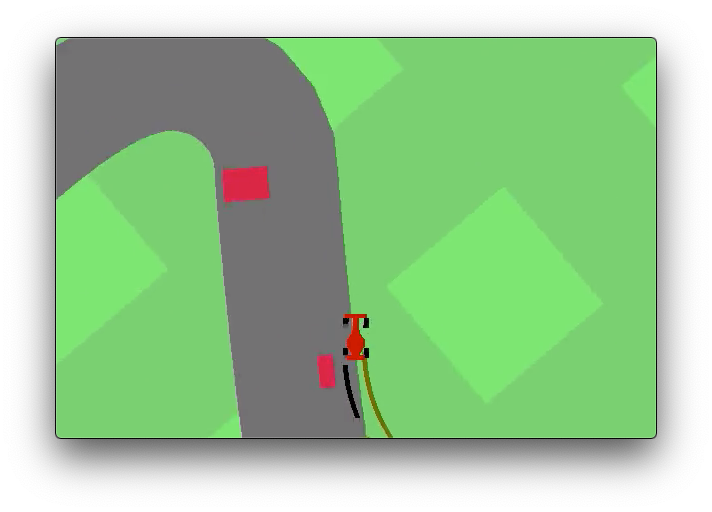}\\
\end{tabular}
\end{tabular}
}
\caption{Key frames of several episodes where the robot violated constraints or only learned a sub-optimal policy.
In Safety Gym, whenever a constraint is violated, a \textcolor{red}{red} sphere is rendered around the robot (the fifth frame of \textsc{CarGoal2} and the fourth frame of \textsc{PointButton2}).
}
\label{fig:failed_episodes}
\end{figure*}

Finally, we show example success and failure modes of \name{} on different tasks in Figure~\ref{fig:success_episodes} and Figure~\ref{fig:failed_episodes}, respectively.
We briefly analyze the failure case of each episode in Figure~\ref{fig:failed_episodes}.
In \textsc{CarGoal2}, the robot faced a crowded set of obstacles in front of it, making its decision very difficult considering the safety requirement.
It took quite some time to drive back and forth, before committing to a path through the two vases in its right front (the fourth frame).
However, when passing a vase, the robot incorrectly estimated its shape and the distance to the vase. 
Even though the majority of its body passed, its left rear tire still hit the vase (the last two frames).
In \textsc{PointButton2}, the robot sped too much in the beginning of the episode, and collided into an oncoming gremlin due to inertia (the floor is slippery!).
It did not learn a precise prediction model of the gremlin's dynamics.
In \textsc{CarPush2}, the robot spent too much time getting the box away from the pillar and did not achieve the goal in time.
These failure cases might be just due to insufficient exploration in similar scenarios. 
In \textsc{SafeRacingObstacle}, the robot learned to take a shortcut for most sharp turns, essentially sacrificing some utility rewards for being safer (skipping obstacles).
The reason is that during every sharp turn with a certain speed, the car's state is quite unstable. 
It requires very precise control to avoid obstacles during this period, which has not been learned by our approach.

\clearpage
\section{Pseudocode}
\begin{algorithm}[h!]
        \SetAlgoLined
        \textbf{Input:} Learning rate $\alpha$\\
        \textbf{Initialize:} Randomize $\theta$, $\phi$, and $\psi$; reset the replay buffer $\mathcal{D}\leftarrow \emptyset$\\
        \For{each training iteration}{
            \textit{// Rollout begins}\\
            Reset the rollout batch $\mathcal{B}_r\leftarrow \emptyset$\\
            \For{each rollout step} {
                Action proposal by UM: $\hat{a} \sim  \piu(\hat{a}|s)$\\
                Action editing by SE: $\Delta a\sim \pis(\Delta a|s,\hat{a})$\\
                Output action $a=h(\hat{a},\Delta a)$\\
                Environment transition $s'\sim \mathcal{P}(s'|s,a)$\\
                Add the transition to the rollout batch $\mathcal{B}_r\leftarrow\mathcal{B}_r\bigcup \{(s,a,s',r(s,a),r_c(s,a))\}$\\
            }
            Store the rollout batch in the buffer $\mathcal{D}\leftarrow\mathcal{D}\bigcup \mathcal{B}_r$\\
            \textit{// Training begins}\\            
            Estimate the gradient of the Lagrangian multiplier $\lambda$ by evaluating Eq.~\ref{eq:lambda_gradient} on $\mathcal{B}_r$\\
            Update the multiplier by Eq.~\ref{eq:lambda0_obj}: $\lambda_0 \leftarrow \lambda_0 - \alpha \Lambda_{\pia}$\\
            Sample a training batch $\mathcal{B}$ from the replay buffer $\mathcal{D}$ for gradient steps below\\
            Perform one gradient step on the critic parameters $\theta$ by TD backup (Eq.~\ref{eq:bellman}) on $Q$ and $Q_c$\\
            Perform one gradient step on UM: $\phi\leftarrow \phi + \alpha\Delta\phi$ (gradient of Eq.~\ref{eq:objectives2}, a)\\
            Perform one gradient step on SE: $\psi\leftarrow \psi + \alpha\Delta\psi$ (gradient of Eq.~\ref{eq:objectives2}, b)\\
            Update other parameters such as entropy weight, target critic network, etc.\\
        }
        \caption{\name{}}
    \label{alg:taac}
\end{algorithm}


\end{document}